\documentclass[10pt,twocolumn,letterpaper]{article}

\usepackage{iccv}
\usepackage{times}
\usepackage{epsfig}
\usepackage{graphicx}
\usepackage{amsmath}
\usepackage{amssymb}

\usepackage{graphicx}
\usepackage{amsmath}
\usepackage{amssymb}
\usepackage{booktabs}
\usepackage{adjustbox}
\usepackage{graphicx}
\usepackage{color}
\usepackage{tabularx}
\usepackage{xfrac}
\usepackage{enumitem}
\usepackage{subfig}
\usepackage{siunitx}
\usepackage{svg}
\usepackage{makecell}
\usepackage{array}
\newcolumntype{P}[1]{>{\centering\arraybackslash}p{#1}}
\newcolumntype{M}[1]{>{\centering\arraybackslash}m{#1}}
\newcolumntype{C}{>{\centering\arraybackslash}X}

\usepackage{titling}
\date{}\predate{}\postdate{}
\pretitle{\begin{center}\Large \bf}
\posttitle{\end{center}\vskip 0.5em}

\usepackage[pagebackref=true,breaklinks=true,letterpaper=true,colorlinks,bookmarks=false]{hyperref}
\newcommand{\Sec}{Section~}
\newcommand{\Fig}{Fig.~}
\newcommand{\Tab}{Table~}

\iccvfinalcopy %

\def\iccvPaperID{12141} %

\ificcvfinal\fi

\begin{document}

\def\MYTITLE{Seeing Behind Dynamic Occlusions with Event Cameras}
\title{\MYTITLE}

\author{Rong Zou
\qquad 
Manasi Muglikar
\qquad 
Nico Messikommer
\qquad 
Davide Scaramuzza \\
\,Robotics and Perception Group, University of Zurich, Switzerland \\
}

\maketitle

\begin{abstract}

Unwanted camera occlusions, such as debris, dust, raindrops, and snow, can severely degrade the performance of computer-vision systems.
Dynamic occlusions are particularly challenging because of the continuously changing pattern. 
Existing occlusion-removal methods currently use  synthetic aperture imaging or image inpainting. 
However, they face issues with dynamic occlusions as these require multiple viewpoints or user-generated masks to hallucinate the background intensity.
We propose a novel approach to reconstruct the background from a single viewpoint in the presence of dynamic occlusions.
Our solution relies for the first time on the combination of a traditional camera with an event camera. %
 When an occlusion moves across a background image, it causes intensity changes that trigger events.
These events provide additional information on the relative intensity changes between foreground and background at a high temporal resolution, enabling a truer reconstruction of the background content. 
We present the first large-scale dataset consisting of synchronized images and event sequences to evaluate our approach.
We show that our method outperforms image inpainting methods by 3dB in terms of PSNR on our dataset. 

\end{abstract}

\section{Introduction}
The majority of computer vision algorithms operate under the assumption that the scene being analyzed is visible in its entirety and without any obstructions. 
However, occlusions resulting from non-ideal environmental factors, such as dirt from road debris, insects, raindrops, snow, and others, have the potential to significantly reduce the quality of captured images. 
The dynamic nature of these occlusions, affects the downstream applications such as autonomous navigation, object detection, tracking, 3D reconstruction, etc. 
Dynamic occlusions pose an interesting challenge for scene understanding, especially for robotics and autonomous driving, where it is crucial to know the occluded scene for safe navigation in unknown environments.
\begin{figure}[t]
    \centering
    \setlength{\tabcolsep}{2pt}
    \includesvg[width=\linewidth]{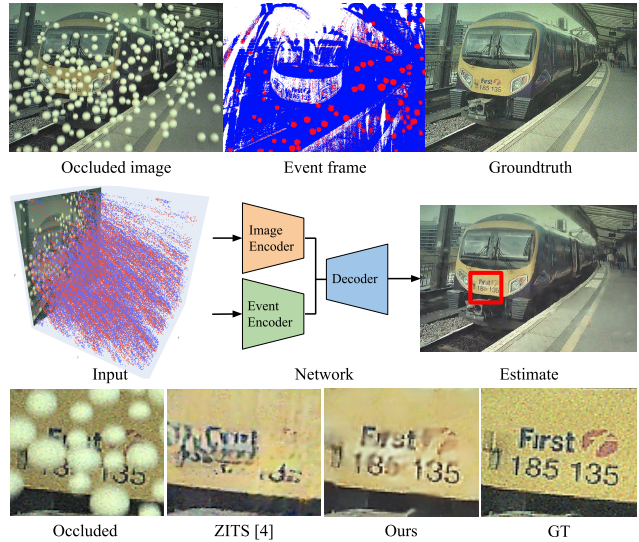}
    \caption{
    (Top) The background image (right) occluded by moving particles observed with a traditional camera (left) and event camera (center).
    (Middle) The proposed method for background reconstruction in the presence of moving objects uses the continuous event stream and the occluded image to learn to reconstruct the background image.
    (Bottom) Our approach outperforms the image inpainting baseline and accurately reconstructs the background intensity.
    }
    \label{fig:eye_catch}
    \vspace{-2ex}
\end{figure}

The literature, however, has focused more on static occlusions and many different approaches have been proposed that use multiple viewpoints (synthetic aperture imaging) \cite{liu20cvpr, wang20wacv}, optical diffraction cloaking \cite{shi22acmtg}, or image inpainting \cite{bertalmio00siggraph, hays07siggraph, liu18eccv,liu19iccv, suvorov22wacv, wan2021cvpr, liu22cvpr, dong22cvpr, li_mat22cvpr}.
The assumption of static occlusion often renders the above approaches infeasible in the presence of dynamic obstructions \cite{Yang14eccv, Tauber07tsmcc}.
Image inpainting approaches are often trained on large-scale datasets to predict intensity in areas specified by a pre-defined mask. 
However, these methods hallucinate the scene since they are trained to favour more aesthetically pleasing images over the true scene content as seen in \Fig \ref{fig:eye_catch}.
While there have been several advancements in these approaches with the advent of deep learning, these methods fail to reconstruct true scene content in the presence of dynamic occlusions.
Synthetic Aperture Imaging (SAI) captures the target scene from multiple viewpoints to create a virtual camera with a large-aperture lens \cite{Wilburn05tog, Yang14eccv}. 
This makes it easier to blur the foreground occlusion and refocus the image on the background scene. 
These methods assume the occlusions are at the same depth and stationary throughout the capture process, which can be up to a few milliseconds, rendering this method infeasible for fast-moving occlusions.

In this paper, we propose a novel approach to estimate the background image in the presence of dynamic occlusions using a single viewpoint.
Our solution complements a traditional camera with an event camera.
Event cameras \cite{Gallego20pami} %
 are bio-inspired sensors that asynchronously measure changes in brightness with microsecond resolution.
When an occlusion crosses over a background image, it causes changes in intensity, thus triggering events as seen in \Fig \ref{fig:eye_catch}.
These events give additional information on the relative intensity changes between the foreground and background at a high temporal resolution, which leads to a more precise reconstruction of the background, compared to image inpainting methods which only use unoccluded image statistics. 
Since this is the first time an event camera is used to tackle dynamic occlusion removal, we collect a large-scale dataset containing 233 challenging scenes with synchronized events, occluded images, and ground truth unoccluded images.

The classical event generation model \cite{Lichtsteiner08ssc} can be used to reconstruct the background image intensity, given contrast threshold and occlusion intensity.
However, due to inherent sensor noise and instability of the contrast threshold, this approach results in poor performance at higher occlusion densities.
We, therefore, propose to use a supervised data-driven approach that learns to reconstruct the unoccluded image, implicitly learning the sensor parameters and noise characteristics.
The network predicts the background intensity using only a single occluded image and events
We show that our method relying on a single frame and events leads to a performance improvement of \SI{3.3}{dB} on our synthetic dataset and \SI{2.8}{dB} on our real dataset over the image-inpainting baselines in terms of PSNR.
Our results show that our method is capable of accurately reconstructing scenes in the presence of dynamic occlusions and has the potential to increase perceptual robustness in unknown environments. %
To summarize, our contributions are:
\begin{itemize}
    \item A novel solution to the problem of background image reconstruction in the presence of dynamic occlusions using a single viewpoint. We are the first to tackle this problem using an event camera in addition to a standard camera.
    \item A large-scale dataset recorded in the real world containing $233$ challenging scenes for background reconstruction with synchronized events, occluded frames, and groundtruth unoccluded images. 
\end{itemize}

\section{Related works}

\subsection{Fame-based methods}
Image inpainting is the task closest related to our tackled problem of removing dynamic occlusions in a single camera view.
In image inpainting, the goal is to reconstruct missing pixel intensities in areas specified by a pre-defined mask.
Traditional image inpainting methods rely on the statistics of the unmasked image areas to directly infer the missing regions~\cite{bertalmio00siggraph, ballester01tip} or copy textures from database images~\cite{criminisicvpr03, hays07siggraph}.
Because of the ability to account for the context in masked images, current research on image inpainting mostly focuses on deep generative approaches~\cite{iizuka17siggraph,liu18eccv,liu19iccv, nazeri19iccv, pathak16cvpr, yu2018cvpr}.
To provide global contextual information early on in the network, LaMa\cite{{suvorov22wacv}} proposes to perform convolutions in the Fourier domain.
In~\cite{guo21iccv}, a two-stream network fuses texture and structure information for a more accurate image generation.
To generalize to different scenes, MISF~\cite{{li_misf22cvpr}} proposes to apply predictive filtering on the feature level to reconstruct semantic information and on the image level to recover details.
More recently, transformers were also applied to the task of image inpainting~\cite{wan2021cvpr, liu22cvpr, dong22cvpr, li_mat22cvpr} and achieve state-of-the-art performance.
In PUT~\cite{liu22cvpr}, an encoder-decoder is applied to reduce the information loss suffered by tokenizing and dekonizing the tokens for the transformer backbone.
Instead of directly predicting the pixel intensities, ZITS~\cite{{dong22cvpr}} uses a transformer model to restore the structures with low resolution in form of edges, which are then further processed by a CNN.
Generally, since the only information for reconstructing image patches is given by the structure and context of the unmasked regions, image inpainting methods aim to reconstruct plausible and aesthetically pleasing images without the focus on reconstructing the real scene content. 
Additionally, the application of image inpainting methods for occlusion removal requires the estimation of the occlusion mask.
Therefore, image inpainting methods tackle a different problem setting and are not designed for the task of removing dynamic occlusions in a single-camera view.

Instead of considering only one image, a sequence of images taken at different positions can be leveraged to remove reflections and static occlusions~\cite{liu20cvpr} or contaminant artifacts on lenses~\cite{li21iccv}.
In the field of synthetic aperture imaging (SAI), different camera views are used to estimate the incoming lights from the background onto a synthetic camera view.
By considering a large number of images captured at different positions, SAI methods can focus on different planes in the scene, effectively removing the static occlusions by blurring them out~\cite{Wilburn05tog, vaish04cvpr, vaish06cvpr, matusik12cvpr, Yang14eccv, pei13pr}.
Furthermore, various imaging setups with hardware modifications are developed.
In DeOccNet~\cite{wang20wacv}, multiple views of a light field camera are used in an encoder-decoder fashion to remove the static occlusion.
Capturing the same scene from multiple viewpoints constrains the occlusion to be stationary during the capture process, which can range from a few milliseconds to seconds depending on the camera speed and exposure time.
Our problem, therefore, becomes more challenging to solve with SAI, as these methods are designed to remove static occlusions by considering multiple camera views.
Another interesting work~\cite{shi22acmtg} proposes to learn a diffractive optical element, which can be inserted in front of the lens to focus better on the objects further away from the camera, which effectively removes thin occluders close to the camera.

\subsection{Event-based methods}
It was shown that event cameras are capable of generating image sequences in high-speed and high-dynamic range scenes in which standard cameras fail~\cite{Rebecq19cvpr, Rebecq19pami, Mostafavi19cvpr, zhang20eccv}.
Moreover, event cameras are also successfully applied in combination with frame cameras for different imaging tasks.
Specifically, the high temporal resolution of event cameras is used for filling in missing information between two frames to create high frame-rate videos~\cite{Tulyakov22cvpr, wang-et-al-2020, Zhiyang21iccv} or to deblur images~\cite{Songnan20eccv, Pan19cvpr}, whereas the high-dynamic range property is leveraged in HDR imaging~\cite{han-et-al-2020,wang-et-al-2020, Messikommer22cvprw}.
Recently, event cameras achieved impressive results for removing static occlusions using a moving camera, which is termed event-based synthetic aperture~\cite{zhang21cvpr, yu22pami}.
In~\cite{{liao22cvpr}}, frames are additionally considered to reconstruct the background information behind a static, dense occlusion.
In contrast to our proposed method for dynamic occlusion removal, the stated event-based synthetic aperture approaches tackle a different task of removing static occlusions, which are captured at different positions using a moving camera.

\section{Method}
\paragraph{Problem formulation}
Let us assume an event-based dynamic occlusion setting, where we are given an occluded frame $I_0$ at timestep $0$ and an event sequence recorded between timestep $0$ and $t$.
We aim to reconstruct the background image from event sequences by integrating information from events and the occluded image.
\paragraph{Basic considerations} 
\label{sec:method:basic}
Event-cameras are novel, bio-inspired sensors that asynchronously measure \emph{changes} (i.e., temporal contrast) in illumination at every pixel, at the time they occur \cite{Lichtsteiner08ssc,Suh20iscas,Finateu20isscc, Posch11ssc}.
When a particle occludes a background pixel at position $x_k$ at time $t$, the intensity changes from $L(\mathbf{x_k},t_k-\Delta t_k)$ to $L(\mathbf{x_k},t_k)$. 
In particular, an event camera generates an event $e_k = (\mathbf{x}_k,t_k,p_k)$ at time $t_k$ when the difference of logarithmic brightness at the same pixel $\mathbf{x}_k=(x_k,y_k)^\top$  reaches a predefined threshold $C$:
\begin{equation}
\label{eq:egm}
    L(\mathbf{x_k},t_k) - L(\mathbf{x_k},t_k-\Delta t_k) = p_k C,
\end{equation}
where $p_k \in \{-1,+1\}$ is the sign (or polarity) of the brightness change, and $\Delta t_k$ is the time since the last event at the pixel $\mathbf{x}_k$.
The result is a sparse sequence of events that are asynchronously triggered by illumination changes.
To reconstruct the original intensity, we use the above event generation model as follows:
\begin{equation}
    L(\mathbf{x_k},t_k) = L(\mathbf{x_k},t_k-\Delta t_k) + \sum p_k C,
\end{equation}
Using the above equation, the intensity of the occluded pixels can be estimated under the condition that the contrast threshold $C$ and the position of the occlusion are known.
Thus, a model-based method requires segmenting the occluded pixels and estimating the contrast threshold.

We design such a method as a baseline for comparison and refer to it as \textit{Accumulation Method}.
The designed method assumes that the occluded pixels have similar intensities and uses this heuristic to segment the occluded pixels from the background.
However, this method breaks down in the presence of multiple overlapping occlusions, which leads to an occluded area in the image with different intensities.
For the estimation of the unknown contrast threshold $C$, as is commonly done, a reasonable value is defined for the event accumulation.

\subsection{Network Architecture}
\begin{figure*}[ht!]
    \centering
    \includegraphics[width=0.95\textwidth]{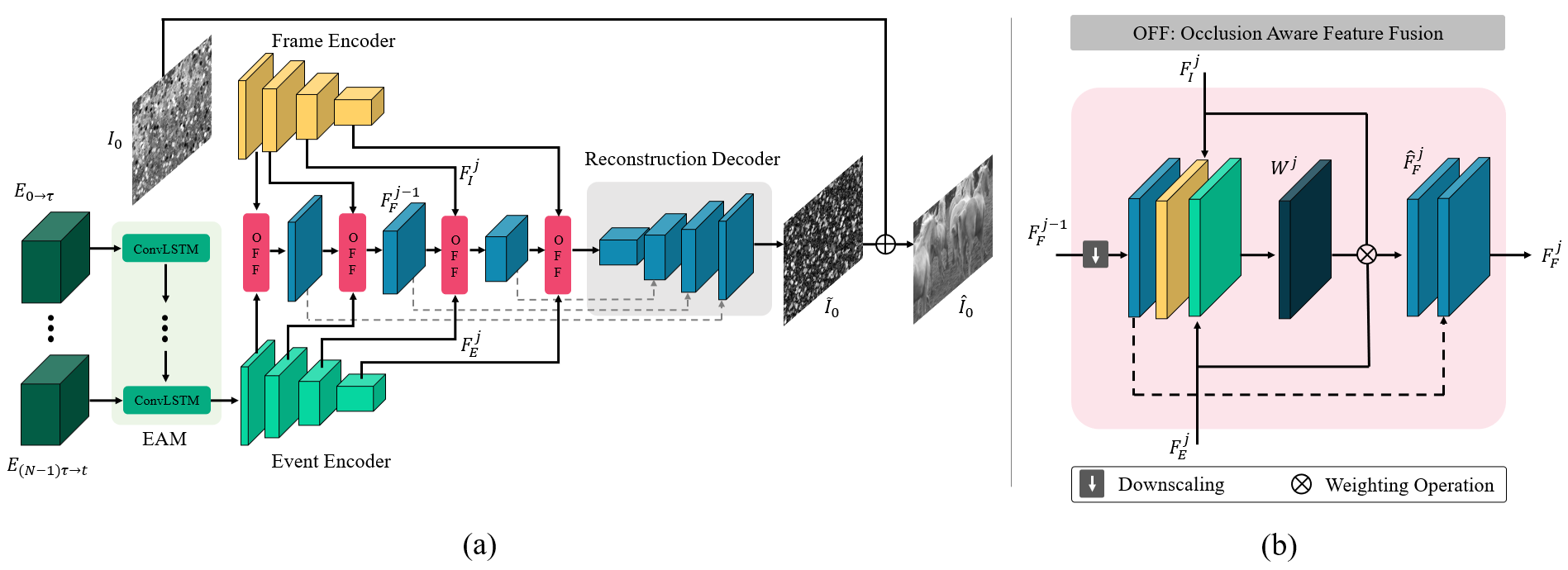}
    \caption{
    As shown in (a), our proposed method first integrates $N$ event representations $E_{0\rightarrow \tau}, ..., E_{(N-1) \tau \rightarrow t}$ using ConvLSTM layers in our \textit{Event Accumation Module} (EAM).
    The final hidden state of the ConvLSTM layers is processed with the multi-scale \textit{Event Encoder}.
    Similarly, the occluded input image $I_0$ is processed with a \textit{Frame Encoder}.
    At each scale $j$, the \textit{Occlusion-aware Feature Fusion} (OFF) module (b) decides in spatial and channel dimensions which frame features $F_I^j$ and event features $F_E^j$ to fuse.
    The fused features $F_F^j$ are reconstructed to a residual image $\tilde{I}_0$ using the \textit{Reconstruction Decoder}.
    The occlusion-free image $\hat{I}_0$ is obtained by adding the residual image to the input image $I_0$.
    }
    \label{fig:method_overview}
\end{figure*}
We design a learning-based method to tackle these deficiencies by implicitly learning the contrast threshold as well as detecting the occluded areas using the input frame and events.
The proposed network takes as input a single occluded image $I_0$ and $N$ event representations $E_{0\rightarrow \tau}, ..., E_{(N-1) \tau \rightarrow t}$, which are constructed in time intervals of $\tau$ based on the events recorded immediately after the image between timestep $0$ and $t$.
The proposed fully convolutional network leverages ConvLSTM layers~\cite{xingjian2015convolutional} to integrate the events and follows a U-Net structure~\cite{Ronneberger15icmicci} to predict a residual image $\tilde{I}_0$.
The residual image is added to the occluded input image to obtain the final occlusion-free intensity image $\hat{I}_0$, which should contain the true background information at the occluded areas of the input image.
The overview of our network is visualized in Fig.~\ref{fig:method_overview} (a).

As a first step, we integrate the event information in the \textit{Event Accumulation Module} (EAM) to retrieve accumulated intensity changes since the events capture instantaneous changes caused by the movement of the occlusions.
The Event Accumulation Module should focus on the intensity changes relevant to the background information, i.e., it should select the intensity information at the timesteps at which the background behind the occlusions becomes visible. 
To achieve this, we integrate the event information at multiple timesteps and let the network select the intensity change corresponding to the unoccluded background. This also helps the network to ignore events triggered by overlapping occlusions. \
We make use of the accumulation and gating mechanism of ConvLSTM layers to model the event integration mechanism, as well as the timestep selection.
Specifically, we apply two ConvLSTM layers in sequence to recurrently process the event representation starting with an initial zero-valued hidden and cell states.
The output of the Event Accumulation Module represents the last hidden state of the last ConvLSTM layer at the timestamp of the final event representation.

The integrated event features are then further encoded using a multi-scale \textit{Event Encoder} consisting of a convolution layer and a batch normalization layer with ReLU activation per scale to generate event features.
In a similar fashion, the occluded image is processed using a corresponding \textit{Frame Encoder}. In total there are $j{=}6$ scales.
Due to the differential principle of event cameras, we noticed that event features are highly beneficial for the reconstruction of textures, e.g., edges, but struggle to reconstruct uniform image areas.
In contrast, standard frames contain absolute intensities and thus are better suited for filling in uniform areas since, in the simplest case of constant pixel values, a border value copying suffices.
Because of this duality, we process event and frame features separately and employ a spatial as well as channel-wise weighting between them at each encoding scale $j$.
The spatial weighting is done in the \textit{Occlusion-aware Feature Fusion} (OFF) module, which computes gating weights $W^j$ based on the event features $F_E^j$ and frame features $F_I^j$ at the considered scale $j$ as well as the fused features from the previous scale $F_F^{j-1}$, which are downscaled by a strided convolution layer.
Specifically, the event and frame features are fused by weighting them with $W^j$ and $1-W^j$ and summing them up.
\begin{equation}
    \hat{F}_F^j = (1 - W^j) F_I^j + W^j F_E^j,
\end{equation}
Together with the fused feature of the previous scale, the weighted sum of the event and frame features is given as input to convolution layers outputting the fused features $F_F^{j}$.
The \textit{Occlusion-aware Feature Fusion} module is visualized in Fig.~\ref{fig:method_overview} (b).

Finally, the fused features at each scale are used as skip connections inside a \textit{Reconstruction Decoder}.
For upsampling, we use a bilinear interpolation followed by a convolution layer.
The output of the \textit{Reconstruction Decoder} is the residual image $\tilde{I}_0$, which is added to the input image $I_0$ and given to a sigmoid function resulting in the final occlusion-free image $\hat{I}_0$.
For more details about the network structure, we refer to the supplementary.

\section{Dataset}
To the best of our knowledge, there exists no dataset aimed at tackling the problem of dynamic occlusion removal using an event camera.
Therefore, we generated a new synthetic dataset containing frames and events to compare our approach against existing baselines in controlled conditions.
Furthermore, to validate our approach in the real world, we record a novel dataset with real events and images.
Both datasets lay the foundation for future research in dynamic occlusion removal using event cameras.

\subsection{Synthetic Dataset}
\global\long\def\figWidth{0.3\linewidth}
\begin{figure}
	\centering
    \setlength{\tabcolsep}{2pt}
	\begin{tabular}{
	M{0.35cm}
	M{\figWidth}
	M{\figWidth}
	M{\figWidth}}
		& Groundtruth & Occluded & Events
		\\
        \rotatebox{90}{\makecell{10\%}}
		& \frame{\includegraphics[width=\linewidth]{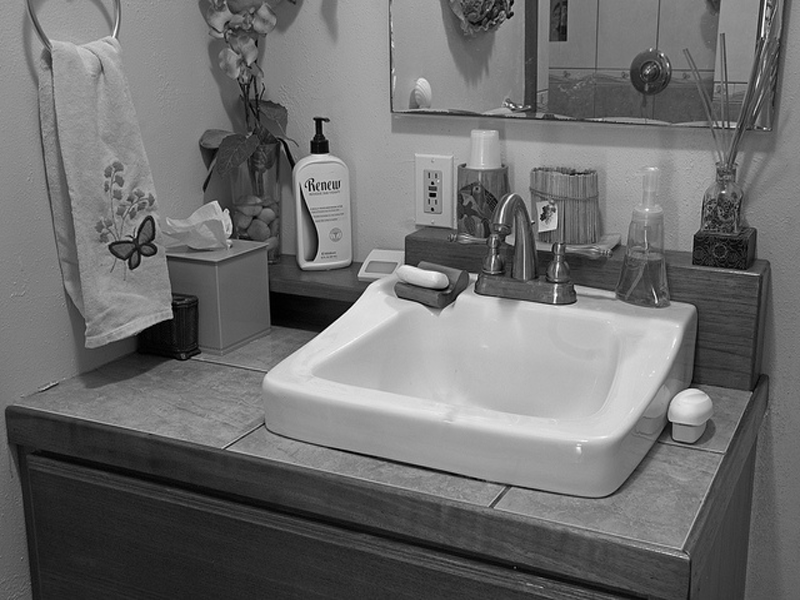}}
		&\frame{\includegraphics[width=\linewidth]{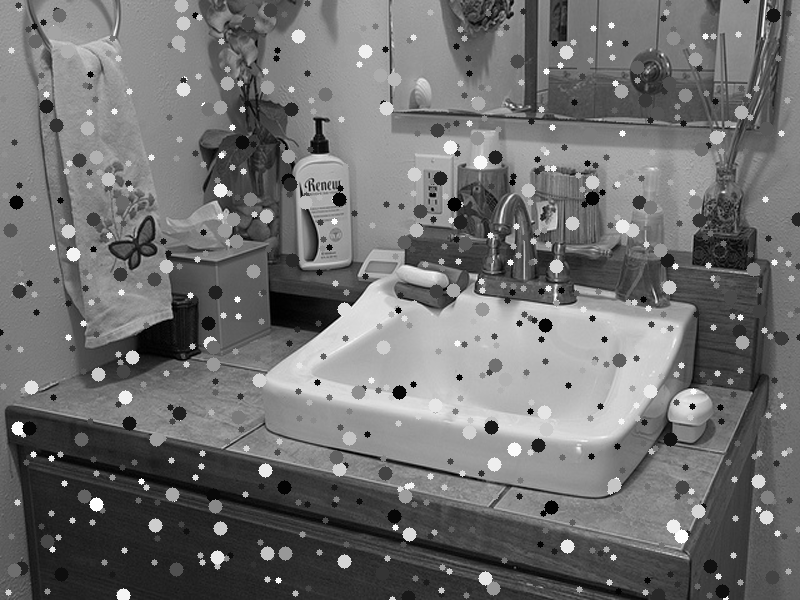}}
		&\frame{\includegraphics[width=\linewidth]{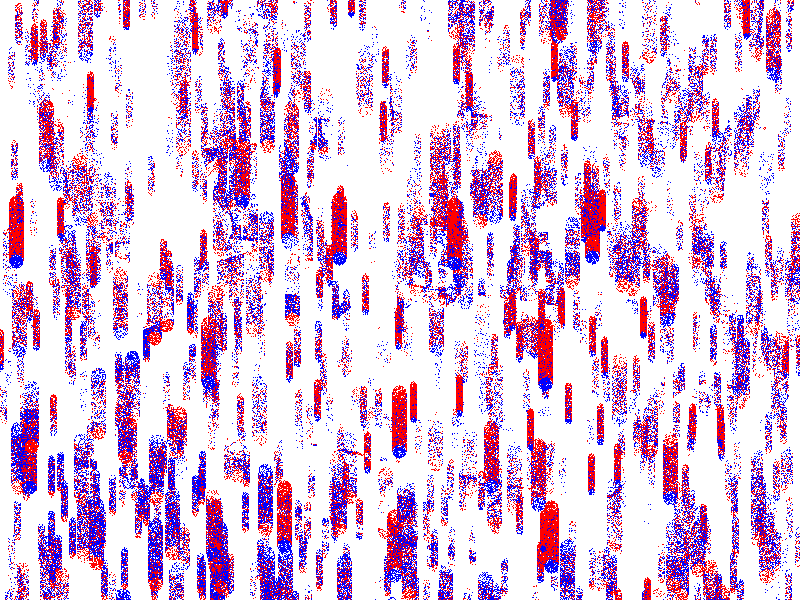}}
		\\
        \rotatebox{90}{\makecell{20\%}}
		& \frame{\includegraphics[width=\linewidth]{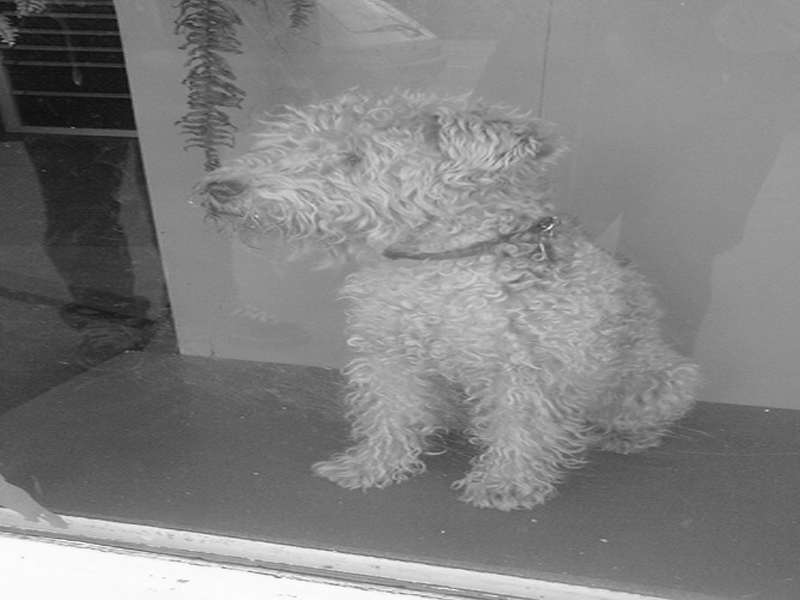}}
		&\frame{\includegraphics[width=\linewidth]{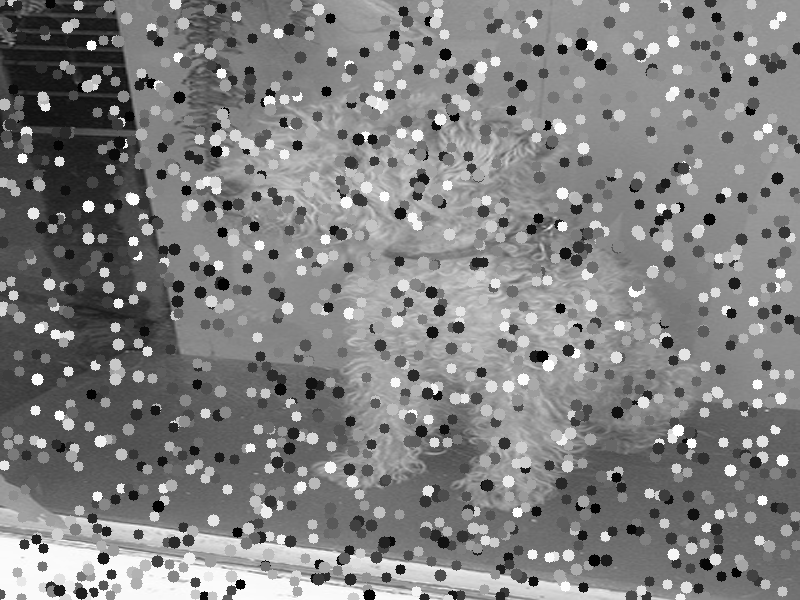}}
		&\frame{\includegraphics[width=\linewidth]{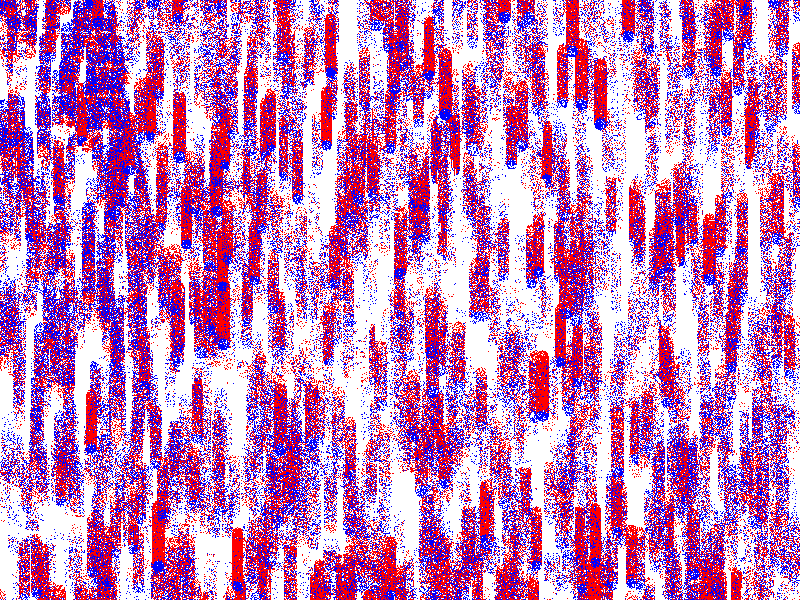}}
		\\

        \rotatebox{90}{\makecell{40\%}}
		& \frame{\includegraphics[width=\linewidth]{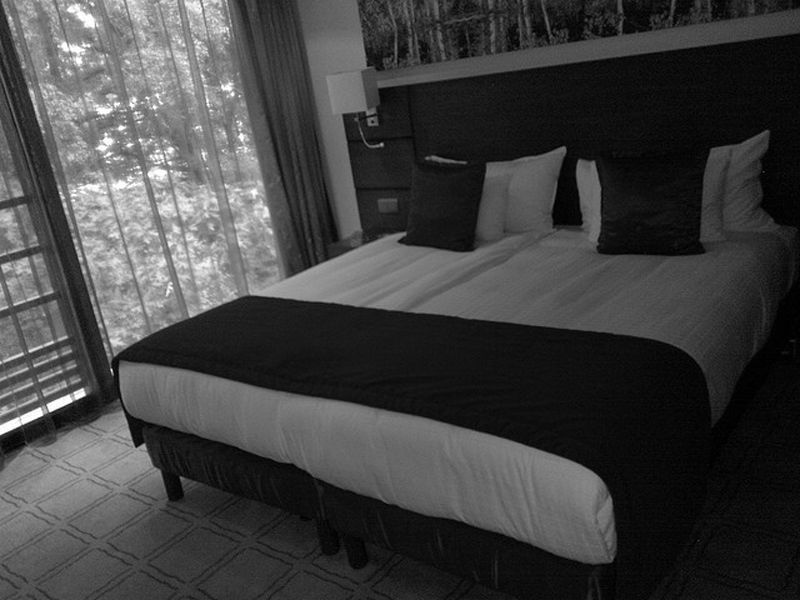}}
		&\frame{\includegraphics[width=\linewidth]{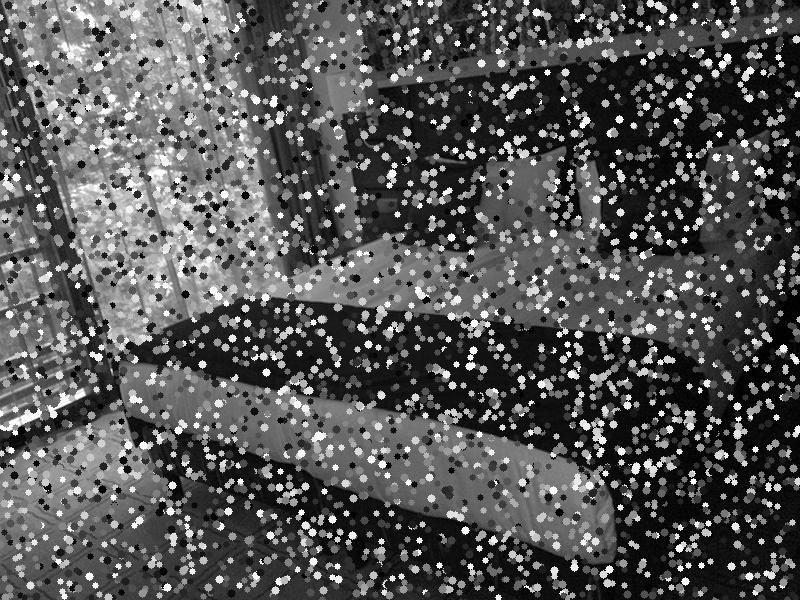}}
		&\frame{\includegraphics[width=\linewidth]{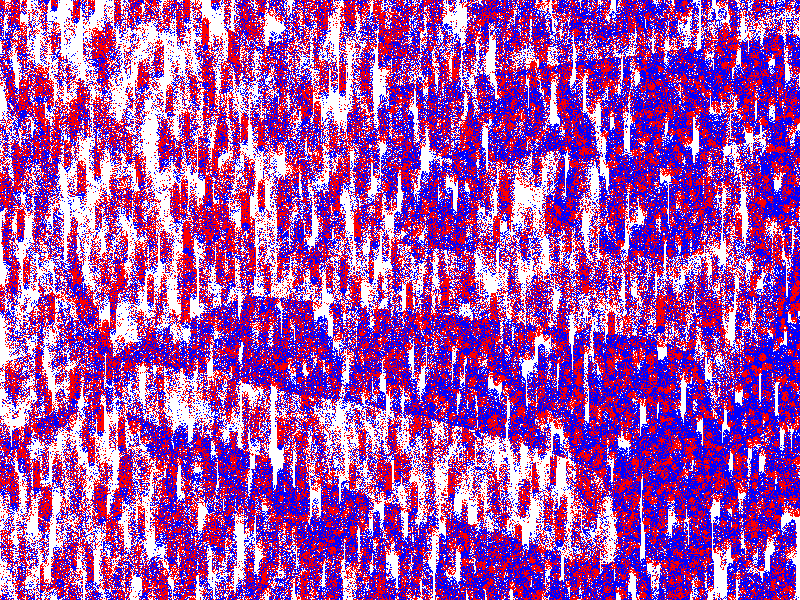}}
		\\
        
        \rotatebox{90}{\makecell{50\%}}
		& \frame{\includegraphics[width=\linewidth]{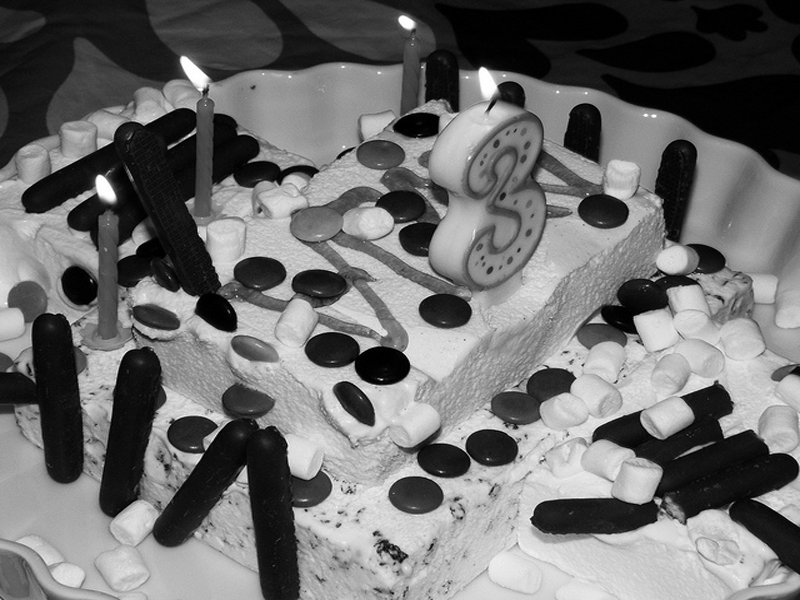}}
		&\frame{\includegraphics[width=\linewidth]{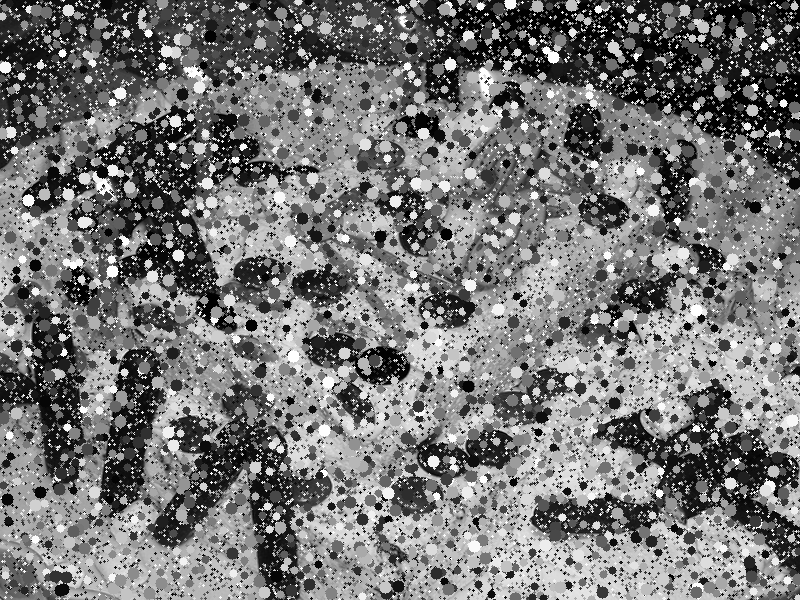}}
		&\frame{\includegraphics[width=\linewidth]{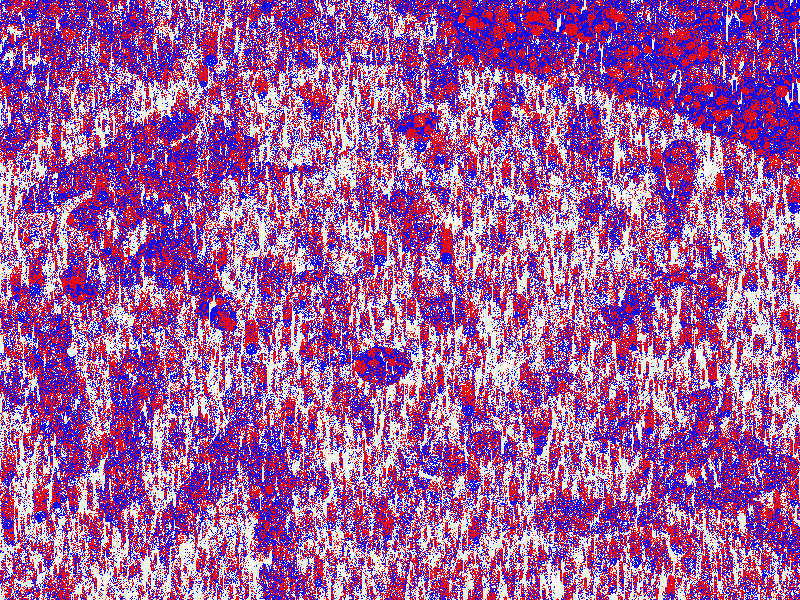}}
		\\
        
        \rotatebox{90}{\makecell{60\%}}
		& \frame{\includegraphics[width=\linewidth]{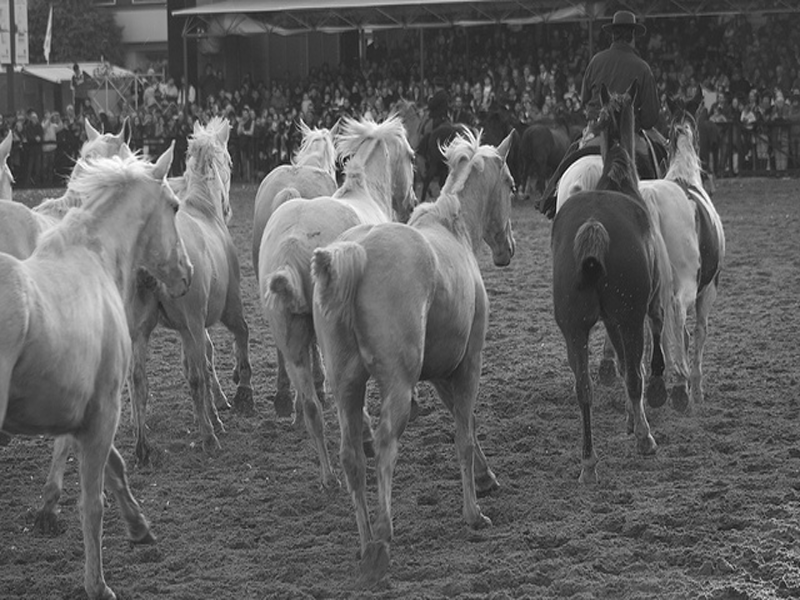}}
		&\frame{\includegraphics[width=\linewidth]{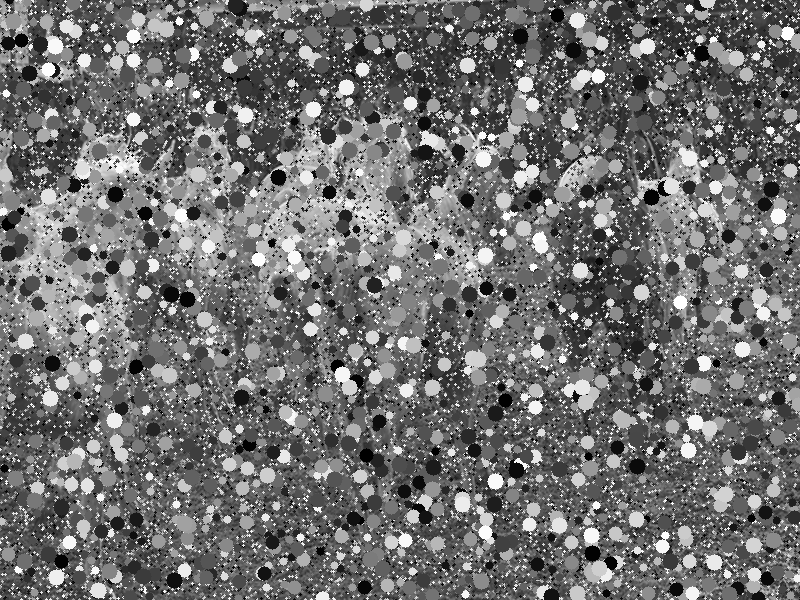}}
		&\frame{\includegraphics[width=\linewidth]{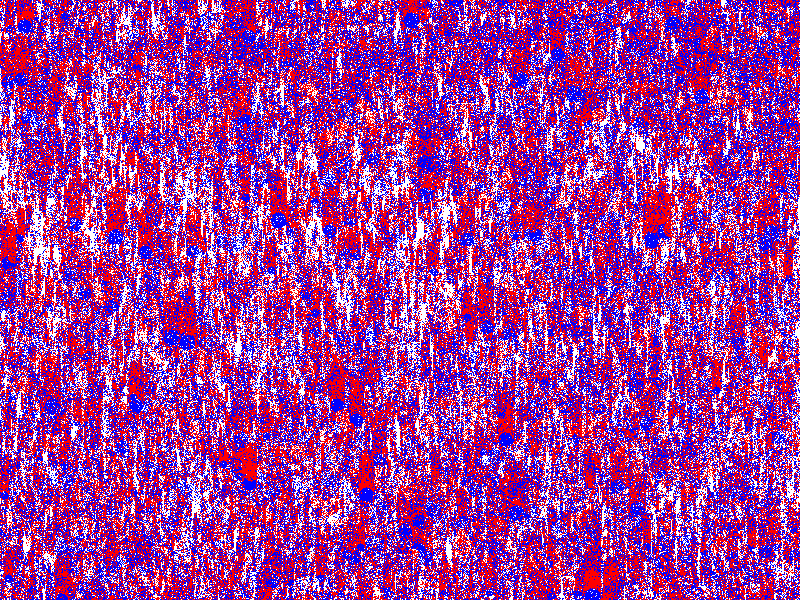}}
		\\

	\end{tabular}
    \vspace{-1ex}
	\caption{Sequence samples of different coverages from our synthetic dataset showing the occlusion-free ground truth frame (left), the occluded frame (middle), and the events accumulated into an event frame (right).}
	\label{fig:syn_data}
\end{figure}

For the synthetic dataset, we simulate $320$ training and $160$ test sequences, each containing a sequence of occluded images, synthetic events, and an occlusion-free ground truth image.
The background images are sampled from the MS-COCO dataset \cite{Lin14eccv} and cropped to a resolution of $384 \times 512$.
Each sequence consists of an image covered with circular-shaped particles which smoothly move across the image. %
To introduce more variability in the data generation process, we randomize the size, intensity, and velocity of the occlusion particles. %
This is done to simulate different occlusion densities from $10$\% to $60$\% occlusion ratio as seen in \Fig \ref{fig:syn_data}.
Since the generation of events requires high-frame-rate videos, we compute the continuous-time trajectory of the particles over the duration of the sequences and render the occluded frames at a high framerate.
Events are generated from these rendered images using ESIM \cite{Rebecq18corl}.
Finally, for each sequence, we create synchronized event sequences, the occluded image, groundtruth occlusion mask, and groundtruth background image.
These groundtruth occlusion masks are only used as an input to the image inpainting baselines.
We consider grayscale images instead of RGB images because events from the event camera only provide relative intensity changes but lack color information.
Therefore, the reconstruction of RGB images using events introduces additional challenges.

\subsection{Real-world Dataset}
We build a dynamic occlusion dataset where the event streams are captured by a Prophesee Gen4 camera \cite{Finateu20isscc}, with a resolution of $1280 \times 720$, and the images with a FLIR Blackfly S camera with a resolution of $4000 \times 3000$.
The images are captured at a framerate of $15$ fps and with an exposure time of $\SI{7}{\milli \second}$.
The cameras are hardware synchronized and mounted in a beam splitter setup (\Fig \ref{fig:setup} (a)), which contains a mirror that splits the incoming light to the event and frame camera, ensuring alignment between events and frames.
Similar to the synthetic dataset, background images are sampled from the MSCOCO dataset \cite{Lin14eccv} and projected on a TV screen, which is recorded with the beam splitter setup pointed towards the screen. 
The occlusions are created by rolling Styrofoam balls over the TV screen as shown in \Fig \ref{fig:setup}  (b).
For each sequence, the ground truth is collected by recording the image projected on the screen without any occlusions.
We sample $3$ frames from each sequence corresponding to low, medium, and high occlusion densities as shown in \Fig \ref{fig:real_data}.
Overall, the real-world dataset consists of $154$ training sequences and $79$ test sequences.

\global\long\def\figHeight{4.1cm}
\begin{figure}[t]
\centering
\subfloat[]{\label{fig:setup:camera}\includegraphics[height=\figHeight]{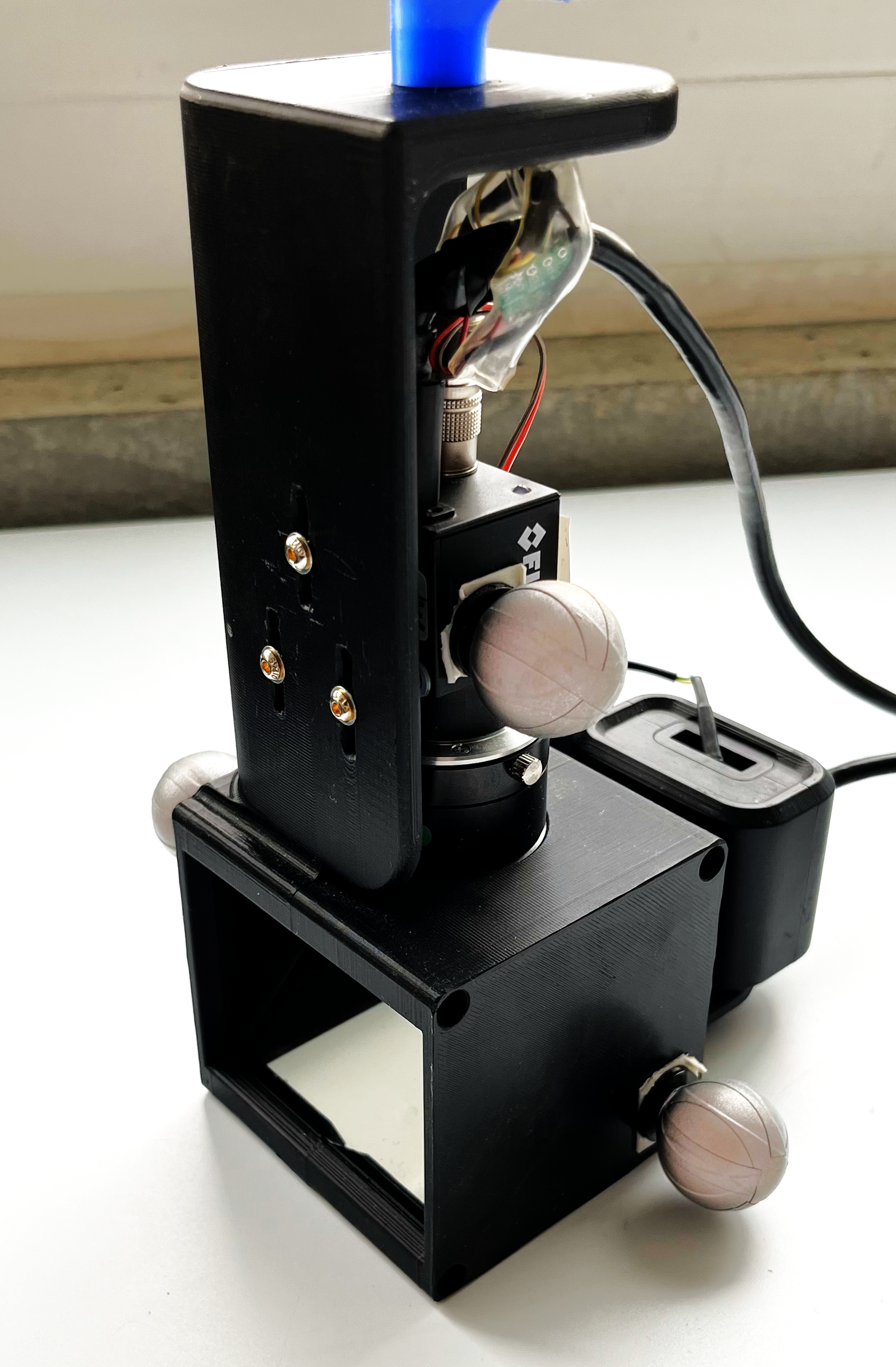}}\;
\subfloat[]{\label{fig:setup:experiment}{\includegraphics[trim={0, 5cm, 5cm, 15cm},clip,width=0.34\linewidth]{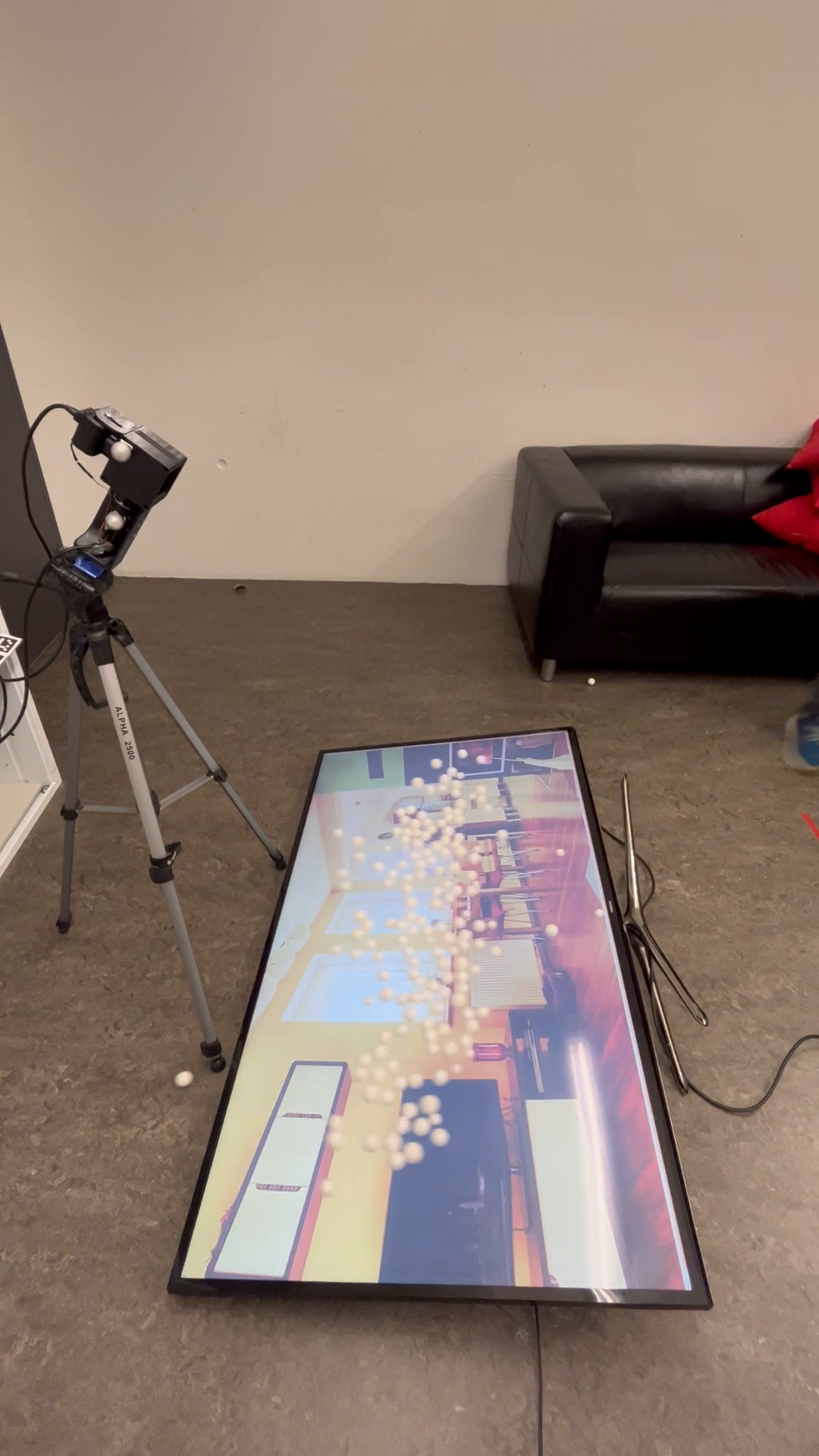}}}
\vspace{-1.5ex}
\caption{For the recording of our real-world dataset, we use a beamsplitter setup consisting of an event and a frame camera (a) to record particles moving on top of a screen (b). The experimental setup consists of the TV screen displaying a background image with the occluding particles rolling on the screen.}
\vspace{-2ex}
\label{fig:setup}
\end{figure}
\global\long\def\figWidth{0.25\linewidth}
\begin{figure}
	\centering
    \setlength{\tabcolsep}{2pt}
	\begin{tabular}{
	M{\figWidth}
	M{\figWidth}
	M{\figWidth}
	M{\figWidth}}
		Groundtruth & Low & Medium & High  
		\\

		\includegraphics[trim={10cm 5cm 10cm 6cm},clip,width=\linewidth]{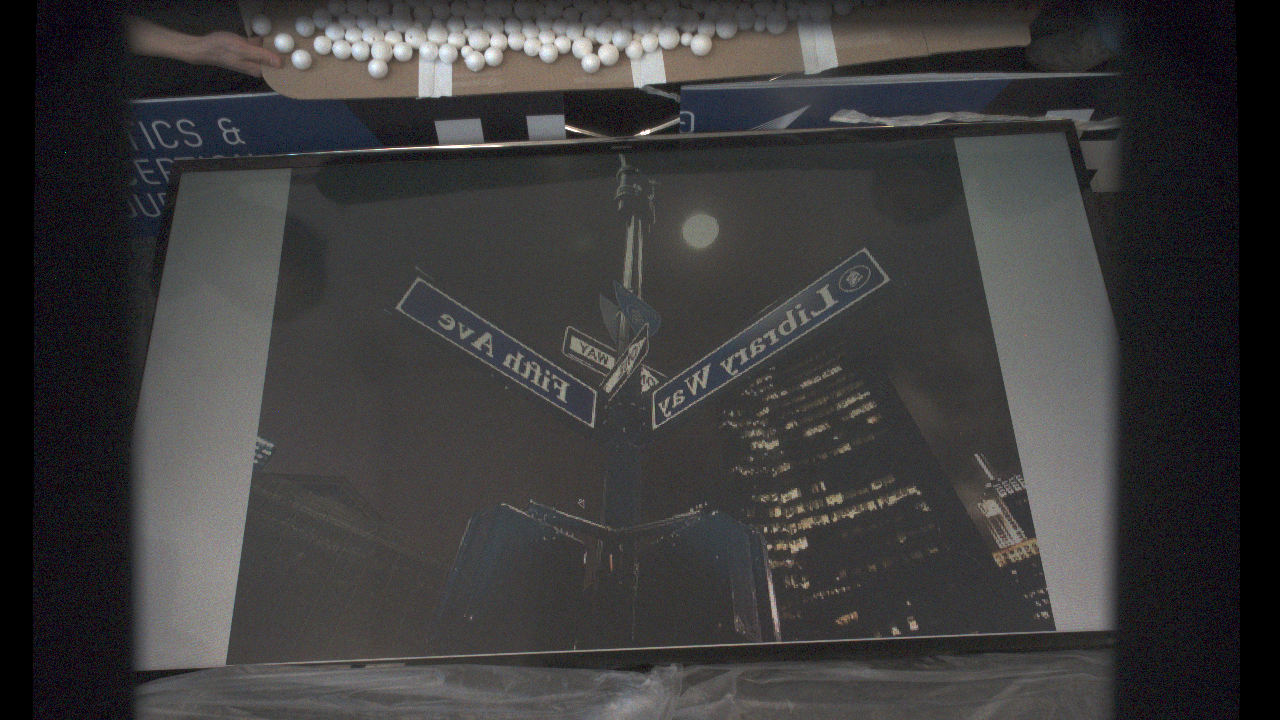}
		&\frame{\includegraphics[trim={10cm 5cm 10cm 6cm},clip,width=\linewidth]{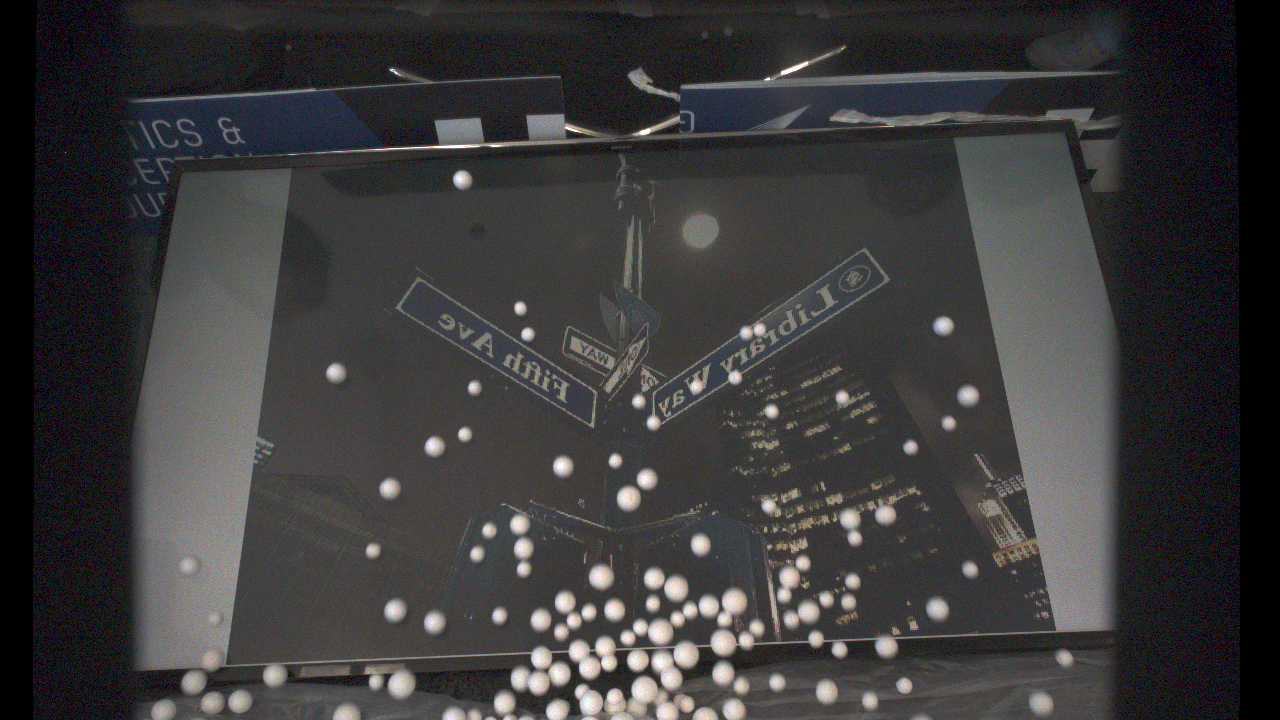}}
		&\frame{\includegraphics[trim={10cm 5cm 10cm 6cm},clip,width=\linewidth]{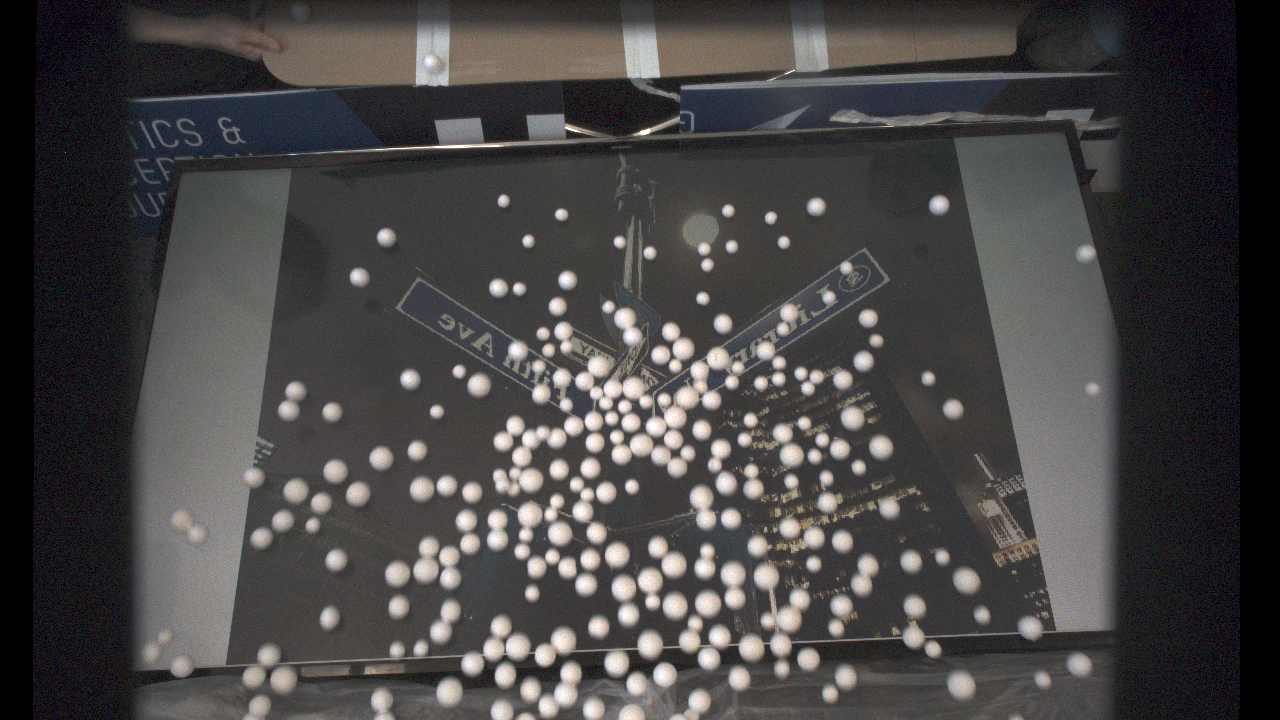}}
		&\frame{\includegraphics[trim={10cm 5cm 10cm 6cm},clip,width=\linewidth]{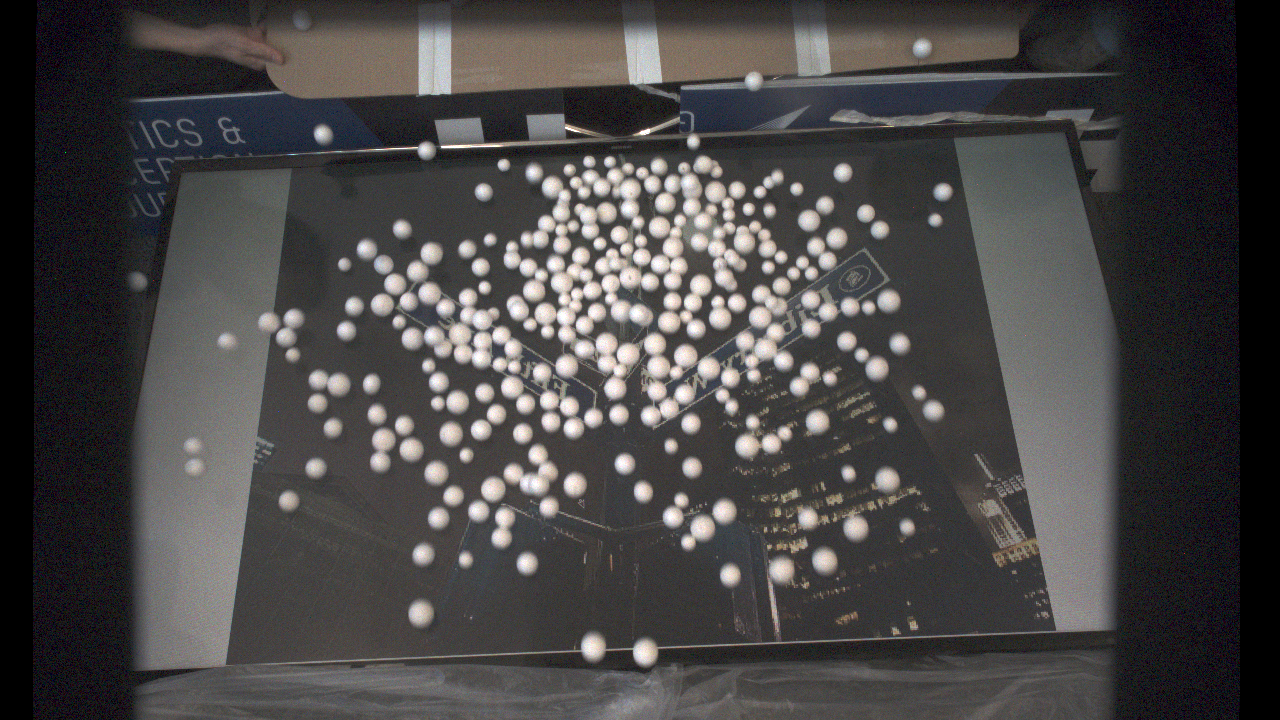}}
        \\
		\includegraphics[trim={10cm 5cm 10cm 6cm},clip,width=\linewidth]{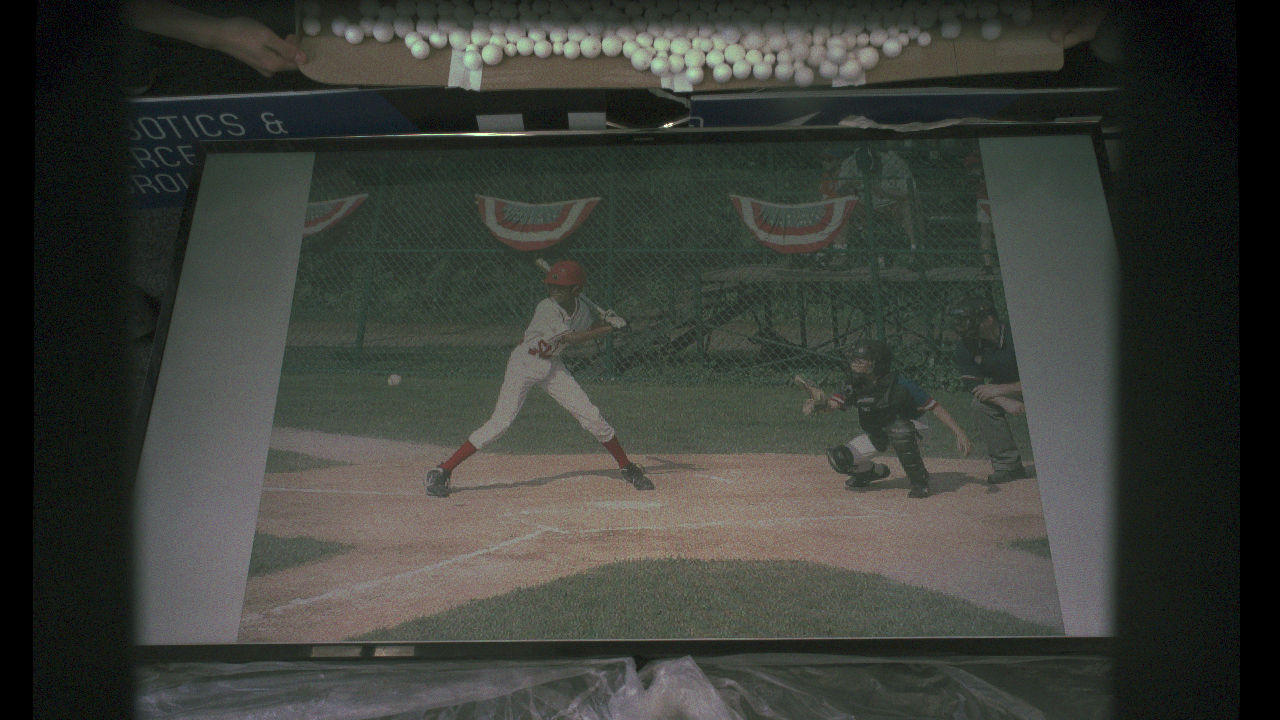}
		&\frame{\includegraphics[trim={10cm 5cm 10cm 6cm},clip,width=\linewidth]{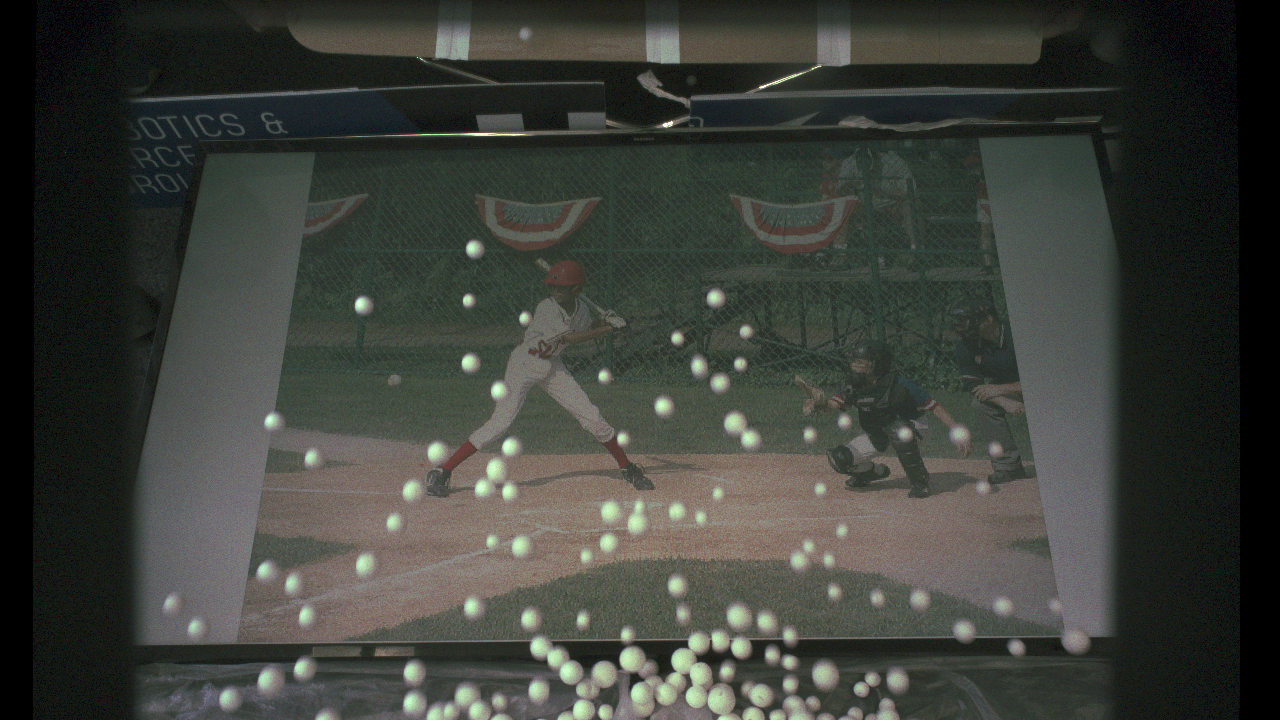}}
		&\frame{\includegraphics[trim={10cm 5cm 10cm 6cm},clip,width=\linewidth]{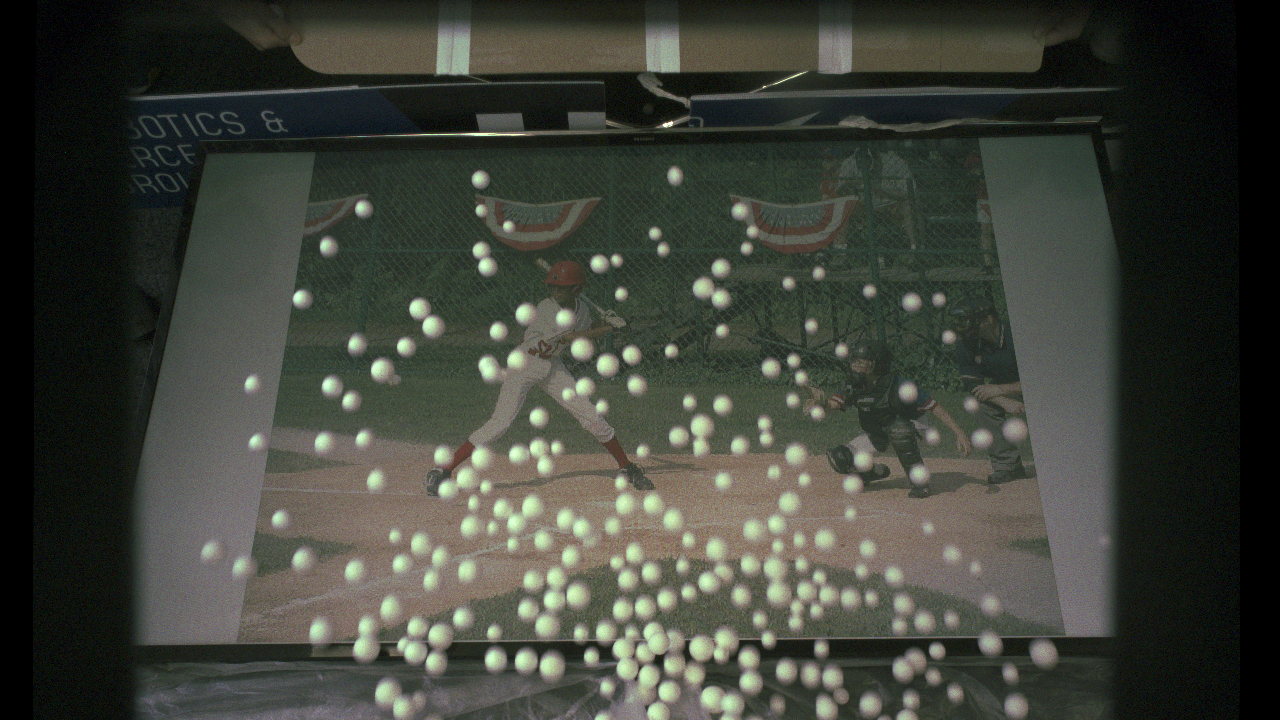}}
		&\frame{\includegraphics[trim={10cm 5cm 10cm 6cm},clip,width=\linewidth]{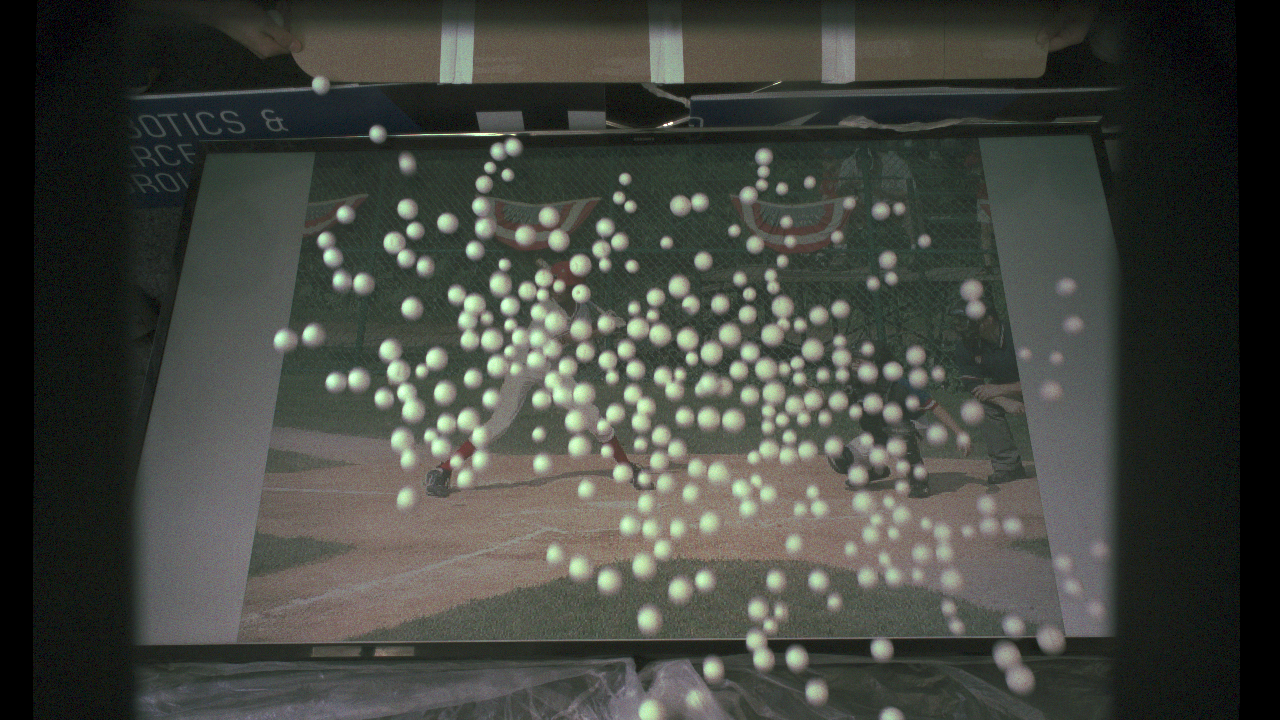}}
        \\
	\end{tabular}
    \vspace{-1ex}
	\caption{Our real-world dataset contains samples with low, medium, and high occlusion density.}
	\label{fig:real_data}
\end{figure}

\section{Experiments}
\paragraph{Implementation details} Our models are implemented in Pytorch~\cite{Paszke17nipsw} and trained from scratch with random weights on our synthetic dataset. 
For the experiments on our real dataset, we finetune our pre-trained network on the real training data. 
We use Adam optimizer\cite{Kingma15iclr} with a learning rate of $1\mathrm{e}{-3}$ and batch size of 4 and train for $1250$ epochs on the synthetic and real dataset.
We supervise our network with the $L1$ loss computed on the predicted and ground truth occlusion-free images.
To measure the quality of the reconstructed image, we use structural similarity index measure (SSIM) \cite{Wang04tip}, peak signal-to-noise ratio (PSNR), and mean absolute error (MAE).

\paragraph{Baselines}
We evaluate our approach against four state-of-the-art image-inpainting solutions: MAT \cite{li_mat22cvpr}, MISF \cite{li_misf22cvpr}, PUT \cite{liu22cvpr}, and ZITS \cite{dong22cvpr}.
These baselines are trained on challenging large-scale datasets with over eight million images. 
Thus, for our experiments, we use the weights provided by the authors. %
Since dynamic occlusion removal from a single viewpoint using an event camera is a novel task, there currently do not exist any event-based approaches that tackle this problem directly.
The closest baseline uses events and frames to reconstruct the background image from multiple viewpoints (EF-SAI) \cite{liao22cvpr}.
EF-SAI \cite{liao22cvpr} uses a refocus module with events to blur the foreground and focus on the target depth plane of the background. 
We adapt this baseline for our task by providing it with a single image and events from only a single viewpoint and train this network on our dataset from scratch.
Additionally, we also compare against events-to-image reconstruction methods \cite{Rebecq19cvpr}.
This method is adapted by combining the reconstructed event images with the occluded images using the groundtruth mask.
Note that all of the evaluated baselines were originally not designed for this particular task but are the closest related methods applicable to our task. %

\subsection{Results on Synthetic Dataset}
To compare our method against the baselines in controlled conditions, we evaluate all methods on our synthetic dataset.
\Tab \ref{tab:syb_main} summarizes the quantitative results.
Our method outperforms the best image inpainting baseline by \SI{3}{dB} in terms of PSNR.
\Fig \ref{fig:syn_qual} shows qualitative comparisons between different methods.
Image inpainting methods tend to hallucinate the background, resulting in visually more pleasing information rather than the true scene.
An example of this can be seen in \Fig \ref{fig:syn_qual} third row, where the inpainting method is unable to reconstruct the characters on the bus, whereas our method is able to better preserve this information.
We also outperform the event-based synthetic aperture imaging baseline EF-SAI \cite{liao22cvpr} and the event-to-image reconstruction baseline E2VID \cite{Rebecq19cvpr}.
Although the input to our method and EF-SAI consists of events and a single image, the EF-SAI baseline was designed for multi-view reconstruction.
The simple event accumulation baseline is one of the lowest performing baselines even with the knowledge of the correct contrast threshold, as the occlusions are too complex for the basic event generation model to capture the intensity changes as discussed in the \Sec \ref{sec:method:basic}.
We also analyze the effect of occlusion density on the performance of all the methods and summarize them in \Tab \ref{tab:syn_coverage}.
As expected, increasing the occlusion density decreases the performance of all the approaches.
However, at higher occlusion densities, the image inpainting methods drastically degrade in performance as they tend to hallucinate occluded areas.
In contrast, our method uses events that provide continuous intensity changes, which results in a better performance at higher occlusion densities.
We provide more qualitative results of other baselines in the supplementary material.

\begin{table}[!t]
    \centering
    \begin{adjustbox}{max width=\linewidth}
    \setlength{\tabcolsep}{4pt}
    {\small
    \begin{tabular}{lcccc}
        \toprule
         Method & Input  & PSNR $\uparrow$  &  SSIM $\uparrow$ & MAE $\downarrow$ \\
        \midrule
        MAT \cite{li_mat22cvpr} & I & 30.7620 & 0.9217 & 0.0107 \\
        MISF \cite{li_misf22cvpr}& I & 31.1884 & 0.9229 & 0.0101  \\
        PUT \cite{liu22cvpr} & I & 26.9858 & 0.8608 & 0.0187  \\
        ZITS \cite{dong22cvpr}& I & 31.2971 & 0.9328 & 0.0100  \\
        EF-SAI \cite{liao22cvpr}& I+E & 26.7557 & 0.8688 & 0.0257  \\
        E2VID \cite{Rebecq19cvpr}& E & 19.2278 & 0.6086 & 0.0508  \\
        Ours (Acc. Method)& E & 20.3444 & 0.6955 & 0.0373  \\
        Ours (Learning) & I+E & \textbf{34.6203} & \textbf{0.9536} & \textbf{0.0085}  \\
        \bottomrule
    \end{tabular}}
    \end{adjustbox}
    \caption{Reconstruction performance on our synthetic dataset. 'I' and 'E' stand for image and events, respectively.}
    \label{tab:syb_main}
\end{table}

\global\long\def\figWidth{0.2\linewidth}
\begin{figure*}
	\centering
    \setlength{\tabcolsep}{1pt}
	\begin{tabular}{
	M{\figWidth}
	M{\figWidth}
	M{\figWidth}
	M{\figWidth}
	M{\figWidth}}
		Occluded & EF-SAI \cite{liao22cvpr} & ZITS \cite{dong22cvpr} & Ours & Groundtruth
		\\
		\includegraphics[width=\linewidth]{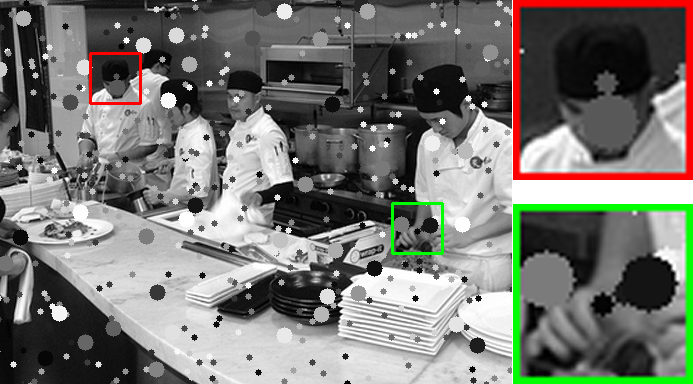}
		&\frame{\includegraphics[clip,width=\linewidth]{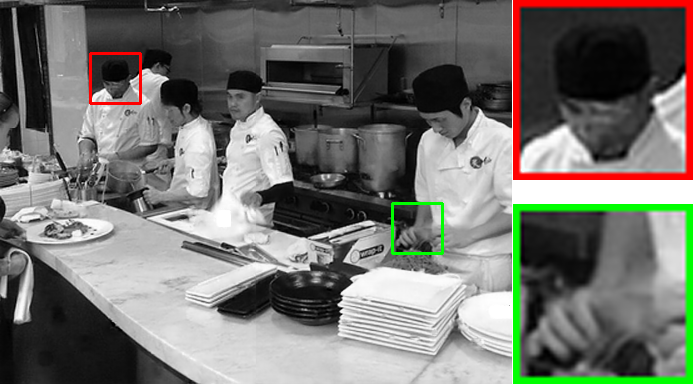}}
		&\frame{\includegraphics[width=\linewidth]{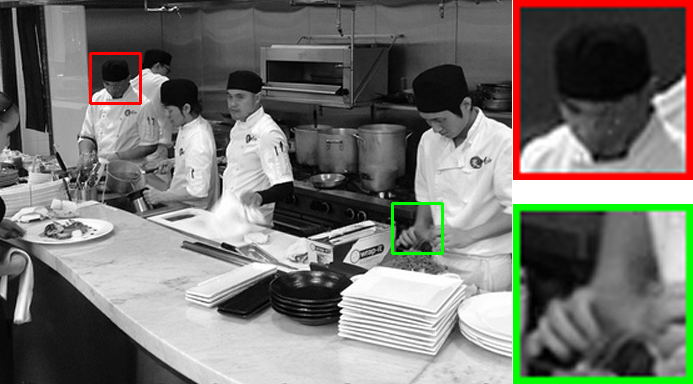}}
		&\frame{\includegraphics[width=\linewidth]{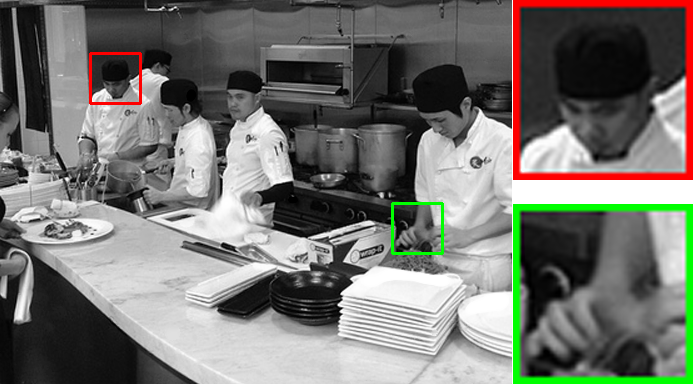}}
		&\frame{\includegraphics[width=\linewidth]{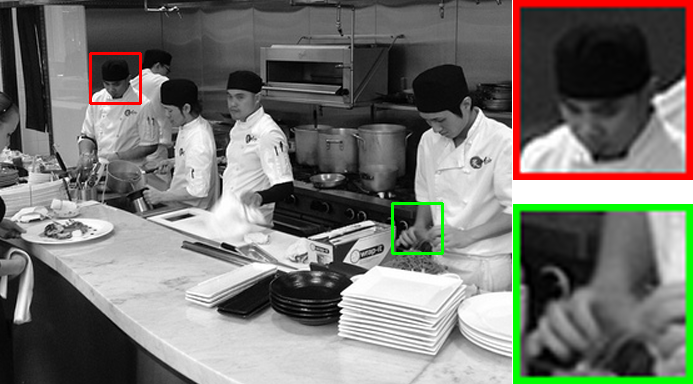}}
        \\
        \includegraphics[width=\linewidth]{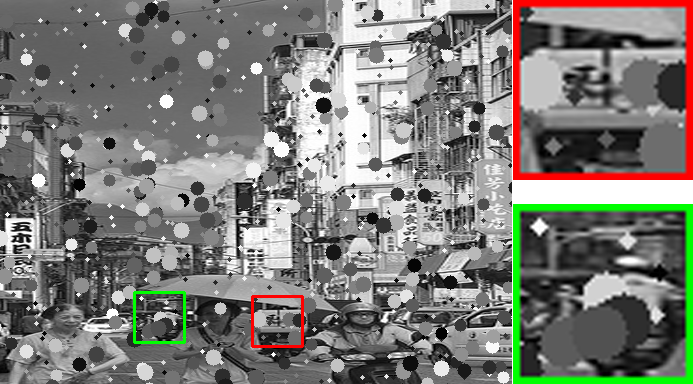}
		&\frame{\includegraphics[clip,width=\linewidth]{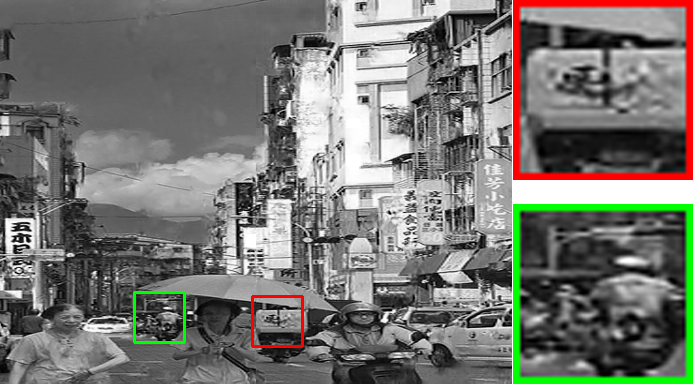}}
		&\frame{\includegraphics[width=\linewidth]{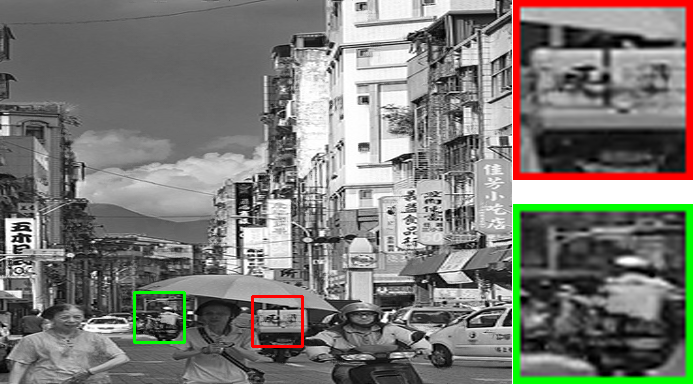}}
		&\frame{\includegraphics[width=\linewidth]{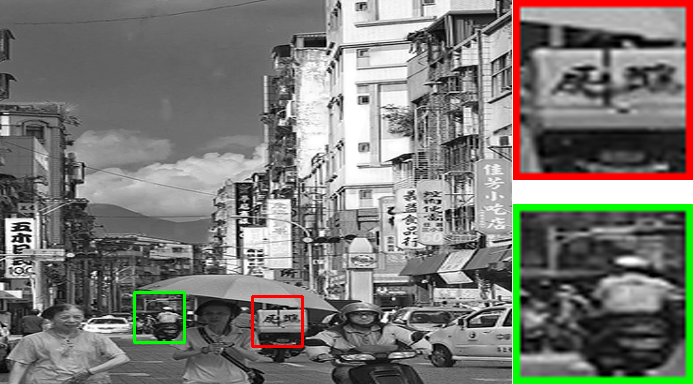}}
		&\frame{\includegraphics[width=\linewidth]{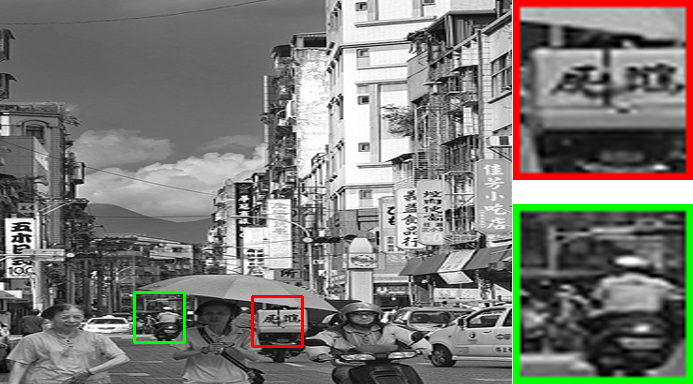}}
        \\
        \includegraphics[width=\linewidth]{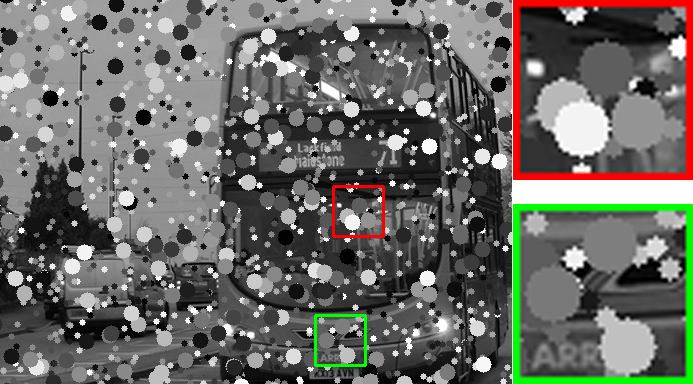}
		&\frame{\includegraphics[clip,width=\linewidth]{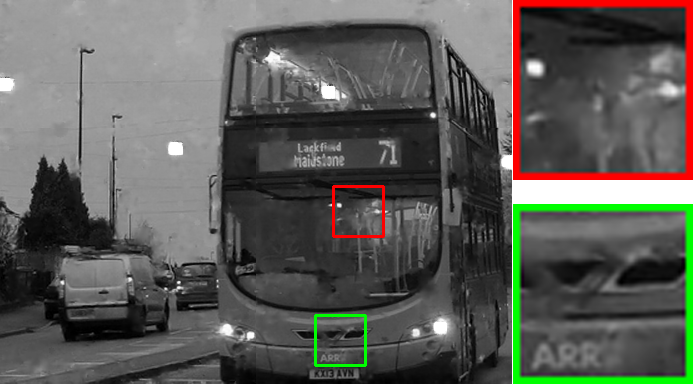}}
		&\frame{\includegraphics[width=\linewidth]{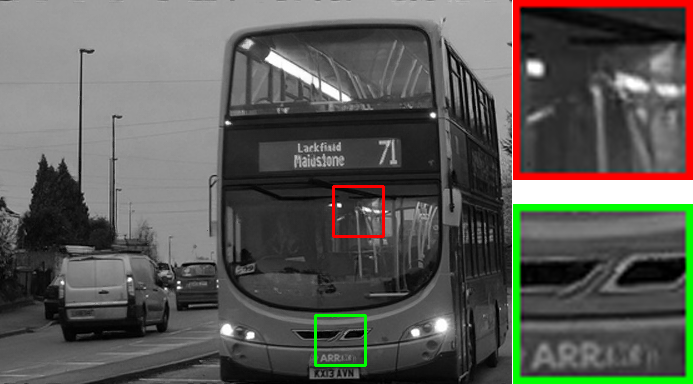}}
		&\frame{\includegraphics[width=\linewidth]{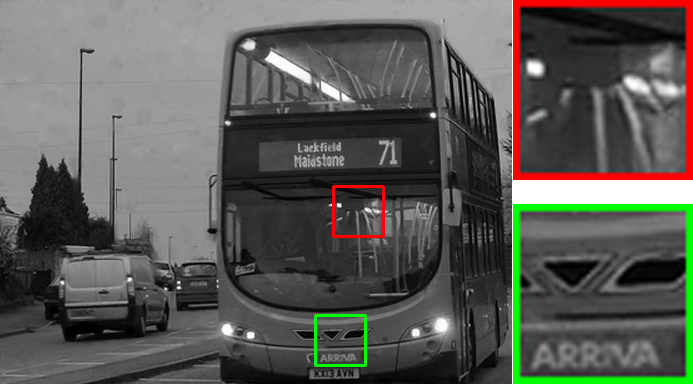}}
		&\frame{\includegraphics[width=\linewidth]{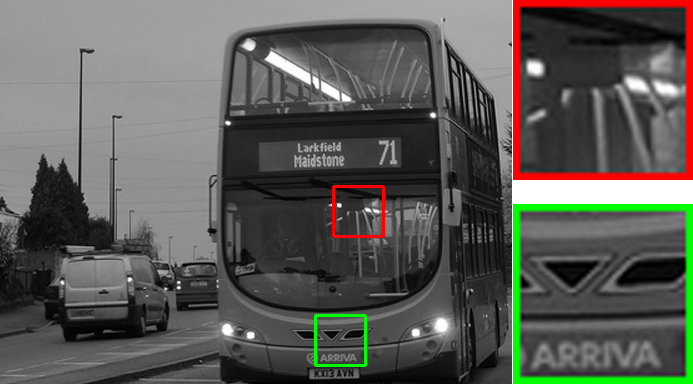}}
        \\
        \includegraphics[width=\linewidth]{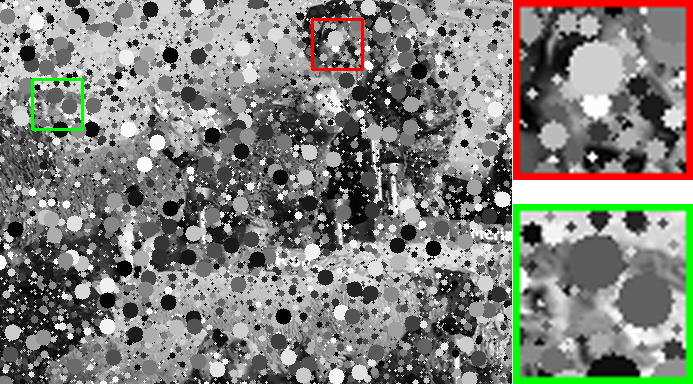}
		&\frame{\includegraphics[clip,width=\linewidth]{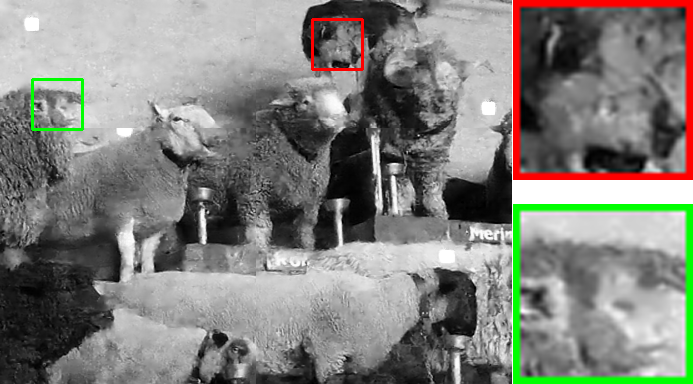}}
		&\frame{\includegraphics[width=\linewidth]{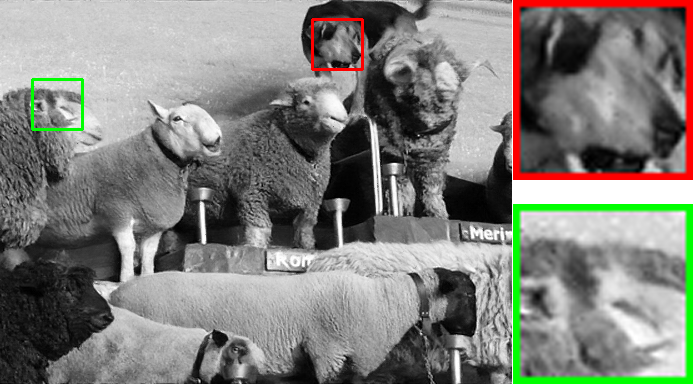}}
		&\frame{\includegraphics[width=\linewidth]{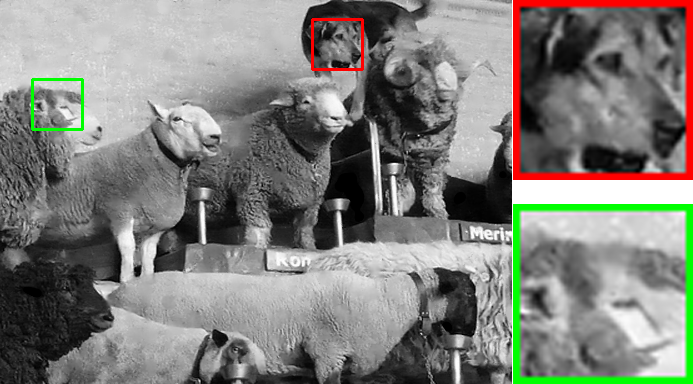}}
		&\frame{\includegraphics[width=\linewidth]{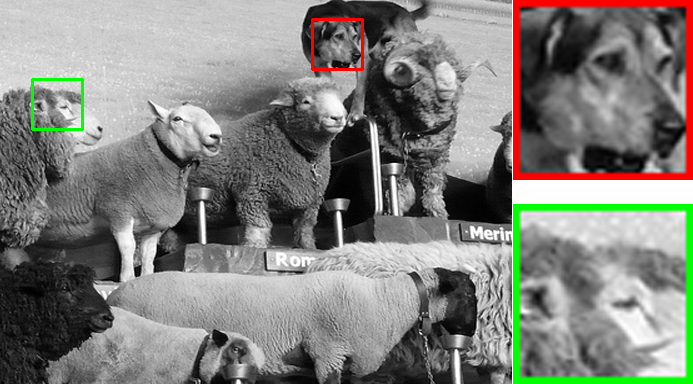}}
        \\
	\end{tabular}
    \vspace{-1ex}
     \caption{Images showing the occluded input frame, the reconstructed frame of the best-performing frame-based and event-based baselines and our method, and the ground truth frame for our synthetic dataset.}
	\label{fig:syn_qual}
\end{figure*}

\begin{table}[!t]
    \centering
    \begin{adjustbox}{max width=\linewidth}
    \setlength{\tabcolsep}{4pt}
    {\small
    \begin{tabular}{lccccccc}
        \toprule
        Method & Input  & \multicolumn{6}{c}{PSNR $\uparrow$} \\
        Coverage &   & 10\% &  20\% &  30\%&  40\%&  50\%&  60\%\\
        \midrule
        MAT \cite{li_mat22cvpr} & I & 35.4937 & 31.4344 & 31.895 & 29.8665 & 27.7057 & 28.1765\\
        MISF \cite{li_misf22cvpr}& I & 35.6868 & 31.8720 & 31.7689 & 30.5100 &  28.3776 & 28.9148 \\
        PUT \cite{liu22cvpr} & I & 32.7979 & 28.6378 & 27.0988 & 26.1156 & 23.7555 & 23.5093 \\
        ZITS \cite{dong22cvpr}& I & 35.6619 &  31.9159 & 32.3340 &  30.4923 & 28.3828 & 28.9959 \\
        EF-SAI \cite{liao22cvpr}& I+E & 29.2994 & 27.9049 & 27.7354 & 26.2447 & 24.7267 & 24.6231\\
        E2VID \cite{Rebecq19cvpr}& E & 22.0048 & 19.8868 & 20.2063 & 18.1452 & 17.5353 & 17.5884 \\
        Ours (Acc. Method)& E & 28.0390 & 22.4906 & 20.6897 & 17.9743 & 17.0200 & 15.8527\\
        Ours (Learning) & I+E & \textbf{40.1286} &\textbf{ 36.3622} & \textbf{35.8476} & \textbf{33.2527} & \textbf{31.0083} & \textbf{31.1224}\\
        \bottomrule
    \end{tabular}}
    \end{adjustbox}
    \caption{Reconstruction performance  on our synthetic dataset in terms of PSNR divided according to the occlusion density of the test samples.}
    \label{tab:syn_coverage}
\end{table}

\paragraph{Ablation Study}
To study the effect of the EAM and OFF modules in our network, we report the network performance by removing the specific modules.
For computational reasons, we performed the ablation of the OFF module without including the EAM module, which reduces significantly the training time.
As shown in \Tab \ref{tab:syn_ablation}, adding the OFF module to the network improves the performance with respect to all metrics. 
This improvement can be explained by the ability of the OFF module to better localize the occlusions and adaptively fuse image and event features throughout the network scales.
Including the EAM module leads to an even higher performance increase of 1.9 dB in PSNR.
One possible reason for the improvement over the network without recurrent event encoding is that the EAM module is better at handling multiple overlapping occlusions by selecting events relevant to the true background and ignoring the redundant event information.
For more ablation studies showing the effect of the event and image features on the textured and uniform image regions, we refer to the supplementary material.
\begin{table}[!t]
    \centering
    \begin{adjustbox}{max width=\linewidth}
    \setlength{\tabcolsep}{4pt}
    {\small
    \begin{tabular}{lcccc}
        \toprule
         Method & Input  & PSNR $\uparrow$  &  SSIM $\uparrow$ & MAE $\downarrow$\\
        \midrule
        Ours (w/o EAM \& OFF) & I+E & 31.0776 &  0.9212 & 0.0139\\
        Ours (w/o EAM) & I+E & 32.7652 & 0.9425 & 0.0102\\
        Ours (Full) & I+E & \textbf{34.6203} & \textbf{0.9536} & \textbf{0.0085}\\
        \bottomrule
    \end{tabular}}
    \end{adjustbox}
    \caption{The reconstruction performance on the synthetic dataset of our network obtained by removing adaptively the introduced modules.}
    \label{tab:syn_ablation}
\end{table}

\subsection{Results on Real Dataset}
We also evaluate our method on our real dataset collected with a custom build beamsplitter setup.
Unlike the synthetic dataset, we do not have an occlusion mask available for the sequences.
We, therefore, approximate the mask by subtracting the occluded frame from the groundtruth frame and applying thresholding.
This mask is used as an input for the image inpainting methods and event image reconstruction baselines.
The quantitative results are shown in \Tab \ref{tab:real_gray}.
Our results on the real dataset follow a similar trend to the synthetic dataset.
In \Fig \ref{fig:real_qual}, we show the qualitative results for different sequences.
While the image inpainting method results in visually clean images, the hallucination artifacts can be clearly seen in the small image patches, e.g., the fingers around the camera are missing in the sample visualized in the first row or the watch in the sample shown in the second row. 
Our method, on the other hand, can reconstruct the background information accurately.

\global\long\def\figWidth{0.2\linewidth}
\begin{figure*}
	\centering
    \setlength{\tabcolsep}{1pt}
	\begin{tabular}{
	M{\figWidth}
	M{\figWidth}
	M{\figWidth}
	M{\figWidth}
	M{\figWidth}}
		Occluded & EF-SAI \cite{liao22cvpr} & ZITS \cite{dong22cvpr} & Ours & Groundtruth
		\\
		\includegraphics[width=\linewidth]{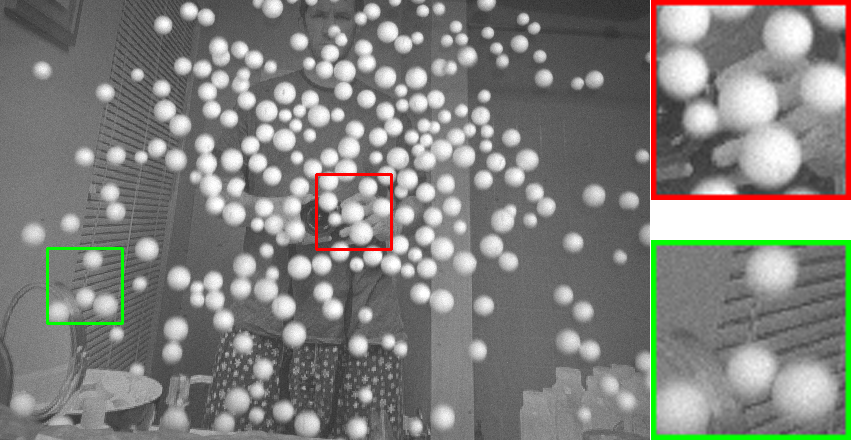}
		&\frame{\includegraphics[width=\linewidth]{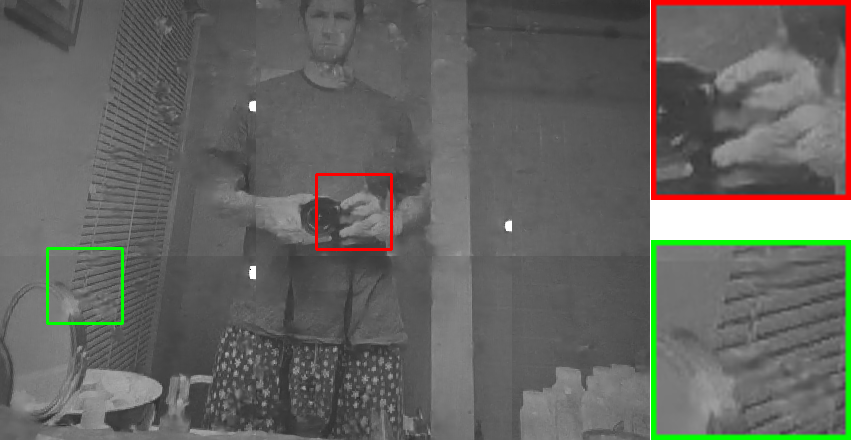}}
		&\frame{\includegraphics[width=\linewidth]{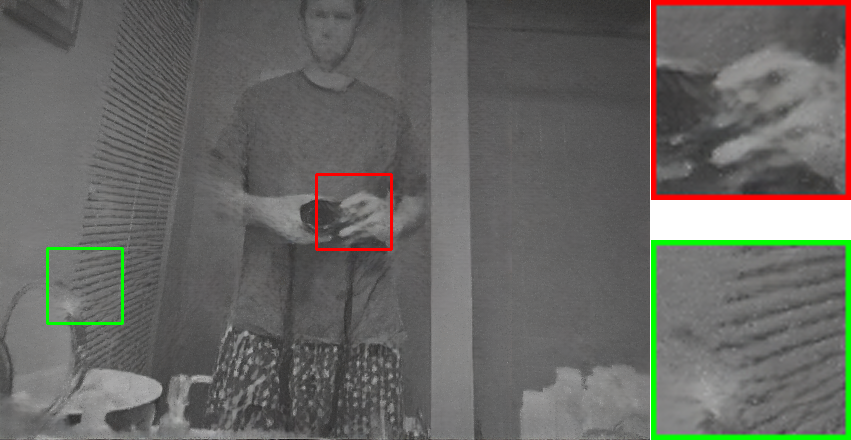}}
		&\frame{\includegraphics[width=\linewidth]{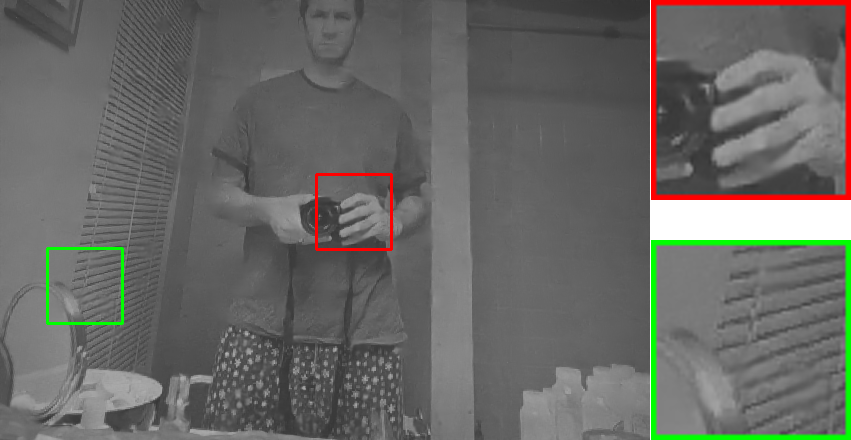}}
		&\frame{\includegraphics[width=\linewidth]{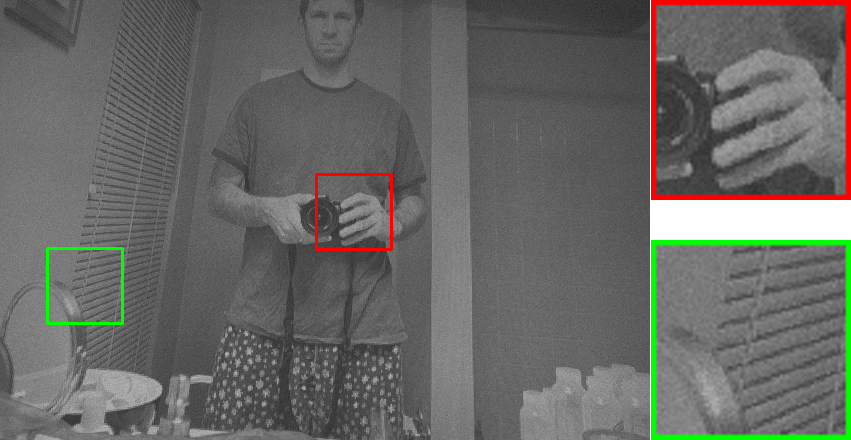}}
      	\\
		\includegraphics[width=\linewidth]{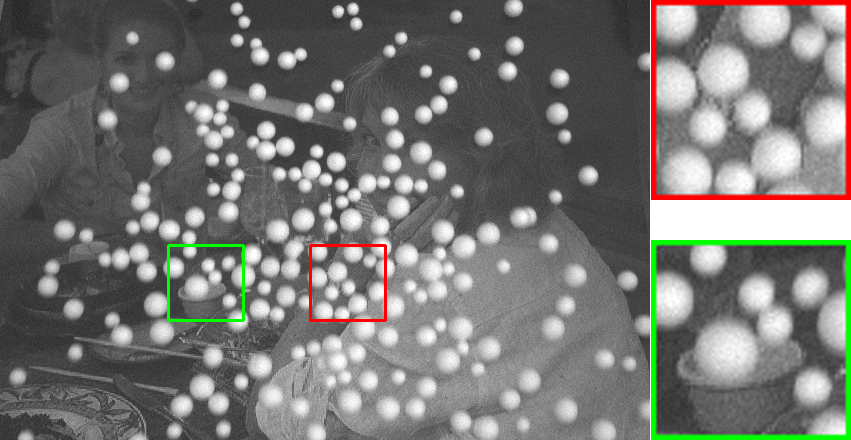}
		&\frame{\includegraphics[width=\linewidth]{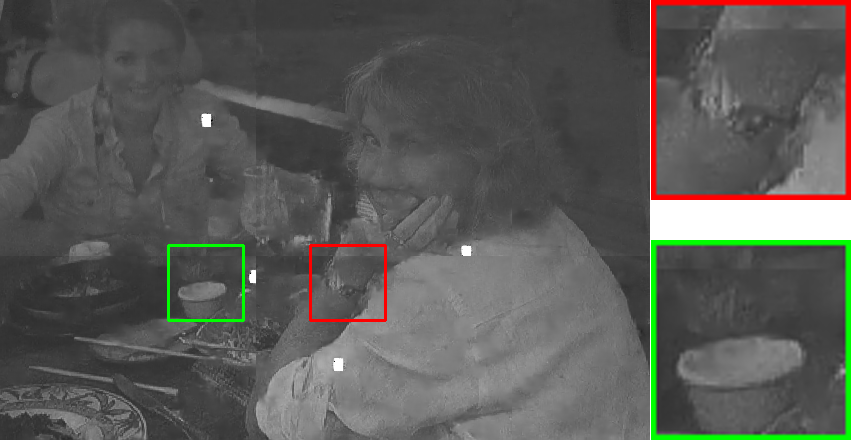}}
		&\frame{\includegraphics[width=\linewidth]{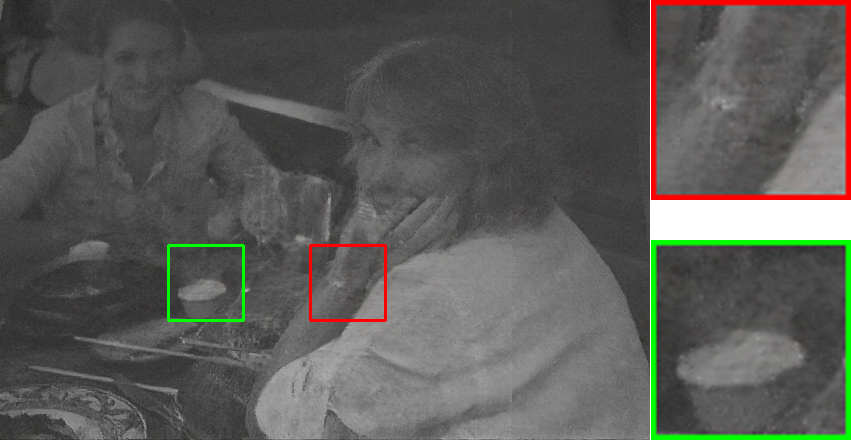}}
		&\frame{\includegraphics[width=\linewidth]{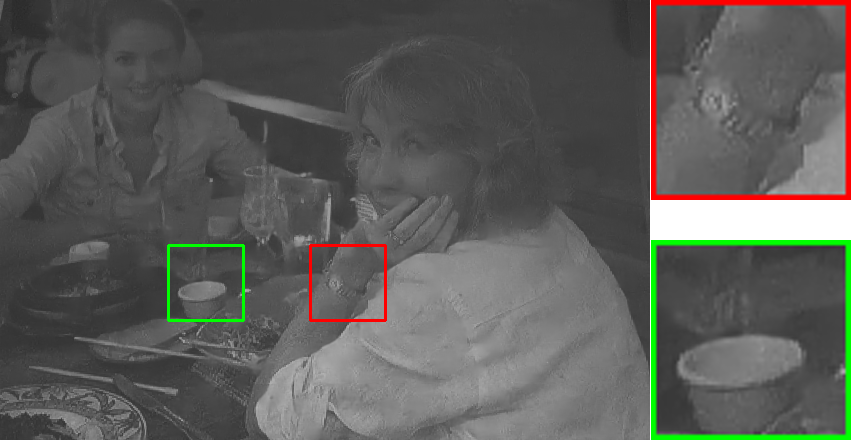}}
		&\frame{\includegraphics[width=\linewidth]{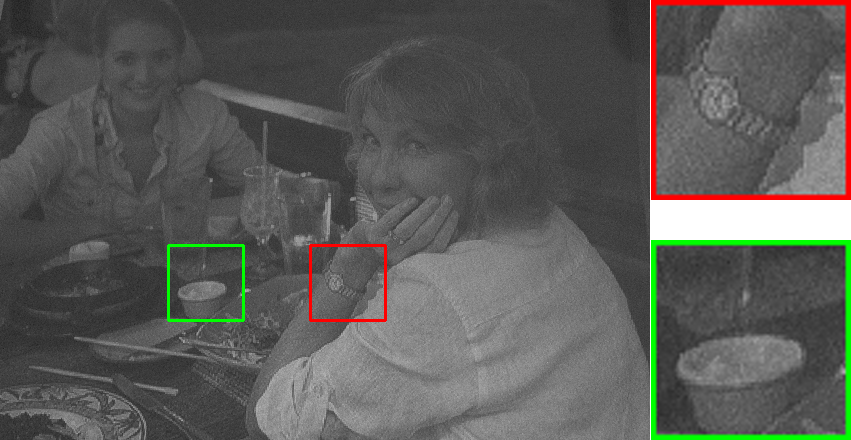}}
        \\
		\includegraphics[width=\linewidth]{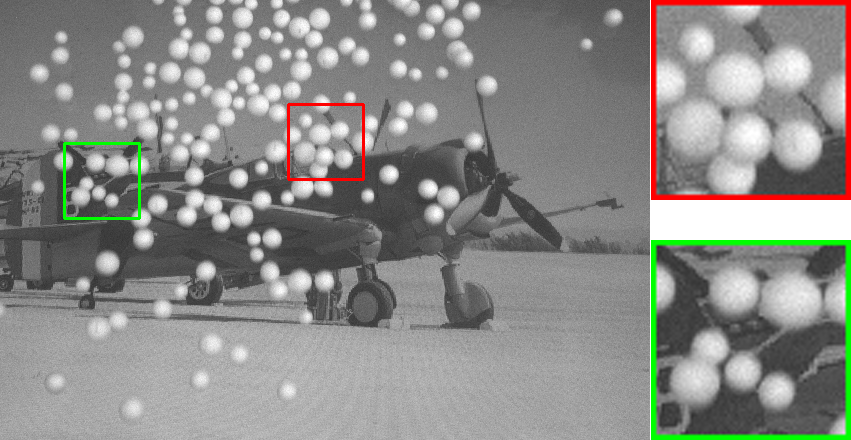}
		&\frame{\includegraphics[width=\linewidth]{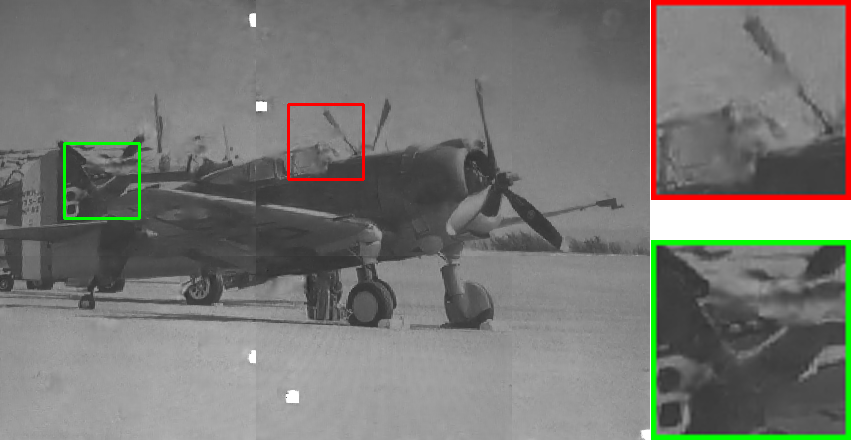}}
		&\frame{\includegraphics[width=\linewidth]{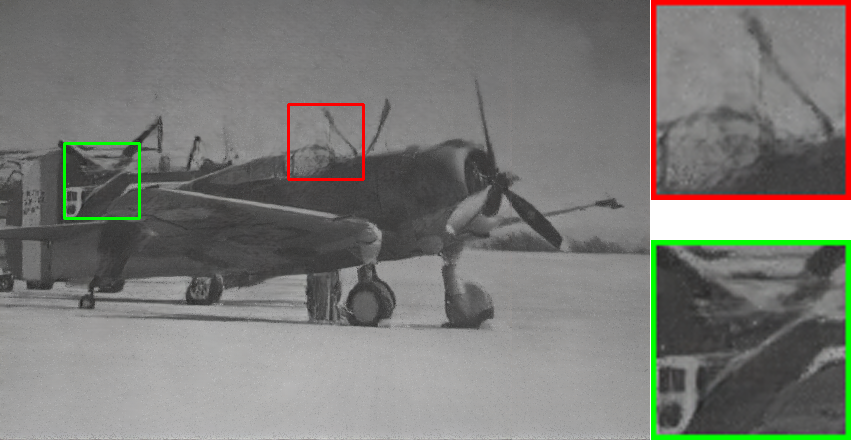}}
		&\frame{\includegraphics[width=\linewidth]{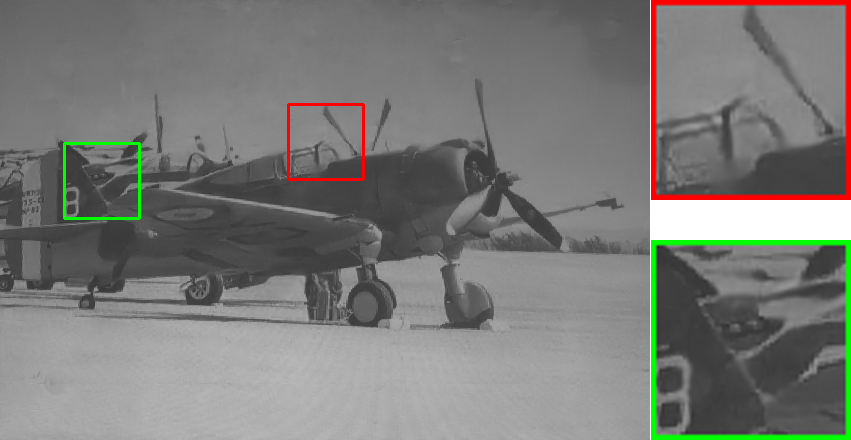}}
		&\frame{\includegraphics[width=\linewidth]{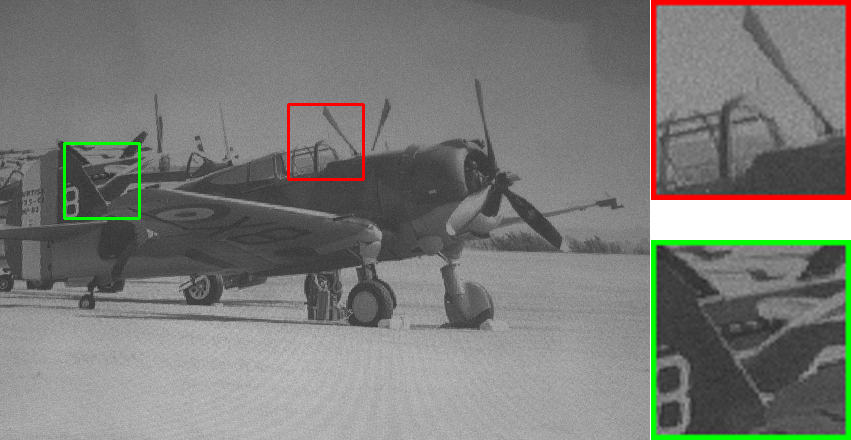}}
        \\
		\includegraphics[width=\linewidth]{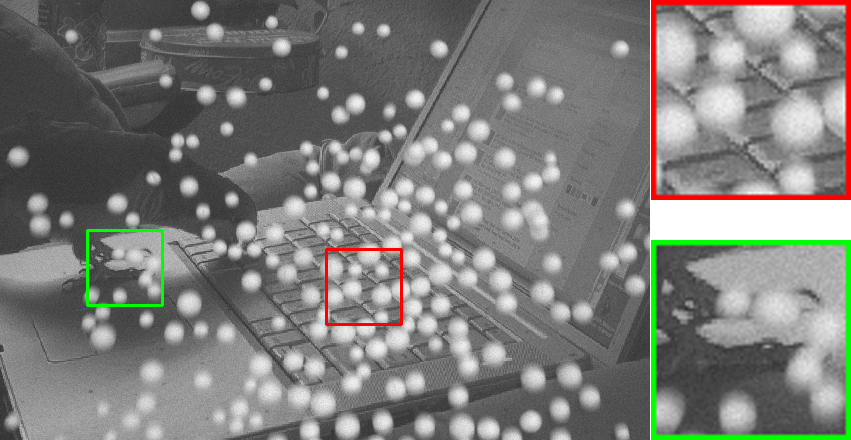}
		&\frame{\includegraphics[width=\linewidth]{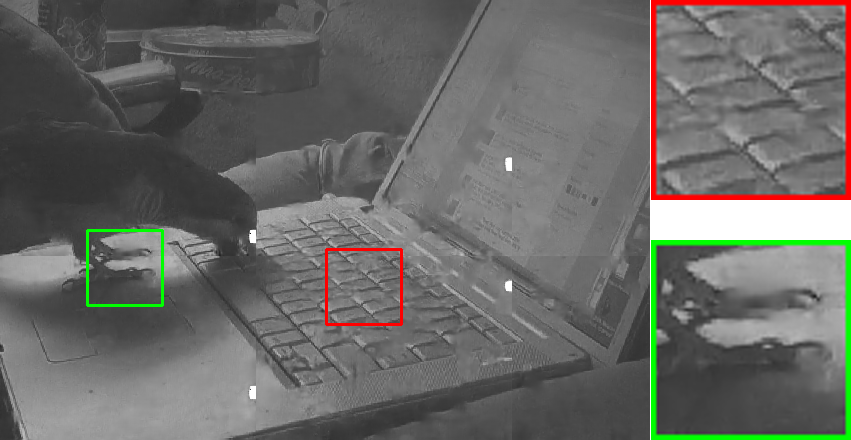}}
		&\frame{\includegraphics[width=\linewidth]{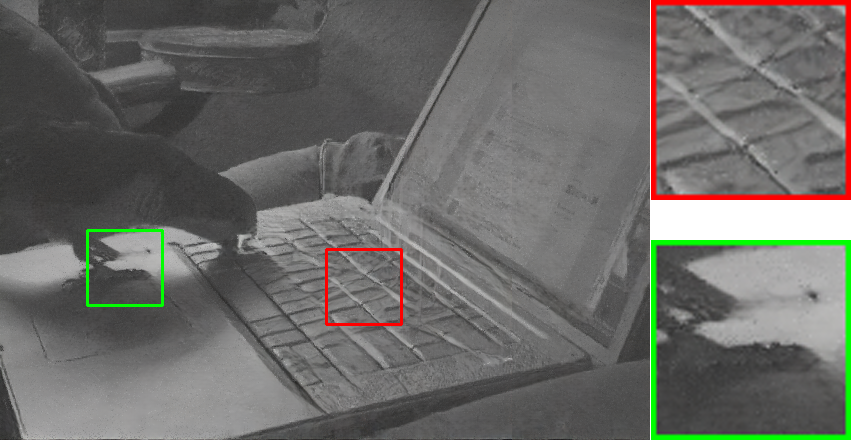}}
		&\frame{\includegraphics[width=\linewidth]{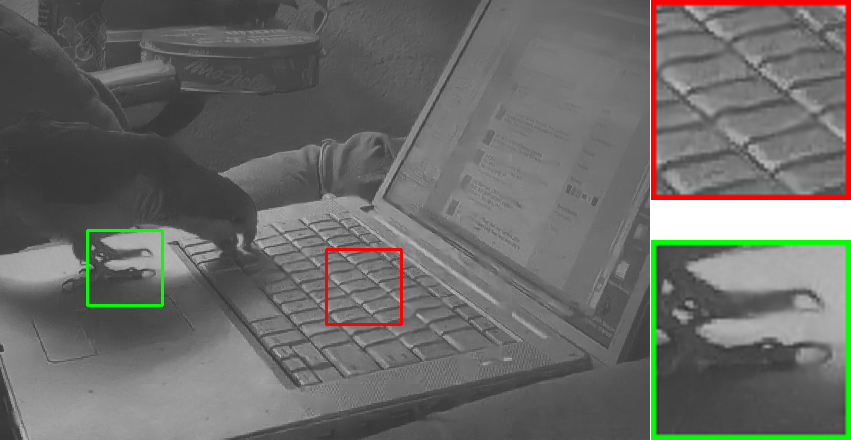}}
		&\frame{\includegraphics[width=\linewidth]{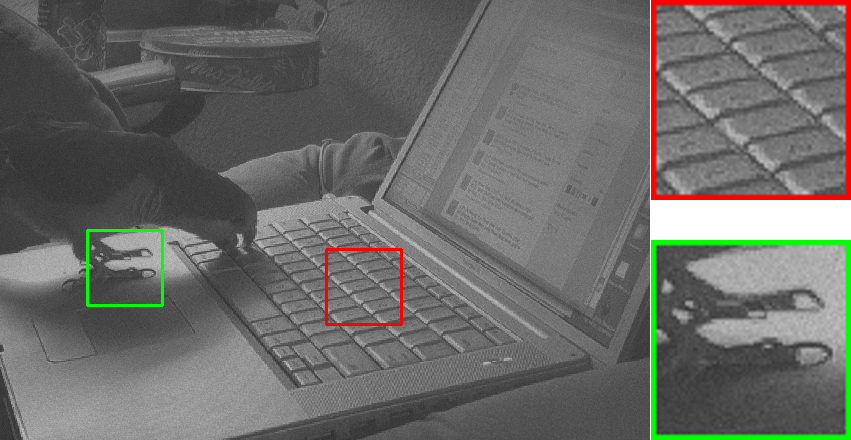}}
        \\
		\includegraphics[width=\linewidth]{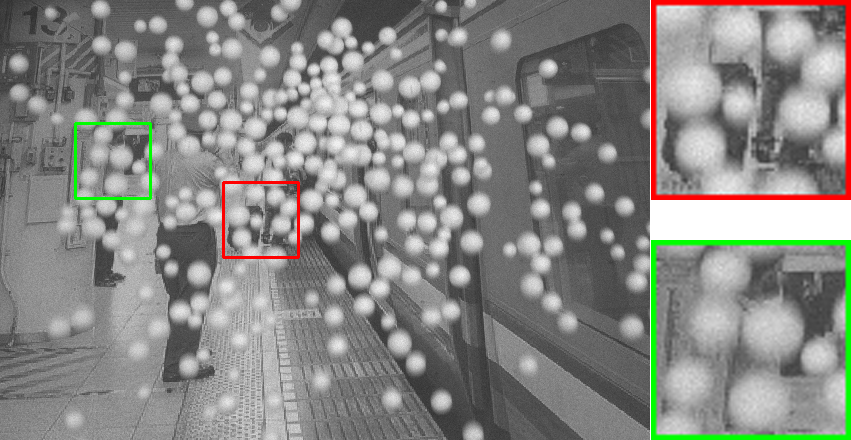}
		&\frame{\includegraphics[width=\linewidth]{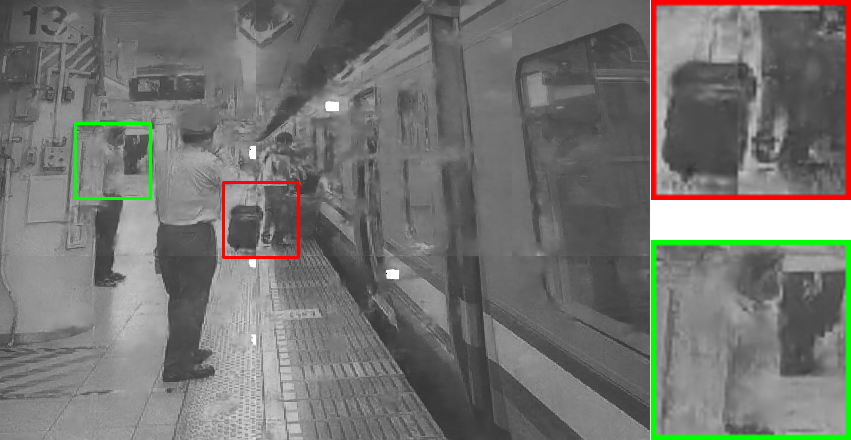}}
		&\frame{\includegraphics[width=\linewidth]{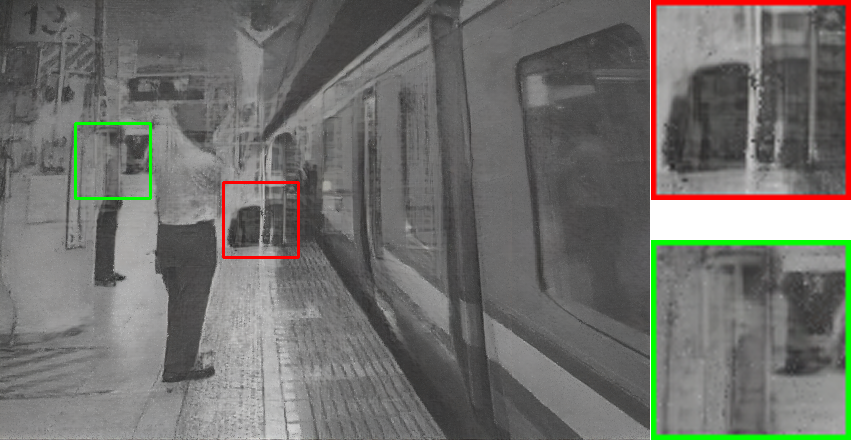}}
		&\frame{\includegraphics[width=\linewidth]{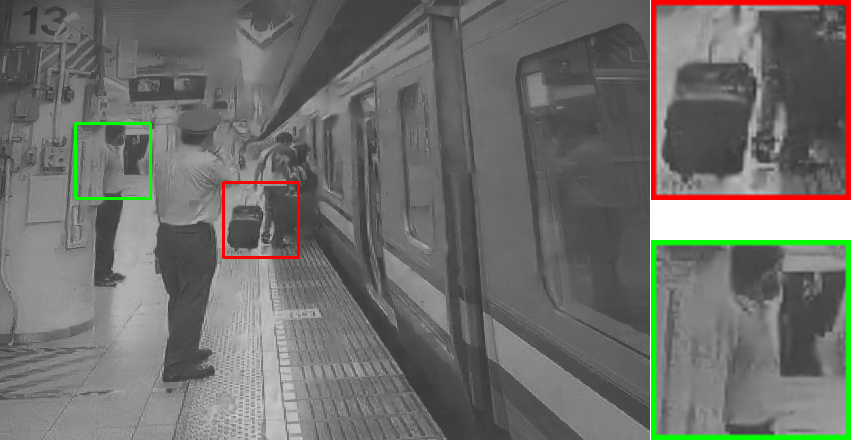}}
		&\frame{\includegraphics[width=\linewidth]{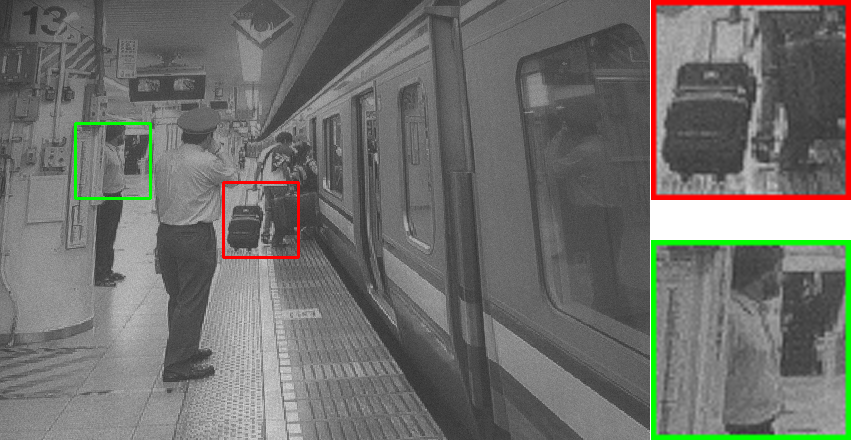}}
        \\
		\includegraphics[width=\linewidth]{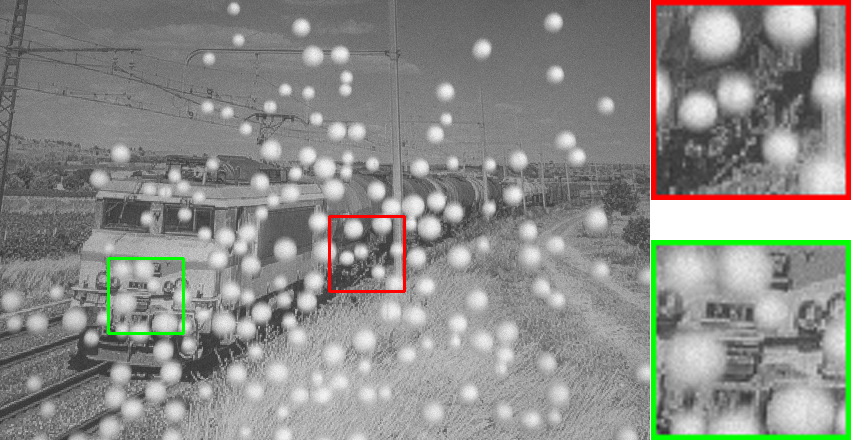}
		&\frame{\includegraphics[width=\linewidth]{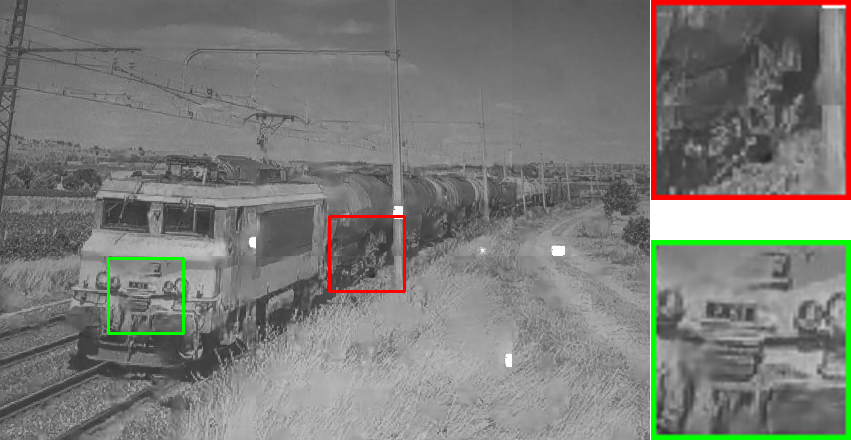}}
		&\frame{\includegraphics[width=\linewidth]{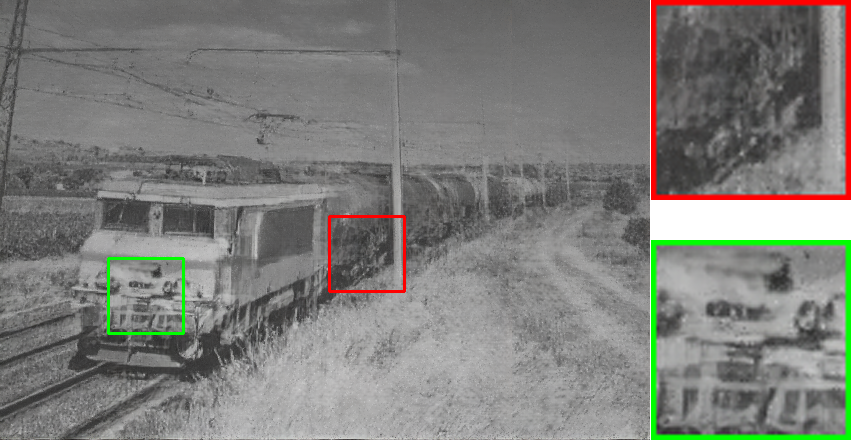}}
		&\frame{\includegraphics[width=\linewidth]{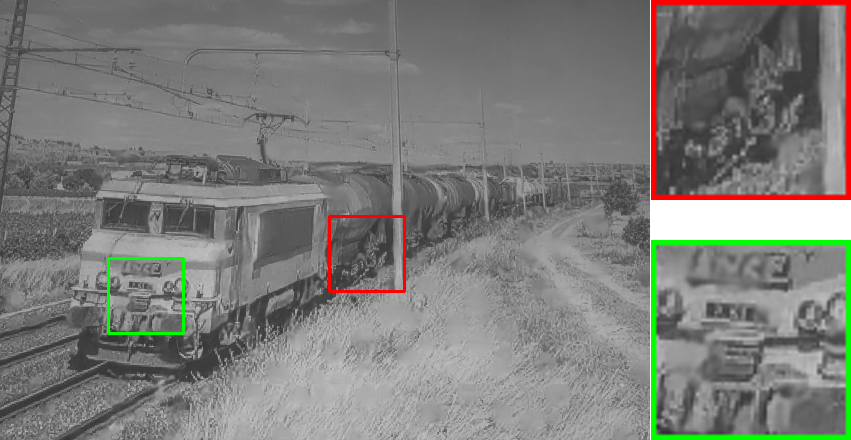}}
		&\frame{\includegraphics[width=\linewidth]{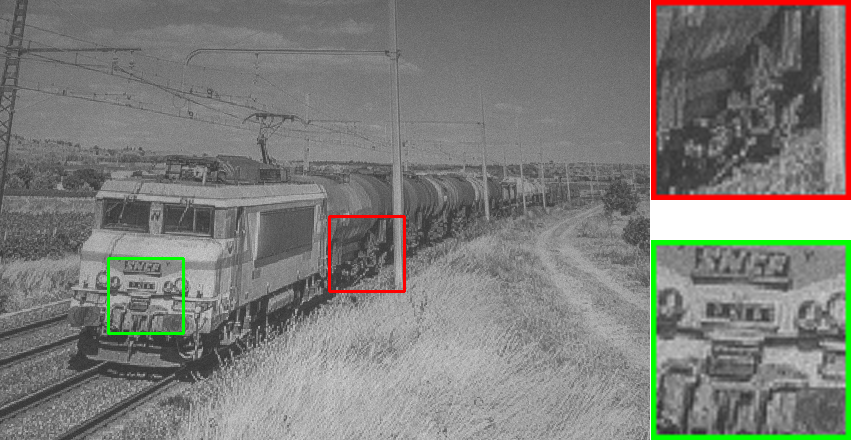}}
	\end{tabular}
    \vspace{-1ex}
     \caption{Images showing the occluded input frame, the reconstructed frame of the best-performing frame-based and event-based baselines and our method, and the ground truth frame for our real-world dataset.
 }
	\label{fig:real_qual}
\end{figure*}

\begin{table}[!t]
    \centering
    \begin{adjustbox}{max width=\linewidth}
    \setlength{\tabcolsep}{4pt}
    {\small
    \begin{tabular}{lcccc}
        \toprule
         Method & Input  & PSNR $\uparrow$  &  SSIM $\uparrow$ & MAE $\downarrow$ \\
        \midrule
        MAT \cite{li_mat22cvpr} & I & 26.7451 & 0.5670 & 0.0285 \\
        MISF \cite{li_misf22cvpr}& I & 29.0281 & 0.6951 & 0.0222  \\
        PUT \cite{liu22cvpr} & I & 18.9135 & 0.2726 & 0.0801  \\
        ZITS \cite{dong22cvpr}& I & 30.2675 & 0.7542 & 0.0190  \\
        EF-SAI \cite{liao22cvpr}& I+E & 30.2390 & 0.8037 & 0.0194  \\
        E2VID \cite{Rebecq19cvpr}& E & 16.5642 & 0.3203 & 0.1066  \\
        Ours ( (Acc. Method)& E & 19.5110 & 0.5685 & 0.0606  \\
        Ours (Learning) & I+E & \textbf{33.0701} & \textbf{0.8173} & \textbf{0.0166}  \\
        \bottomrule
    \end{tabular}}
    \end{adjustbox}
    \caption{Reconstruction performance on our real-world dataset.}
    \label{tab:real_gray}
\end{table}

\section{Conclusion}
We introduce a novel event-based approach for background image reconstruction in the presence of dynamic occlusions.
It leverages the complementary nature of event camera and frames to reconstruct true scene information instead of hallucinating occluded areas as done by image inpainting approaches.
Specifically, our proposed data-driven approach reconstructs the background image using only one occluded image and events.
The high temporal resolution of the events provides our method additional information on the relative intensity changes between the foreground and background, making it robust to dense occlusions. %
To evaluate our approach, we present the first large-scale dataset recorded in the real world containing over $230$ challenging scenes with synchronized events, occluded images, and groundtruth images.
Our method achieves an improvement of 3dB in PSNR over state-of-the-art frame-based and event-based methods on both synthetic and real datasets.
We will release our synthetic and recorded dataset representing the first datasets for background image reconstruction using events and images in the presence of dynamic occlusions.
We believe that our proposed method and dataset lay the foundation for future research.

\title{\MYTITLE\\---Supplementary Material---}
\maketitle
\thispagestyle{empty}

\section{Network Structure}

In this section, we provide details of the structure of our network.
As explained in the main paper, our network consists of the \textit{Frame Encoder}, the \textit{Event Encoder}, the \textit{Event Accumulation Module} (EAM), the \textit{Occlusion-aware Feature Fusion} (OFF) module and the \textit{Reconstruction Decoder}.

\begin{figure}[ht!]
    \centering
    \includegraphics[width=0.45\textwidth]{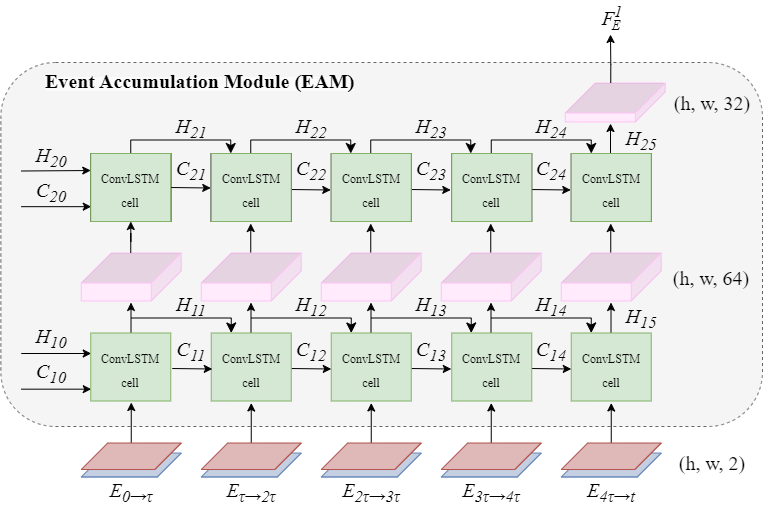}
    \caption{
    The EAM in our proposed method which integrates $N=5$ event representations $E_{0\rightarrow \tau}, ..., E_{(N-1) \tau \rightarrow t}$ using two ConvLSTM layers. The final hidden state of the second ConvLSTM layers formulates the first-scale event feature $F_E^1$ in the multi-scale \textit{Event Encoder}.
    }
    \label{fig:EAM}
\end{figure}

The architecture of the Frame Encoder and the Event Encoder is shown in \Tab \ref{tab:encoder}. 
Both encoders have the same structure except at the first two scales, where the frame encoder is provided with a grayscale image of dimension (h, w, 1) and the event encoder receives the output of the EAM of dimension (h, w, 32).
The architecture of the EAM module is shown in \Fig \ref{fig:EAM}.

We provide the details of the OFF module in \Tab \ref{tab:off}. 
The OFF module computes gating weights $W^{j}$  at each scale $j$ based on the current-scale event features $F^j_E$ and frame features $F^j_I$ as well as the previous-scale fused features $F^{j-1}_F$ downscaled to  $\bar{F}^{j-1}_F$. 
The fused feature $F^j_F$ is then obtained based on the current-scale event and frame features $F^j_E$ and $F^j_I$, the gating weights $W^{j}$, and the downsampled previous-scale fused feature $\bar{F}^{j-1}_F$.
The OFF module can be described with the following equations:
\begin{equation}
    \tilde{F}^j_F = G(F^j_E, F^j_I)
\end{equation}
\begin{equation}
    W^{j} =   G(\tilde{F}^j_F , \bar{F}^{j-1}_F)
\end{equation}
\begin{equation}
    \hat{F}^j_F = (1-W^{j}) F^j_I +   W^j F^j_E 
\end{equation}
\begin{equation}
    F^j_F = G(\hat{F}^j_F , \bar{F}^{j-1}_F)
\end{equation}
where $G$ indicates concatenation followed by convolution operations.

Finally, the structure of the \textit{Reconstruction Decoder} is shown in \Tab \ref{tab:decoder} which uses the fused features at each scale as skip connections to predict the residual image.
\begin{table}[!t]
        \centering
        \begin{adjustbox}{max width=\linewidth}
        \setlength{\tabcolsep}{4pt}
        {\small
        \begin{tabular}{lcc|ccl}
            \toprule
            &\multicolumn{2}{c}{\bfseries Frame Encoder}
            &\multicolumn{2}{c}{\bfseries Event Encoder}\\
            \cmidrule{2-3}  \cmidrule{4-5} 
            Scale $j$ & Layer  & Out Size  $(h^j, w^j, c^j)$ & Layer & Out Size  $(h^j, w^j, c^j)$\\
            \midrule 
            0 & input image & (h, w, 1) & -- &  -- \\
            1 & conv(5,1,2) & (h, w, 32) & -- & -- \\
            2 & conv(5,2,2) & (h/2, w/2, 64) & conv(5,2,2) & (h/2, w/2, 64)\\
            3 & conv(5,2,2) & (h/4, w/4, 128) & conv(5,2,2) & (h/4, w/4, 128)\\
            4 & conv(5,2,2) & (h/8, w/8, 256) & conv(5,2,2) & (h/8, w/8, 256) \\
            5 & conv(5,2,2) & (h/16, w/16, 512) & conv(5,2,2) & (h/16, w/16, 512)\\
            6 & conv(3,1,1) & (h/16, w/16, 512) & conv(3,1,1) & (h/16, w/16, 512) \\
            \bottomrule
        \end{tabular}}
        \end{adjustbox}
        \caption{
         The architecture of the \textit{Frame Encoder} and \textit{Event Encoder} in our proposed method, in which $(h^j, w^j, c^j)$ denotes the height, width, and number of channels of the output of the layer corresponding to scale $j$, and conv($k$, $s$, $p$) denotes a convolution block with kernel size $k$, stride $s$ and padding $p$. 
        The input layer corresponds to $j=0$, and the input size used in our paper is h$=h^0=384$, w$=w^0=512$.  
        }
        \label{tab:encoder}
    \end{table}
    
    \begin{table}
        \begin{adjustbox}{max width=\linewidth}
        \setlength{\tabcolsep}{4pt}
        {\small
        \begin{tabular}{@{}cccc@{}}
            \hline
            \multicolumn{4}{c}{\bfseries Occlusion-aware Feature Fusion (OFF)}\\
            \hline
            Input & Input Size & Layer & Output\\
            \midrule
            $F_E^j$ \& $F_I^j$ & $(h^j, w^j, c^j)$ & conv(5,1,2) $\times$ 2 & $\tilde{F}_F^j$ \\
            $F_F^{j-1}$ &$(h^{j-1}, w^{j-1}, c^{j-1})$ & conv(5,$s^j$,2) & $\bar{F}_F^{j-1}$ \\
            $\bar{F}_F^{j-1}$ \& $\tilde{F}_F^j$  & $(h^j, w^j, c^j)$ & conv(5,1,2) $\times$ 2 & $W^{j}$ \\
            $F_E^j$, $F_I^j$ \& $W^{j}$ & $(h^j, w^j, c^j)$ & Feature Selection  & $\hat{F}_F^{j}$ \\
            $\hat{F}_F^{j}$ \& $\bar{F}_F^{j-1}$ & $(h^j, w^j, c^j)$ &  conv(5,1,2) $\times$ 2 & $F_F^{j}$ \\
            \bottomrule
        \end{tabular}}
        \end{adjustbox}
        \caption{The structure of the \textit{Occlusion-aware Feature Fusion} module in the proposed method.}
        \label{tab:off}
    \end{table}

    \begin{table}
        \begin{adjustbox}{max width=\linewidth}
        \setlength{\tabcolsep}{4pt}
        {\small
        \begin{tabular}{@{}lccc@{}}
            \toprule
            &\multicolumn{3}{c}{\bfseries Reconstruction Decoder}\\
            \cmidrule{2-4}
            Scale & Input Size & Layer & Output Size\\
            \midrule
            6 &(h/16, w/16, 512+512) & up(2)-conv(5,1,2) & (h/8, w/8, 512) \\
            5 &(h/8, w/8, 512+256) & up(2)-conv(5,1,2) & (h/4, w/4, 256)\\
            4 &(h/4, w/4, 256+128) & up(2)-conv(5,1,2) & (h/2, w/2, 128) \\
            3 &(h/2, w/2, 128+64) & up(2)-conv(5,1,2) & (h, w, 64) \\
            2 &(h, w, 64+32) & conv(3,1,1) & (h, w, 32) \\
            1 &(h, w, 32)    & conv(1,1,0) & (h, w, 1) \\
            \bottomrule
        \end{tabular}}
        \end{adjustbox}
        \caption{The structure of the \textit{Reconstruction Decoder}, in which 'up(2)' denotes updampling by 2.}
        \label{tab:decoder}
    \end{table}

\section{Detailed Ablation Results}
Here we present a more detailed ablation study on the network architecture.
\subsection{Importance of Event and Frame Encoders}

\global\long\def\figWidth{0.17\linewidth}
\begin{figure*}
	\centering
    \setlength{\tabcolsep}{1pt}
	\begin{tabular}{
	M{\figWidth}
	M{\figWidth}
	M{\figWidth}
	M{\figWidth}
	M{\figWidth}
    M{\figWidth}
	M{\figWidth}}
		Occluded & Shared F \& E Encoder &  Indep. F Encoder& Indep. E Encoder & Indep. F \& E Encoder w. OFF & Groundtruth
		\\
		\includegraphics[width=\linewidth]{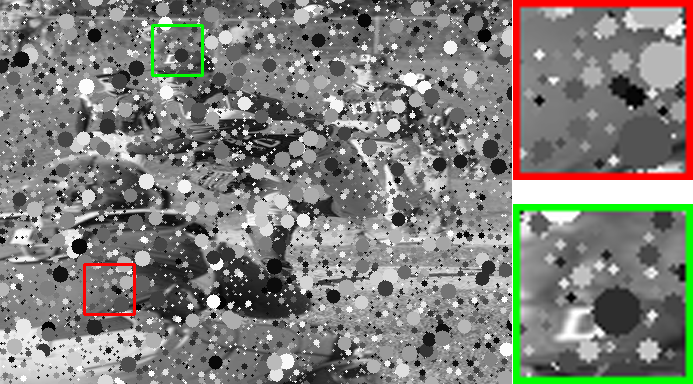}
		&\frame{\includegraphics[width=\linewidth]{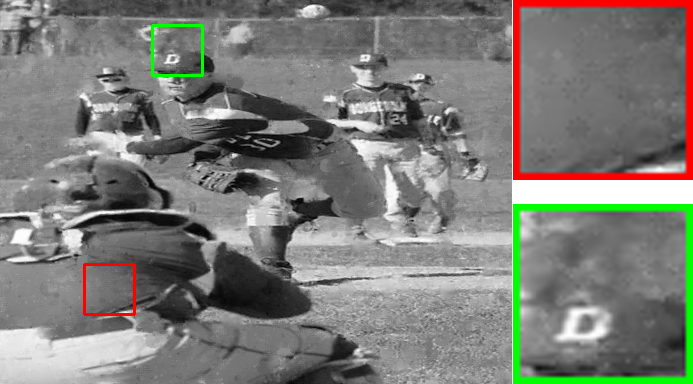}}
		&\frame{\includegraphics[width=\linewidth]{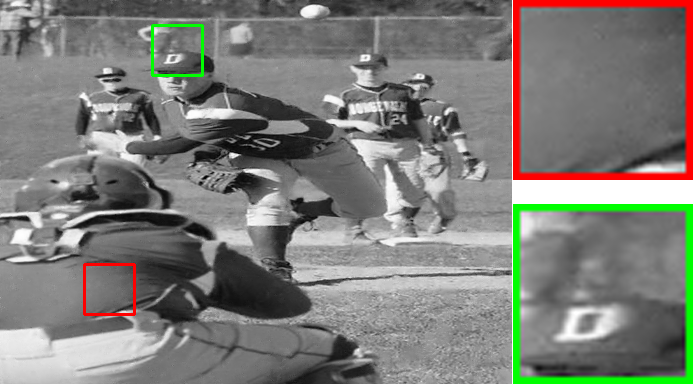}}
		&\frame{\includegraphics[width=\linewidth]{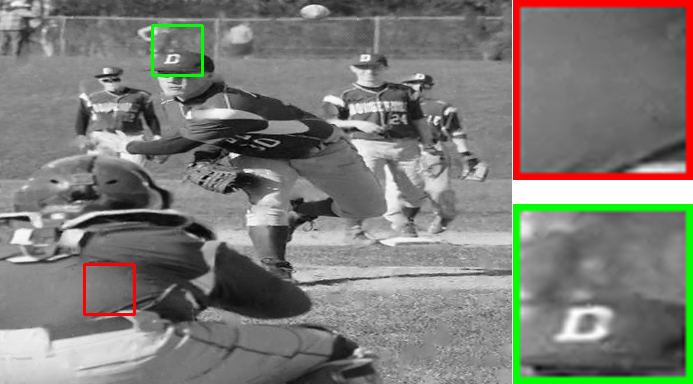}}
        &\frame{\includegraphics[clip,width=\linewidth]{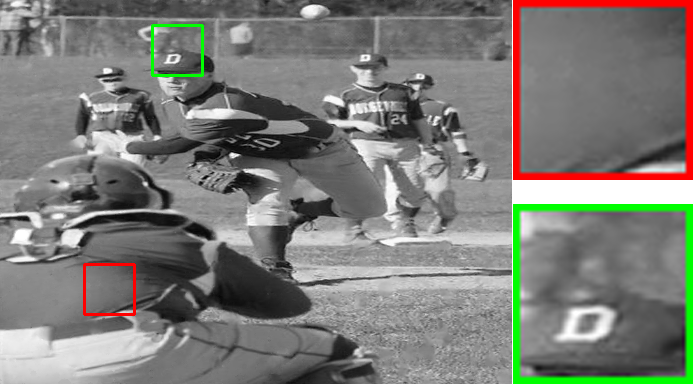}}
		&\frame{\includegraphics[width=\linewidth]{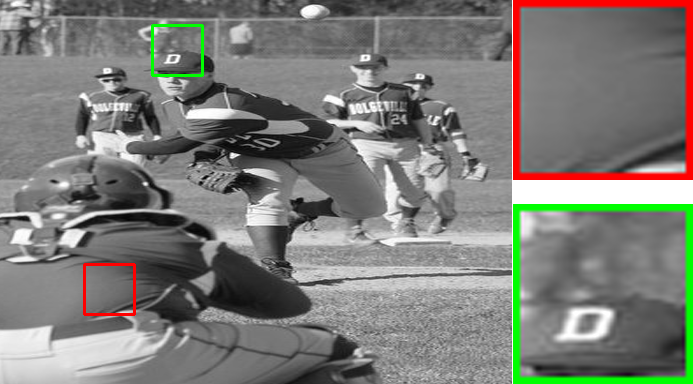}}
        \\
		\includegraphics[width=\linewidth]{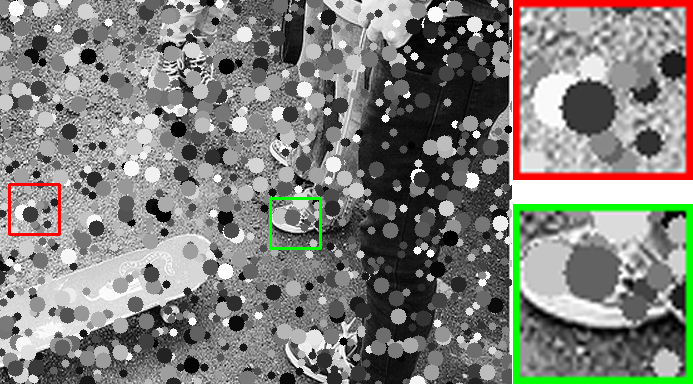}
		&\frame{\includegraphics[width=\linewidth]{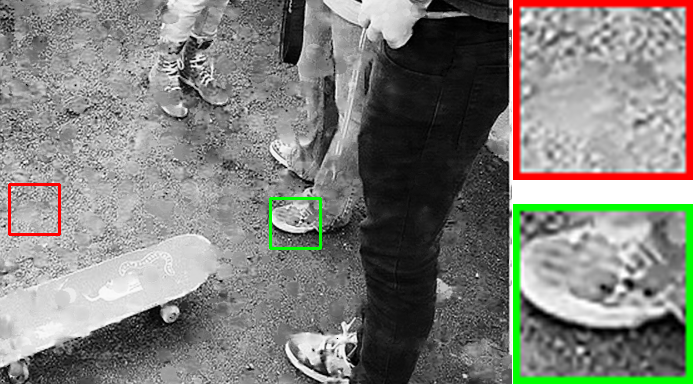}}
		&\frame{\includegraphics[width=\linewidth]{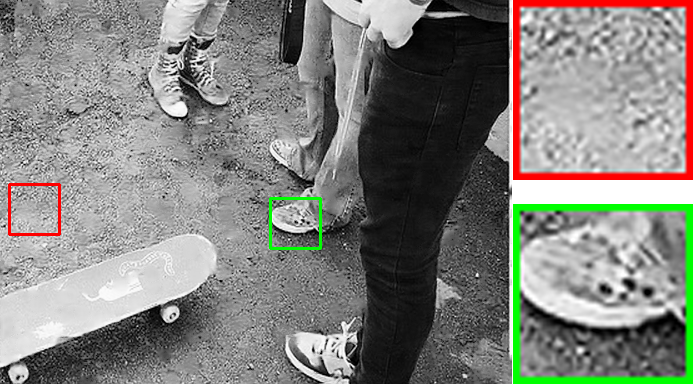}}
		&\frame{\includegraphics[width=\linewidth]{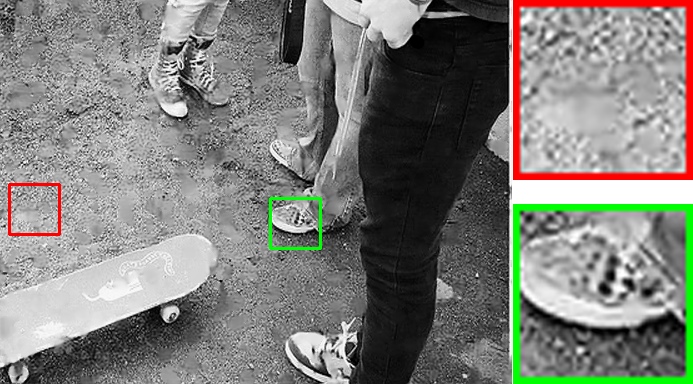}}
        &\frame{\includegraphics[clip,width=\linewidth]{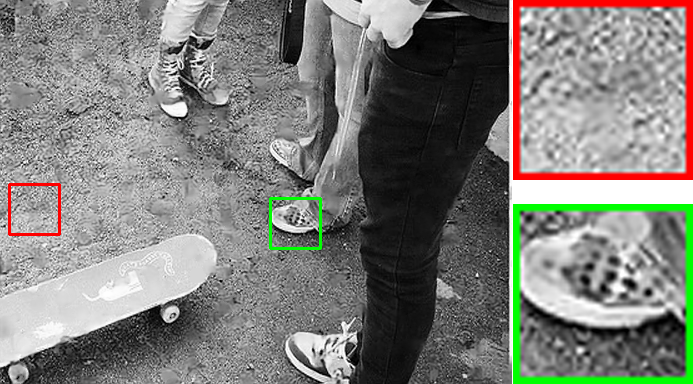}}
		&\frame{\includegraphics[width=\linewidth]{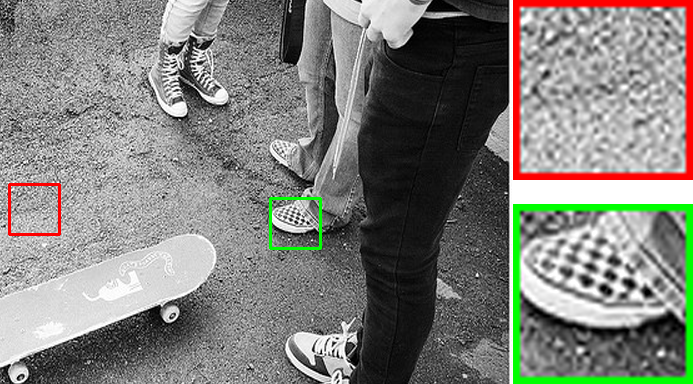}}
        \\
        \includegraphics[width=\linewidth]{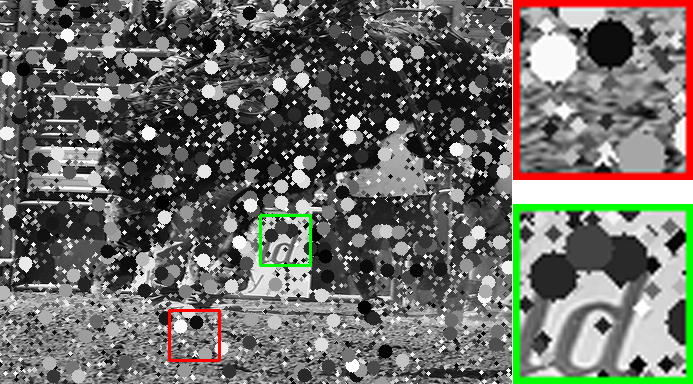}
		&\frame{\includegraphics[width=\linewidth]{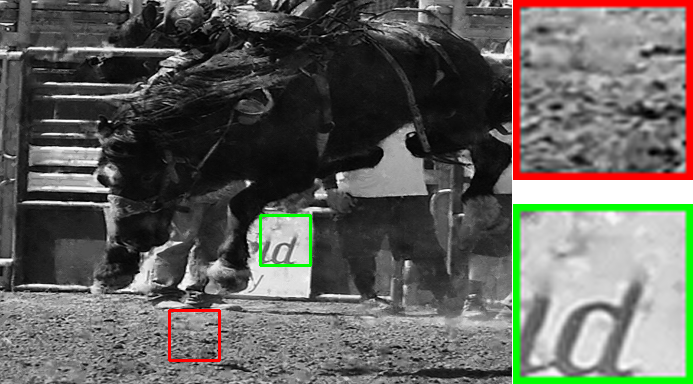}}
		&\frame{\includegraphics[width=\linewidth]{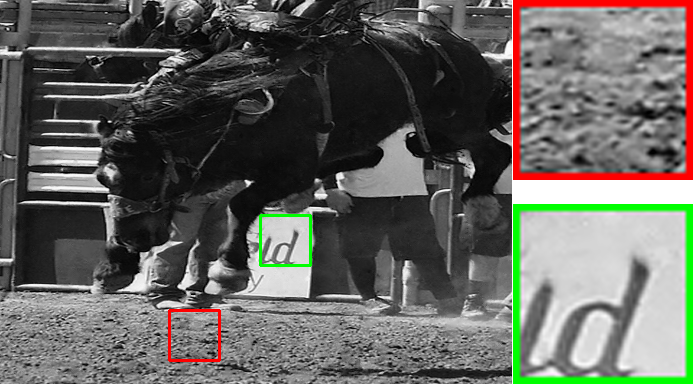}}
		&\frame{\includegraphics[width=\linewidth]{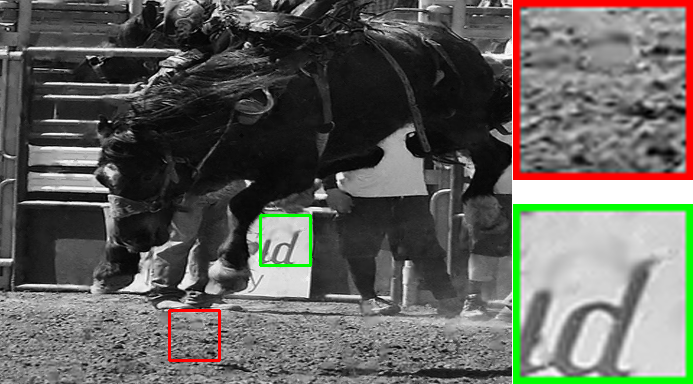}}
        &\frame{\includegraphics[clip,width=\linewidth]{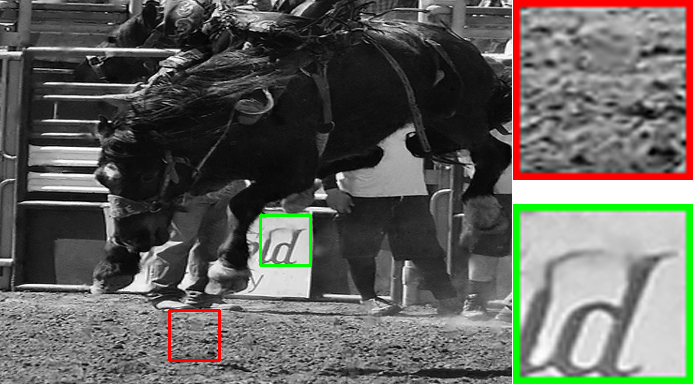}}
		&\frame{\includegraphics[width=\linewidth]{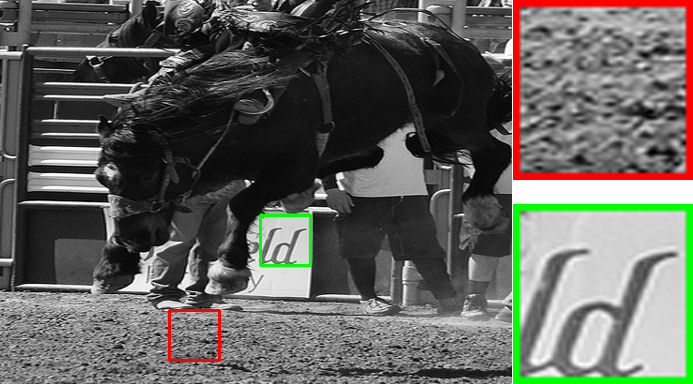}}
        \\
        \includegraphics[width=\linewidth]{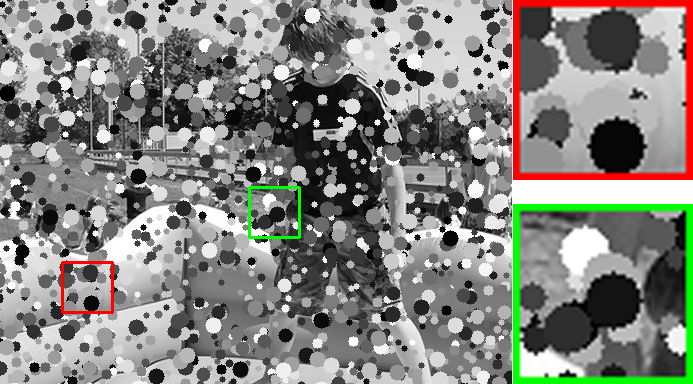}
		&\frame{\includegraphics[width=\linewidth]{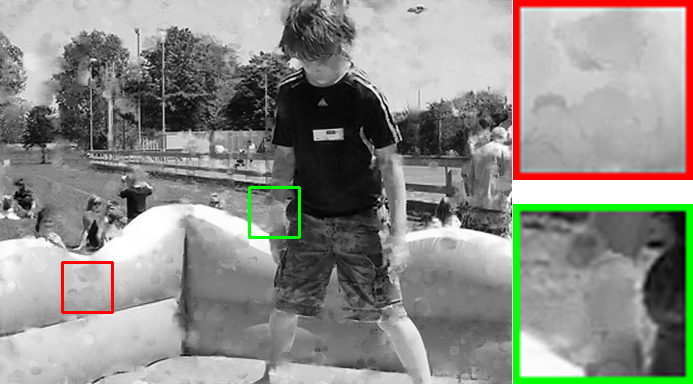}}
		&\frame{\includegraphics[width=\linewidth]{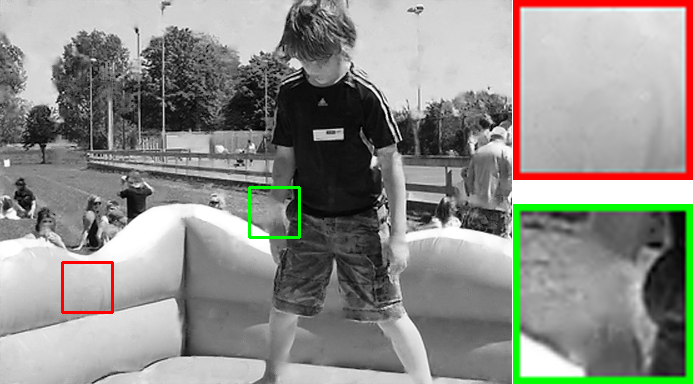}}
		&\frame{\includegraphics[width=\linewidth]{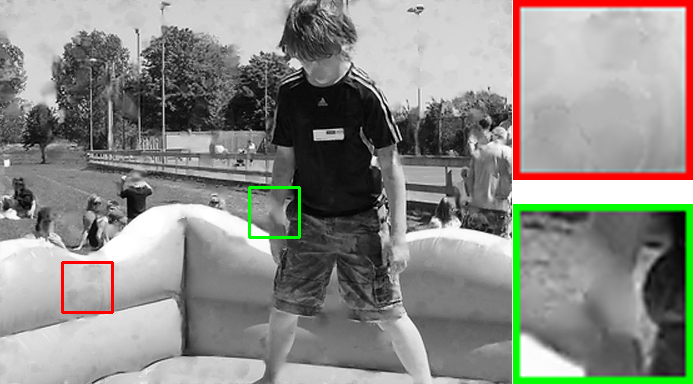}}
        &\frame{\includegraphics[clip,width=\linewidth]{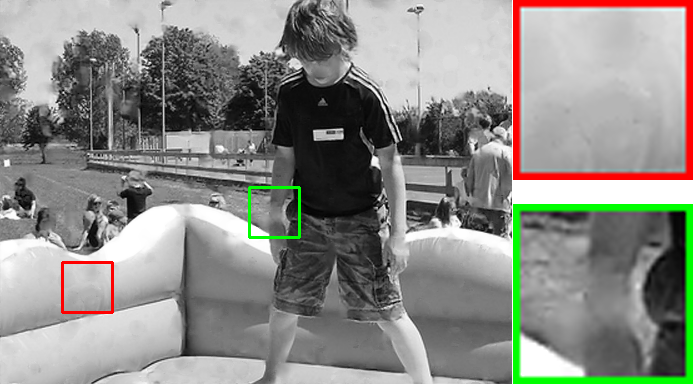}}
		&\frame{\includegraphics[width=\linewidth]{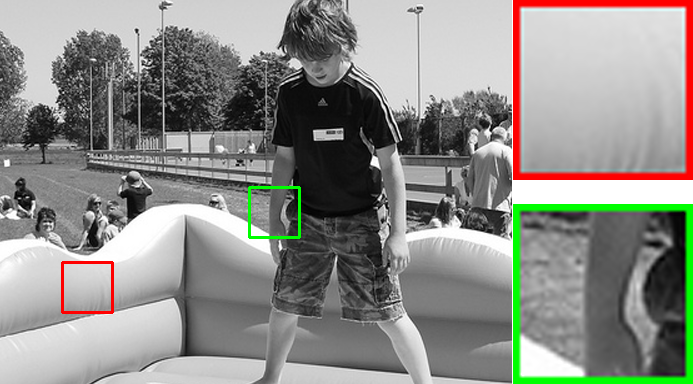}}
        \\
	\end{tabular}
    \vspace{-1ex}
	\caption{Images showing the importance of independent image and event encoders, conducted without EAM. A uniform region and a region containing interesting textures are shown in the red and green square boxes, respectively.}
	\label{fig:feature_importance}
\end{figure*}

\begin{table}[!h]
    \centering
    \begin{adjustbox}{max width=\linewidth}
    \setlength{\tabcolsep}{4pt}
    {\small
    \begin{tabular}{lccccc}
        \toprule
         Method & \multicolumn{2}{c} {\thead {Used Feature Scale \\ Image || Event}} & PSNR $\uparrow$  &  SSIM $\uparrow$ &  MAE $\downarrow$\\
        \midrule
        Shared F \& E Encoder & All scale & All scale & 27.8297 & 0.8642 & 0.0217\\
        Indep. F Encoder & All scale & 1st scale  & 31.6567 & 0.9297 & 0.0133\\
        Indep. E Encoder & 1st scale & All scale & 31.3298 & 0.9263 & 0.0125\\
        Indep. F \& E Encoder Simple Fusion & All scale & All scale & 31.0776 &  0.9212 & 0.0139\\
        Indep. F \& E Encoder with OFF & All scale & All scale & \textbf{32.7652} & \textbf{0.9425} &\textbf{0.0102}\\
        
        \bottomrule
    \end{tabular}}
    \end{adjustbox}
    \caption{ The effects of event and frame encoders and feature fusion methods on reconstruction performance evaluated on the synthetic dataset}
    \label{tab:feature_importance}
\end{table}

We study four different strategies to combine the event and frame information and train each corresponding network on our synthetic dataset. 
The first strategy combines events and frame by concatenating them at the input stage and uses a single encoder-decoder structure to predict the reconstructed image, termed as \textit{Shared F \& E Encoder}.
The second strategy, termed \textit{Independent Frame Encoder}, uses an independent frame encoder, which computes the frame features without considering the event information.
A second encoder takes as input the events and fuses for each subsequent scale the frame features from the independent frame encoder with the feature from the previous scale.
As the third strategy, we switch the inputs of the second strategy, i.e., we use an independent event encoder and a second encoder for fusing the event features with shallow frame features (referred to as \textit{Independent Event Encoder}).
Since we want to focus on the effect of processing the features for each sensing modality separately, we use two independent encoders for events and frame respectively and fuse features at each scale using a simple convolution layer.
This is referred to as \textit{Independent Event and Frame Encoder Simple Fusion}.
Lastly, to show the effect of a more sophisticated fusion, we also show the reconstructions using our OFF module to perform the fusion of the event and frame feature at each scale (\textit{Independent Event and Frame Encoder with OFF}).

We show the qualitative comparison of these approaches in \Fig \ref{fig:feature_importance} and provide quantitative evaluations in \Tab \ref{tab:feature_importance}.
The \textit{Shared F \& E Encoder} results in the worst performance in both the textured and untextured regions of the image.
The \textit{Independent Frame Encoder} results in an improved performance in uniform areas, whereas textured regions are poorly reconstructed, as can be observed in the third column of \Fig \ref{fig:feature_importance}.
The uniform texture of the red patch is captured perfectly by this network, indicating the importance of image features for uniform areas.
This can be explained by the fact that images contain absolute intensity information of surrounding spatial regions, which provides more information to the network for filling in the uniform patches.
However, the edges are not well preserved, which results in bleeding edges, as seen in the edge of the letter `D' in the first row of \Fig \ref{fig:feature_importance}.
The \textit{Independent Event Encoder}, in comparison to the \textit{Independent Frame Encoder}, shows better ability in reconstructing textured regions, which indicates the importance of event features for reconstructing textured areas. However, it struggles to remove occlusions in uniform regions,

Both the second and the third strategy performs significantly better than the first one, highlighting the importance of having separate encoders for frames and events to enable high-quality reconstruction of both textured and untextured regions. 
However, a vanilla fusion of \textit{Independent Event and Frame Encoder Simple Fusion} does not take full advantage of the strengths of the two respective modalities and therefore results in lower performance, as can be seen in the fourth row in \Tab \ref{tab:feature_importance}.
In contrast, the \textit{Independent Event and Frame Encoder with OFF} achieves higher performance than all the above-mentioned strategies. 
Especially, it can better reconstruct textured regions, e.g., the edge of the letter `D' is recovered, as visualized in the first row of \Fig \ref{fig:feature_importance}.
We believe that the differential nature of event cameras enables the preservation of high-frequency structures such as edges.

In summary, the proposed model using independent encoders for frame and events with a sophisticated fusion shows the best overall performance.

\subsection{Importance of EAM and OFF Modules}

\global\long\def\figWidth{0.17\linewidth}
\begin{figure*}
	\centering
    \setlength{\tabcolsep}{1pt}
	\begin{tabular}{
	M{\figWidth}
	M{\figWidth}
	M{\figWidth}
	M{\figWidth}
	M{\figWidth}
	M{\figWidth}}
		Occluded & w/o both & w/o EAM & w/o OFF & Ours & Groundtruth
		\\
		\includegraphics[width=\linewidth]{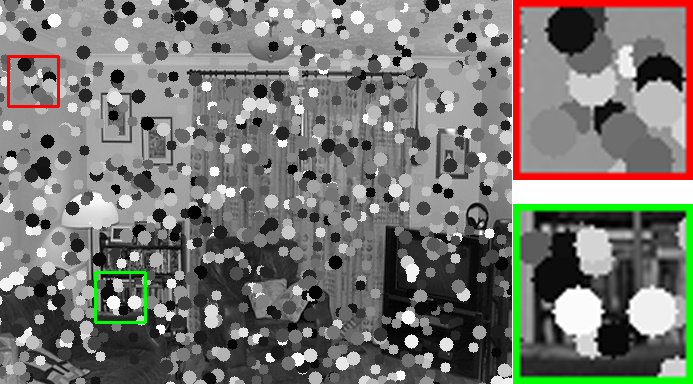}
		&\frame{\includegraphics[width=\linewidth]{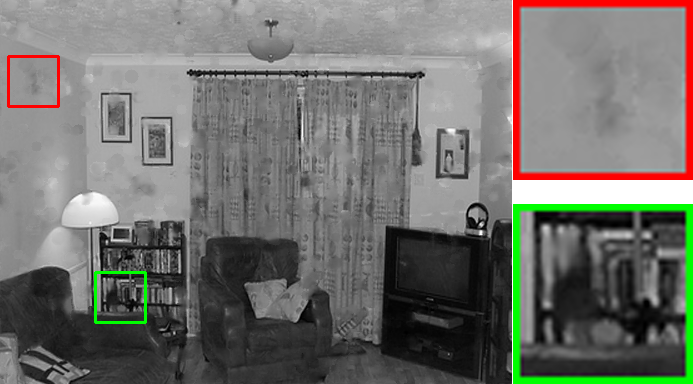}}
		&\frame{\includegraphics[clip,width=\linewidth]{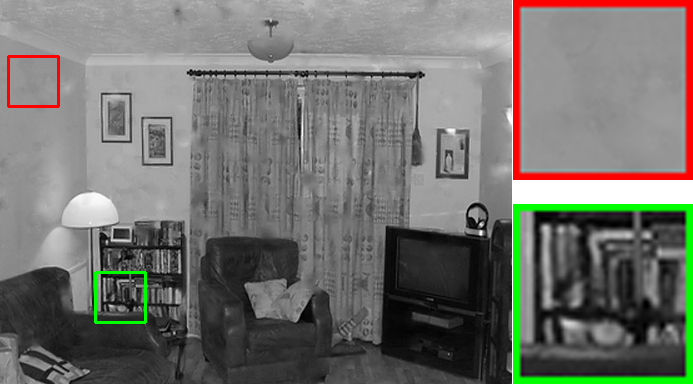}}
		&\frame{\includegraphics[width=\linewidth]{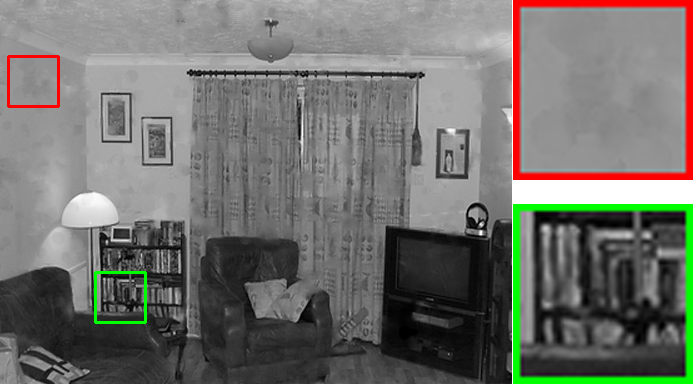}}
		&\frame{\includegraphics[width=\linewidth]{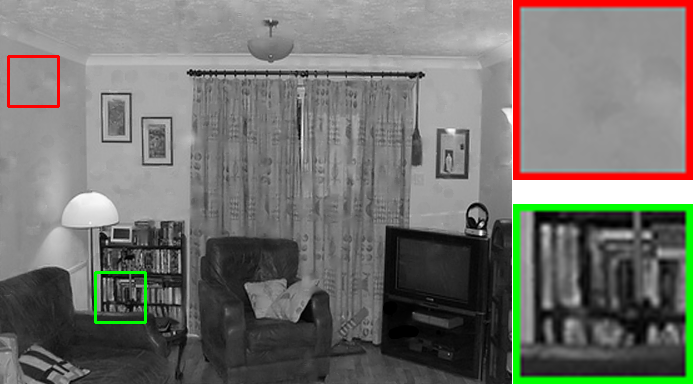}}
		&\frame{\includegraphics[width=\linewidth]{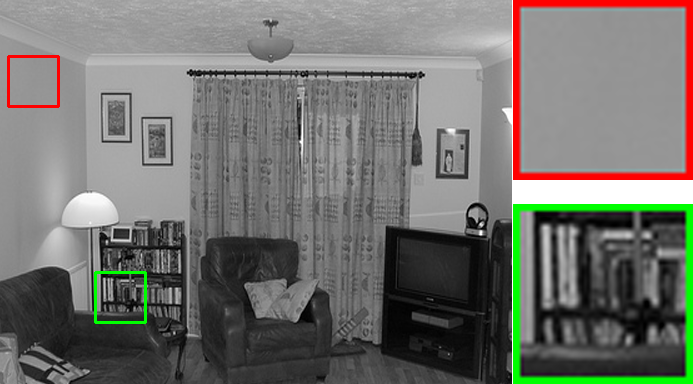}}
        \\
		\includegraphics[width=\linewidth]{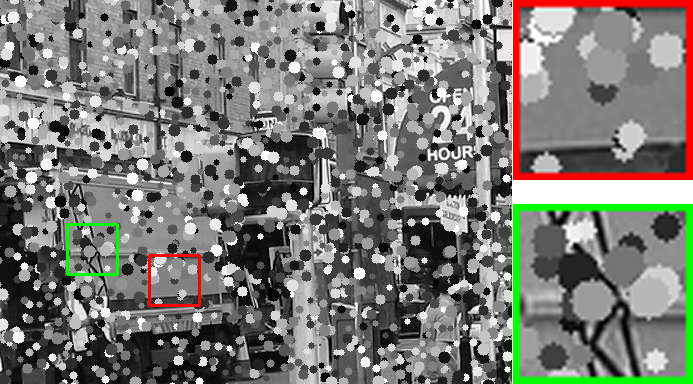}
		&\frame{\includegraphics[width=\linewidth]{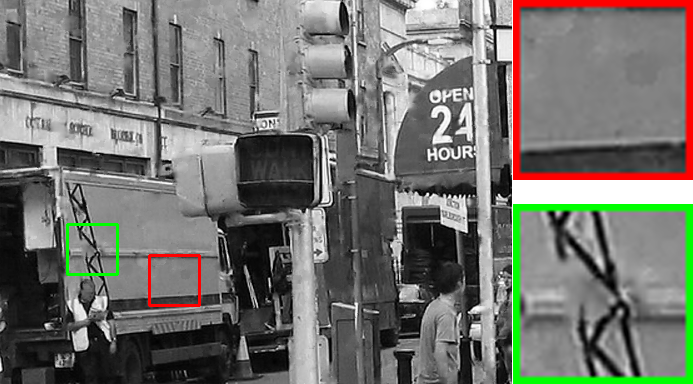}}
		&\frame{\includegraphics[clip,width=\linewidth]{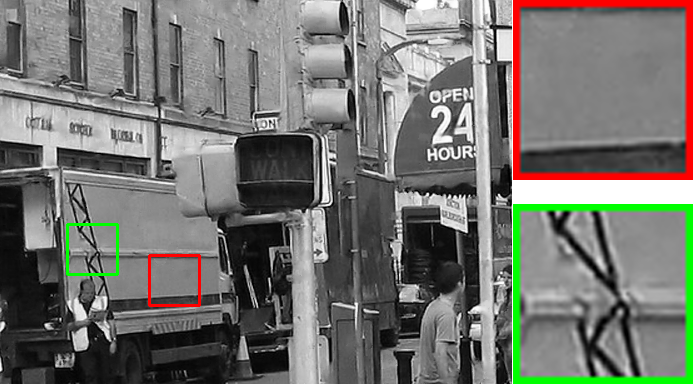}}
		&\frame{\includegraphics[width=\linewidth]{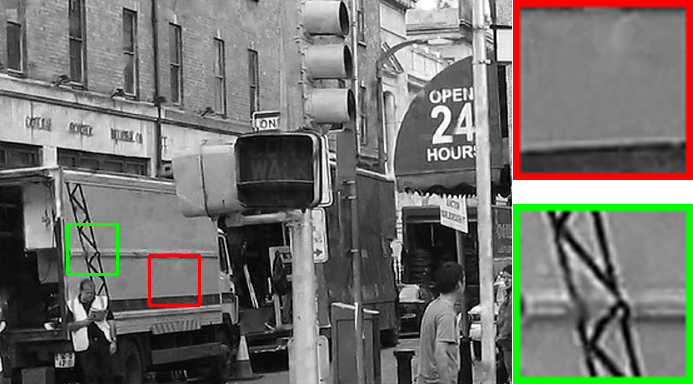}}
		&\frame{\includegraphics[width=\linewidth]{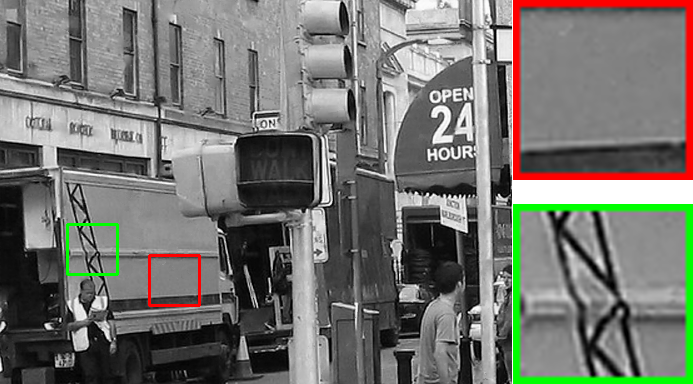}}
		&\frame{\includegraphics[width=\linewidth]{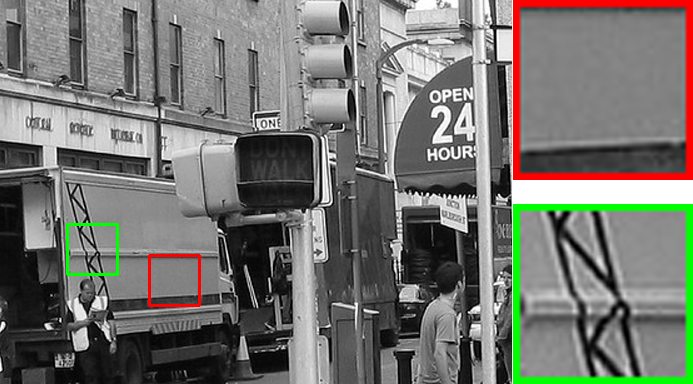}}
        \\
        \includegraphics[width=\linewidth]{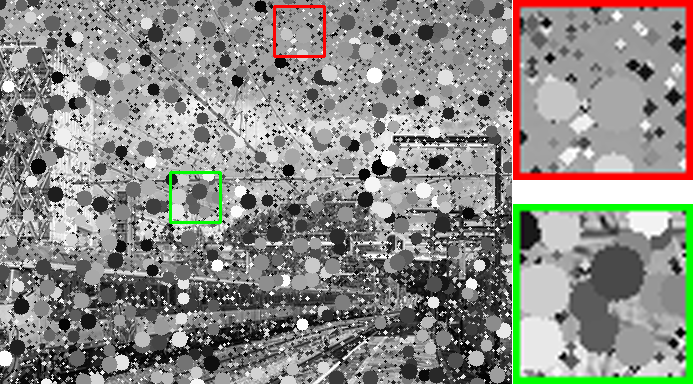}
		&\frame{\includegraphics[width=\linewidth]{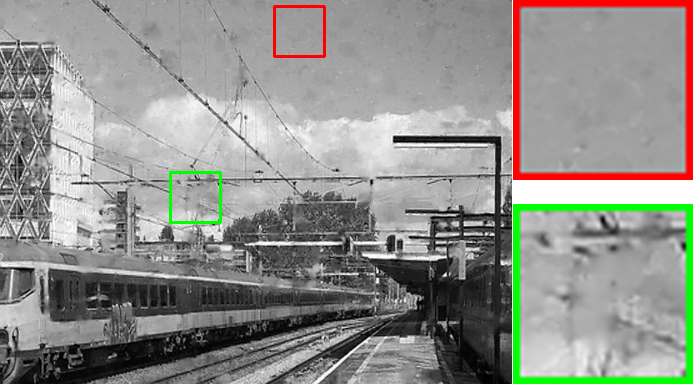}}
		&\frame{\includegraphics[clip,width=\linewidth]{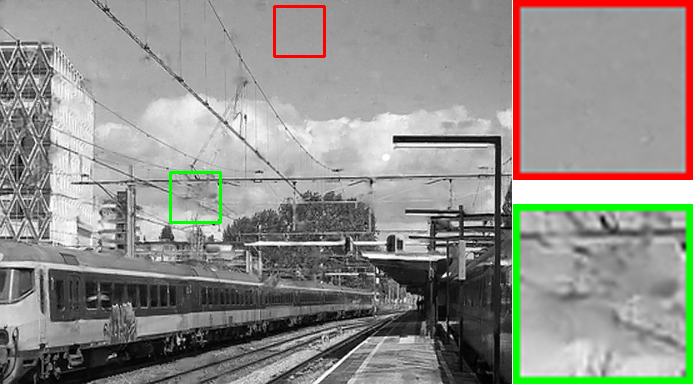}}
		&\frame{\includegraphics[width=\linewidth]{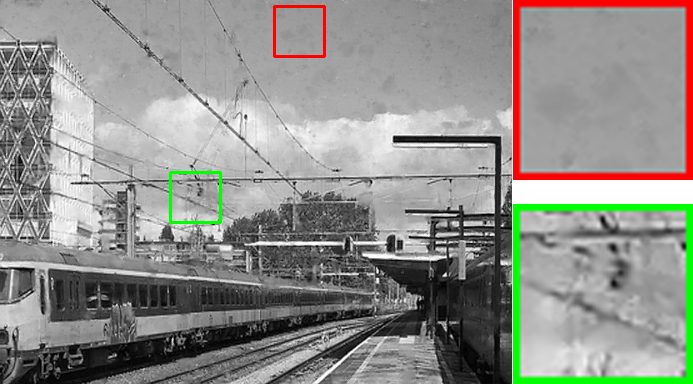}}
		&\frame{\includegraphics[width=\linewidth]{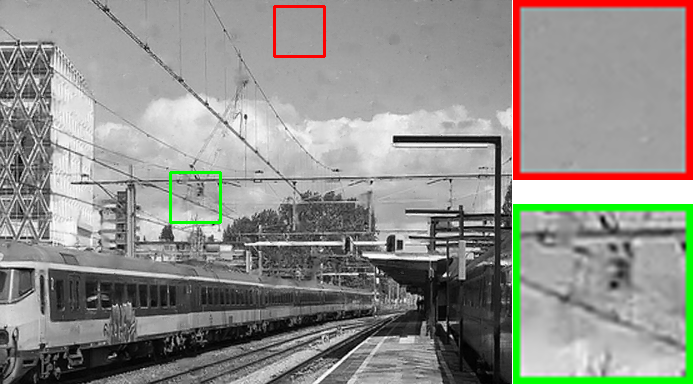}}
		&\frame{\includegraphics[width=\linewidth]{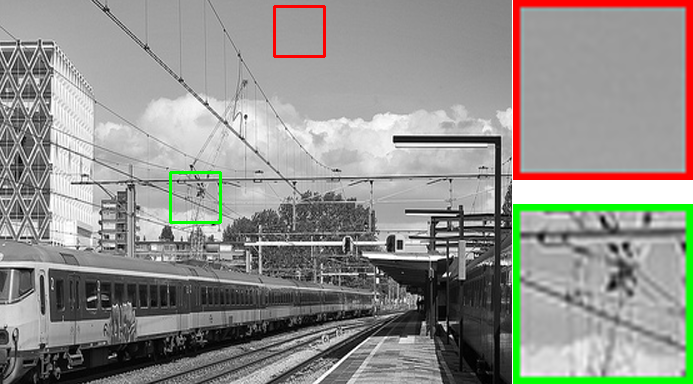}}
        \\
         \includegraphics[width=\linewidth]{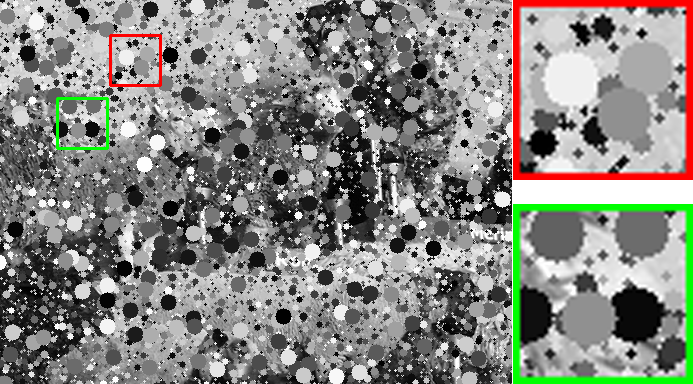}
		&\frame{\includegraphics[width=\linewidth]{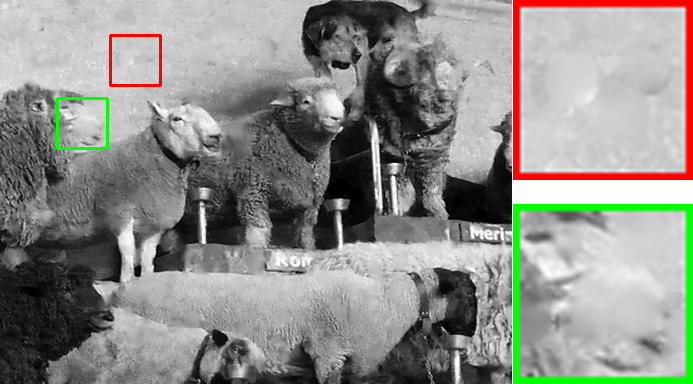}}
		&\frame{\includegraphics[clip,width=\linewidth]{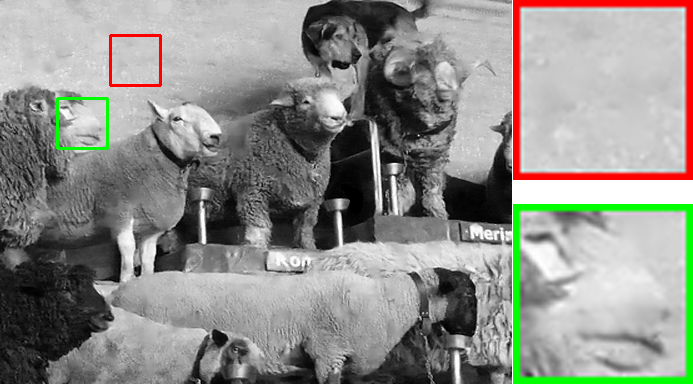}}
		&\frame{\includegraphics[width=\linewidth]{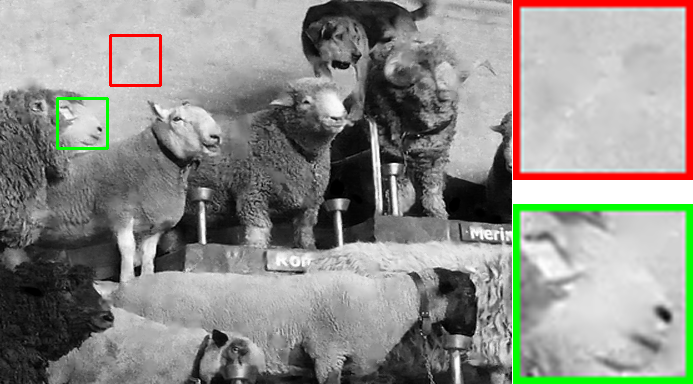}}
		&\frame{\includegraphics[width=\linewidth]{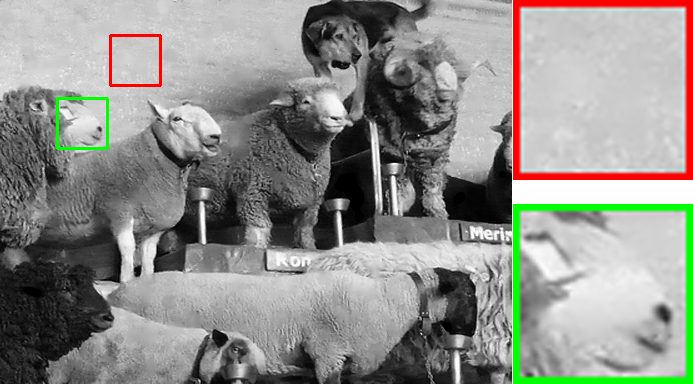}}
		&\frame{\includegraphics[width=\linewidth]{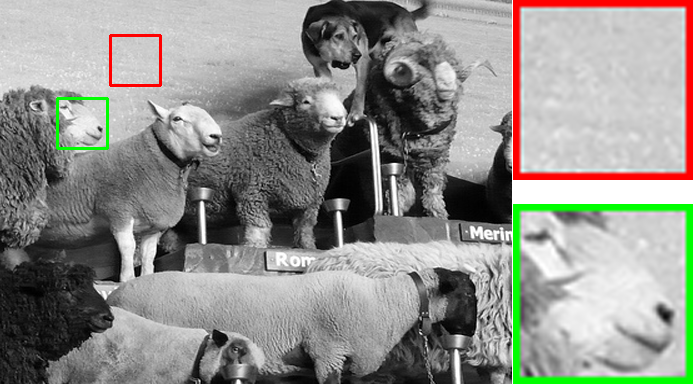}}
        \\
	\end{tabular}
    \vspace{-1ex}
	\caption{Images showing the importance of EAM and OFF modules. A uniform region and a region containing interesting textures are shown in the red and green square boxes, respectively.}
	\label{fig:ablation_module}
\end{figure*}

\begin{table}
    \centering
    \begin{adjustbox}{max width=\linewidth}
    \setlength{\tabcolsep}{4pt}
    {\small
    \begin{tabular}{lcccc}
        \toprule
         Method & Input  & PSNR $\uparrow$  &  SSIM $\uparrow$ & MAE$\downarrow$\\
        \midrule
        w/o EAM \& OFF & I+E & 31.0776 &  0.9212 & 0.0139\\
        w/o EAM & I+E & 32.7652 & 0.9425 & 0.0102\\
        w/o OFF & I+E & 32.8892 & 0.9378 & 0.0119\\
        Full & I+E & \textbf{34.6203} & \textbf{0.9536}  & \textbf{0.0085} \\
        \bottomrule
    \end{tabular}}
    \end{adjustbox}
    \caption{The reconstruction performance on the synthetic dataset of our network obtained by removing adaptively the introduced modules.}
    \label{tab:syn_ablation_supp}
\end{table}

In this section, we study the impact of \textit{Event Accumulation module} (EAM) and \textit{Occlusion-aware Feature Fusion} (OFF) on network performance.
The quantitative performance is shown in \Tab \ref{tab:syn_ablation_supp} and the qualitative results can be seen in \Fig \ref{fig:ablation_module}.
The base model without EAM and OFF results in the worst performance of $31dB$ PSNR, and adding EAM improves it by $1.8dB$ in terms of PSNR.
This gain is achieved because the EAM is designed to select event data that is relevant to the true background and filter out redundant information.
As can be seen in \Fig \ref{fig:ablation_module}, the network with EAM ($4^{th}$ column) shows better results in both uniform and textured regions compared to the base model. 
Also, in contrast to the two methods without EAM, the true background information is recovered to a significantly better extent in textured areas, even where there are multiple overlapping occlusions.

The performance of the network gains a substantial improvement from the OFF module as well, with a $1.7dB$ increase in PSNR. 
As illustrated in \Fig \ref{fig:ablation_module}, the OFF module also improves reconstruction quality in both uniform and textured areas when compared with the base model. 
In addition, compared to the two models without OFF, the one with OFF is significantly better at reconstructing uniform or non-textured regions.

Combining EAM and OFF, the full model achieves an increase of more than $3.5dB$ PSNR over the base model. Qualitatively, the full model performs better than all other models in both types of regions, implying that the two modules benefit from each other.

\section{Additional Qualitative Results}
\global\long\def\figWidth{0.165\linewidth}
\begin{figure*}
	\centering
    \setlength{\tabcolsep}{1pt}
	\begin{tabular}{
	M{0.25cm}
	M{\figWidth}
	M{\figWidth}
	M{\figWidth}
	M{\figWidth}
	M{\figWidth}
	M{\figWidth}
 }
		\\
		\rotatebox{90}{\makecell{Occluded}} 
        & \includegraphics[width=\linewidth]{figures/images/qual_results_syn/occ/0013.png} 
        &  \includegraphics[width=\linewidth]{figures/images/qual_results_syn/occ/0134.png}
        &  \includegraphics[width=\linewidth]{figures/images/qual_results_syn/occ/0251.png}
        &  \includegraphics[width=\linewidth]{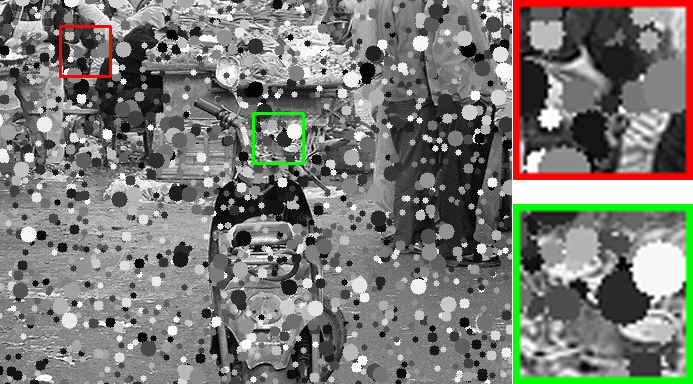}
        &  \includegraphics[width=\linewidth]{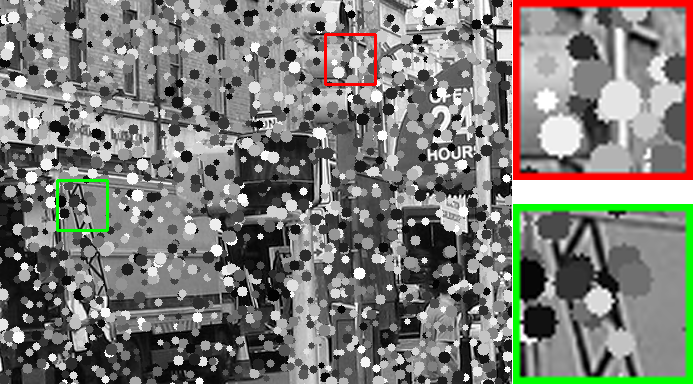}
        &  \includegraphics[width=\linewidth]{figures/images/qual_results_syn/occ/0454.png}
        \\
        \rotatebox{90}{\makecell{E2VID \cite{Rebecq19cvpr}}}    
        &\frame{\includegraphics[width=\linewidth]{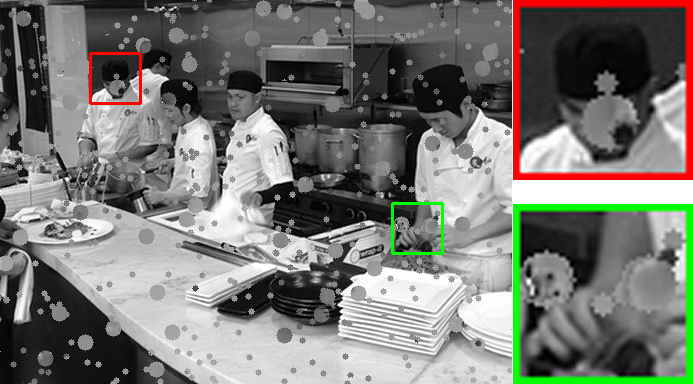}}
        &\frame{\includegraphics[width=\linewidth]{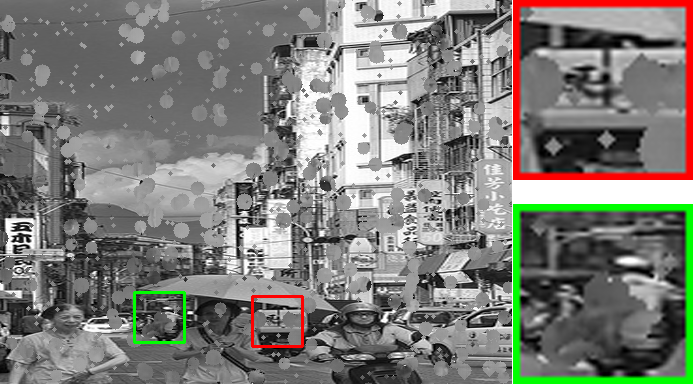}}
        &\frame{\includegraphics[width=\linewidth]{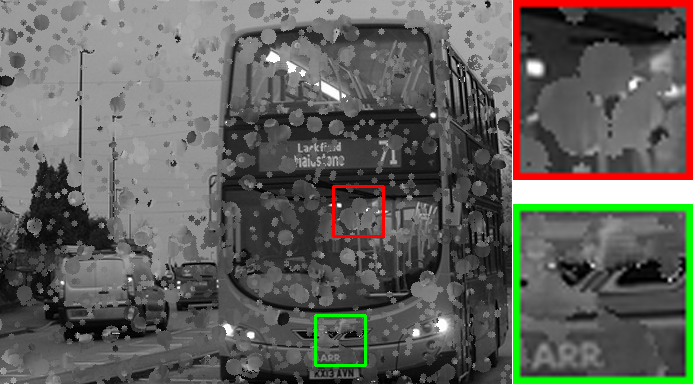}}
        &\frame{\includegraphics[width=\linewidth]{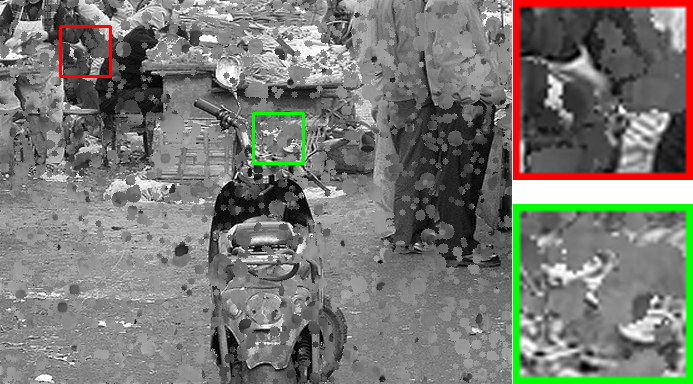}}
        &\frame{\includegraphics[width=\linewidth]{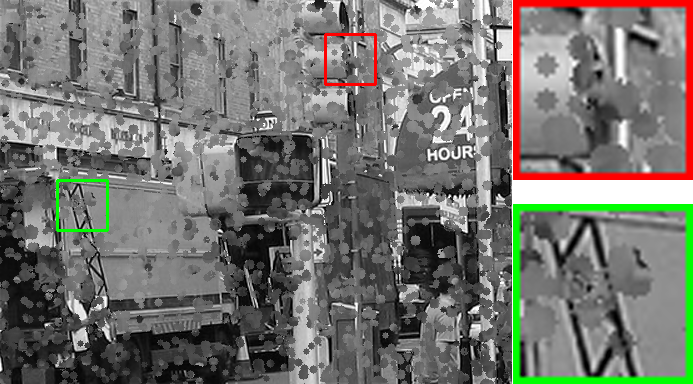}}
        &\frame{\includegraphics[width=\linewidth]{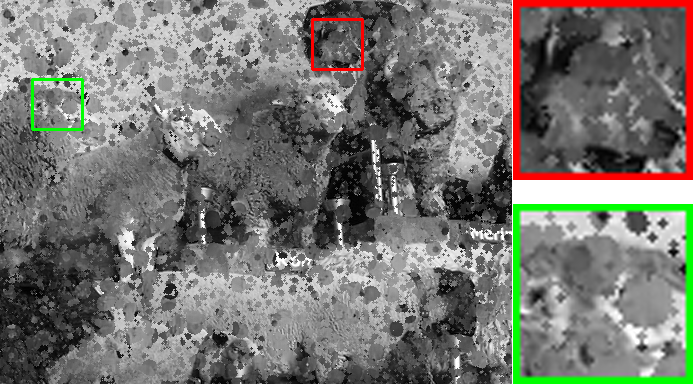}}
        \\
        \rotatebox{90}{\makecell{PUT \cite{liu22cvpr}}}  
        &\frame{\includegraphics[width=\linewidth]{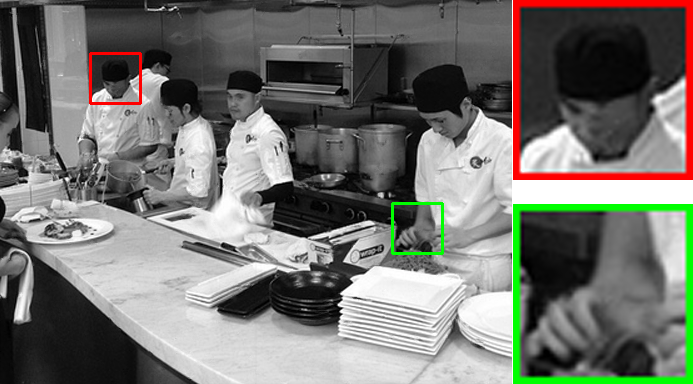}}
        &\frame{\includegraphics[width=\linewidth]{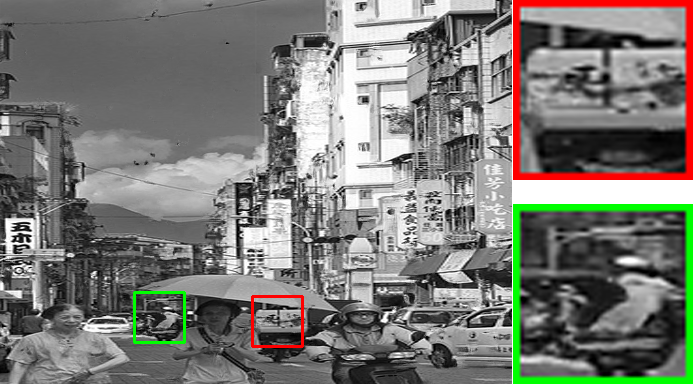}}
        &\frame{\includegraphics[width=\linewidth]{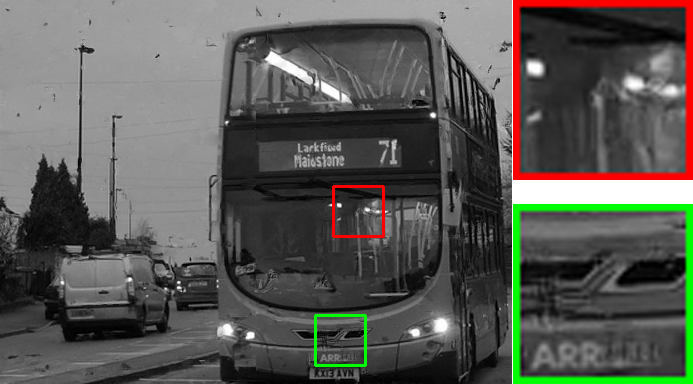}}
        &\frame{\includegraphics[width=\linewidth]{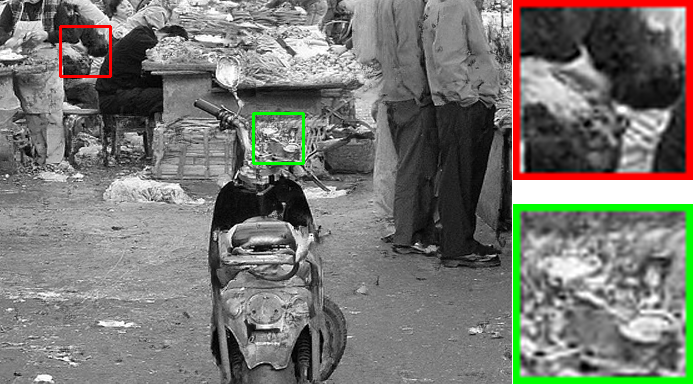}}
        &\frame{\includegraphics[width=\linewidth]{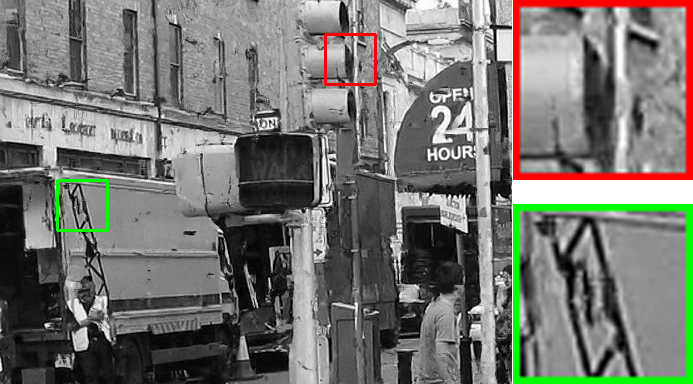}}
        &\frame{\includegraphics[width=\linewidth]{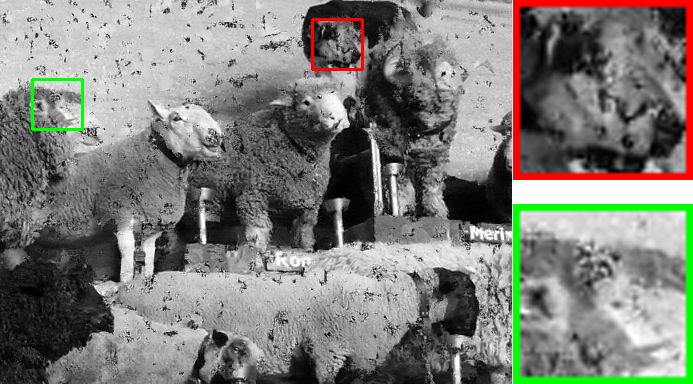}}
        \\
		\rotatebox{90}{\makecell{Ours (Acc.)}}  
        &\frame{\includegraphics[width=\linewidth]{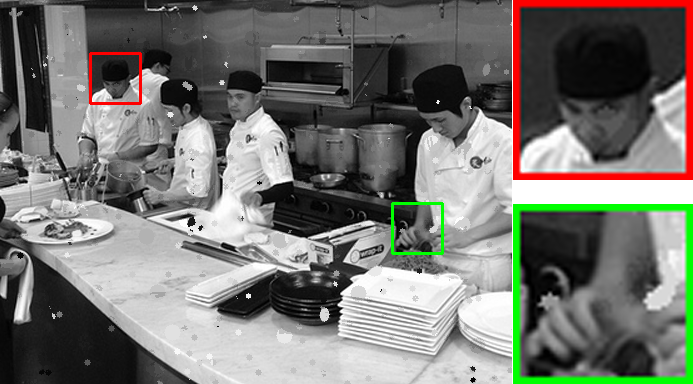}}
        &\frame{\includegraphics[width=\linewidth]{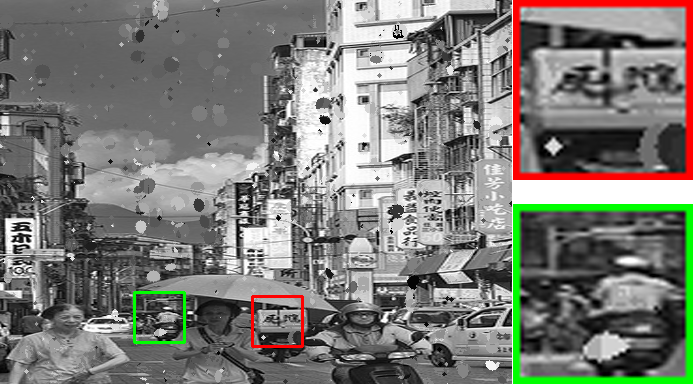}}
        &\frame{\includegraphics[width=\linewidth]{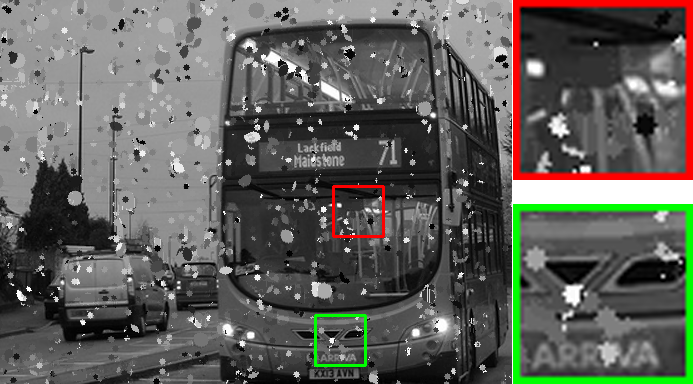}}
        &\frame{\includegraphics[width=\linewidth]{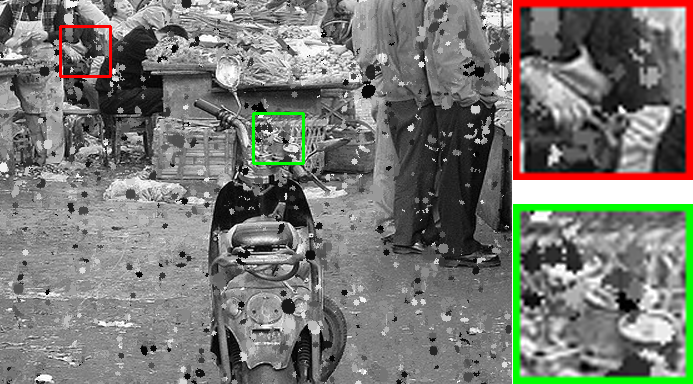}}
        &\frame{\includegraphics[width=\linewidth]{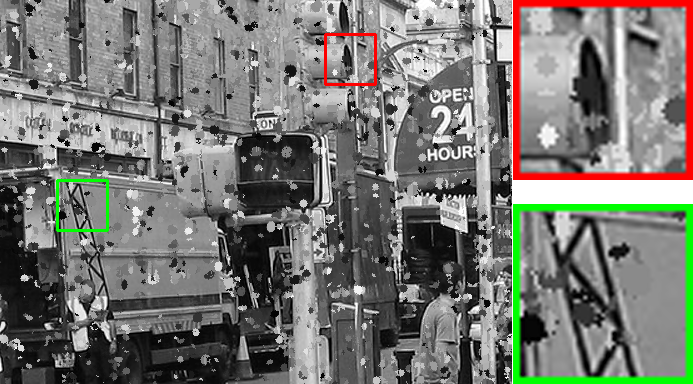}}
        &\frame{\includegraphics[width=\linewidth]{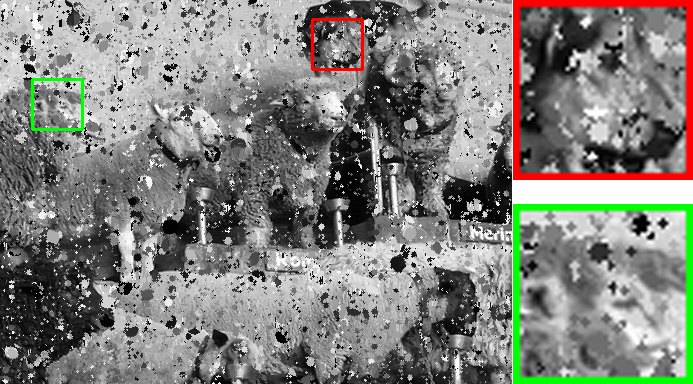}}
        \\ 
        \rotatebox{90}{\makecell{EF-SAI \cite{liao22cvpr}}}    
        &\frame{\includegraphics[width=\linewidth]{figures/images/qual_results_syn/efsai/0013.png}}
        &\frame{\includegraphics[width=\linewidth]{figures/images/qual_results_syn/efsai/0134.png}}
        &\frame{\includegraphics[width=\linewidth]{figures/images/qual_results_syn/efsai/0251.png}}
        &\frame{\includegraphics[width=\linewidth]{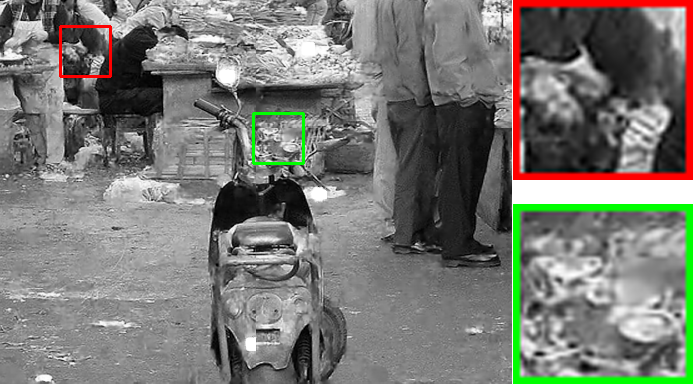}}
        &\frame{\includegraphics[width=\linewidth]{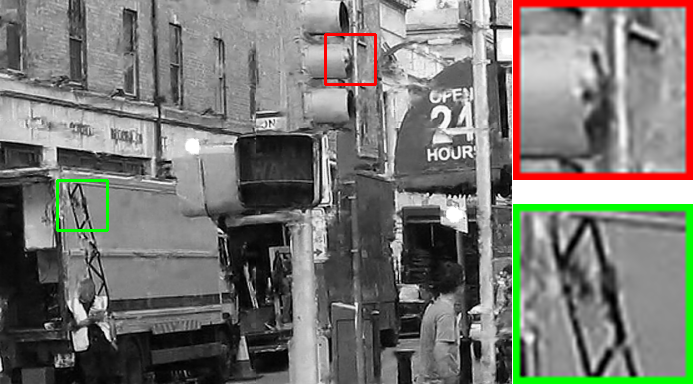}}
        &\frame{\includegraphics[width=\linewidth]{figures/images/qual_results_syn/efsai/0454.png}}
        \\       
        \rotatebox{90}{\makecell{MAT \cite{li_mat22cvpr}}}  
        &\frame{\includegraphics[width=\linewidth]{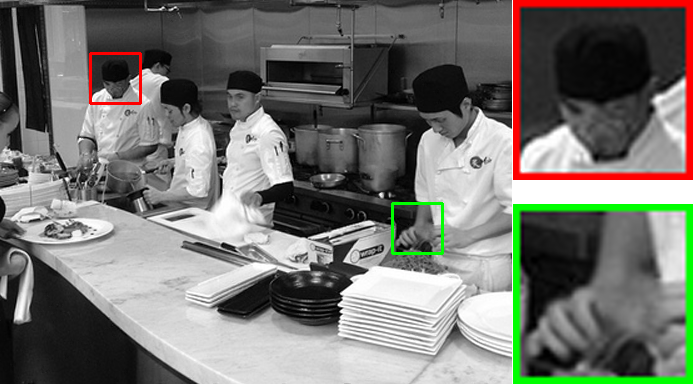}} 
        &\frame{\includegraphics[width=\linewidth]{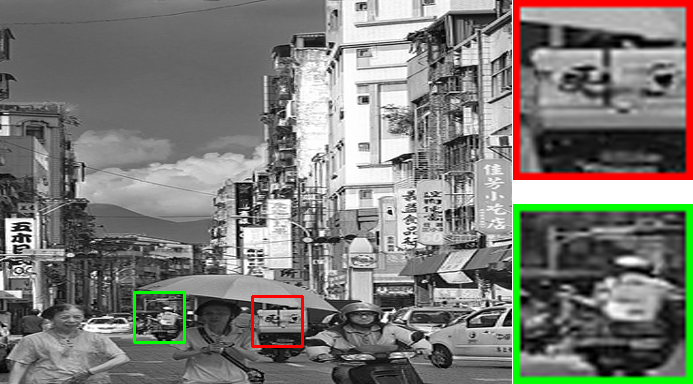}}
        &\frame{\includegraphics[width=\linewidth]{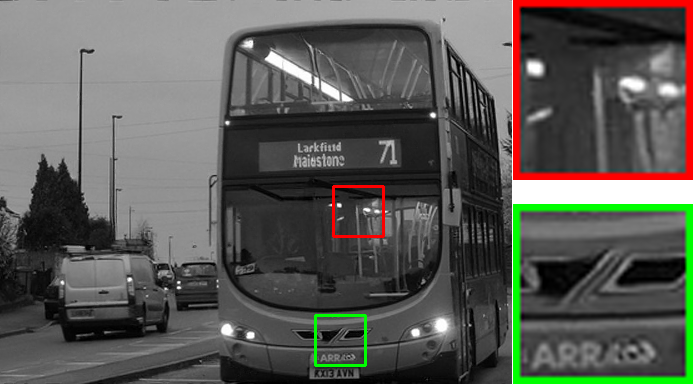}}
        &\frame{\includegraphics[width=\linewidth]{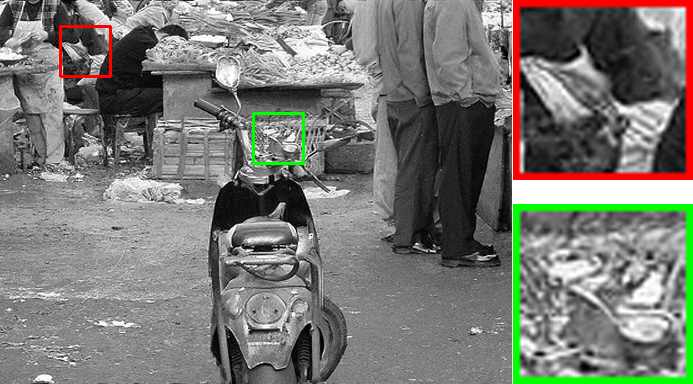}}
        &\frame{\includegraphics[width=\linewidth]{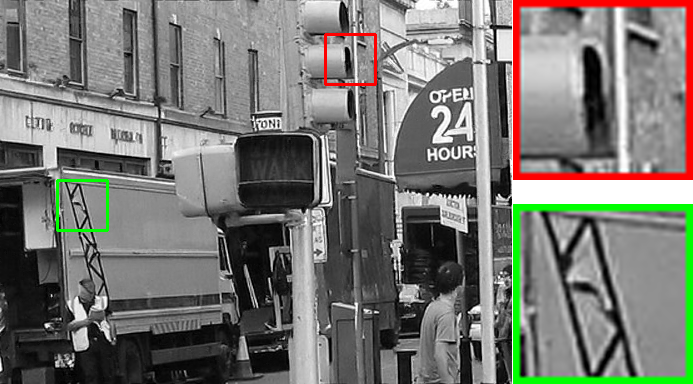}}
        &\frame{\includegraphics[width=\linewidth]{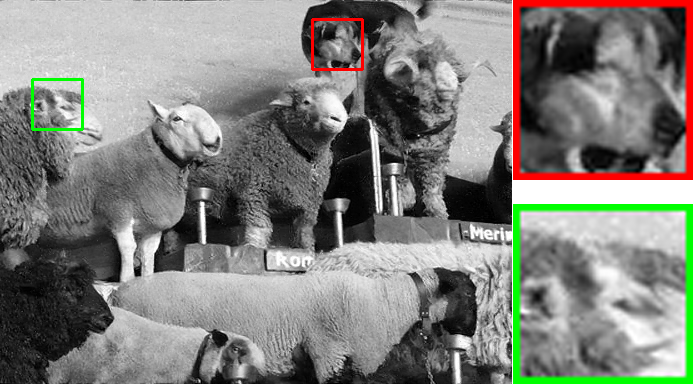}}
        \\        
		\rotatebox{90}{\makecell{MISF \cite{li_misf22cvpr}}}  
        &\frame{\includegraphics[width=\linewidth]{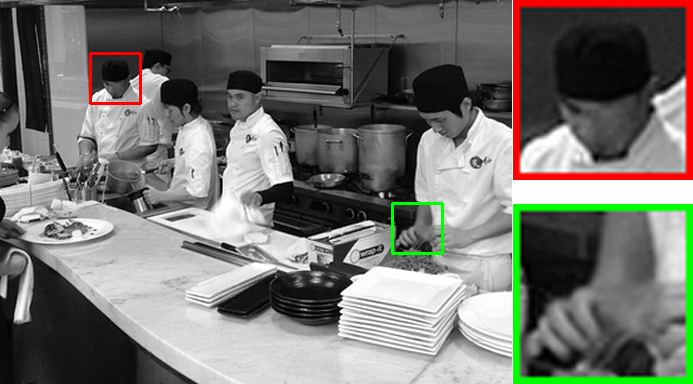}} 
        &\frame{\includegraphics[width=\linewidth]{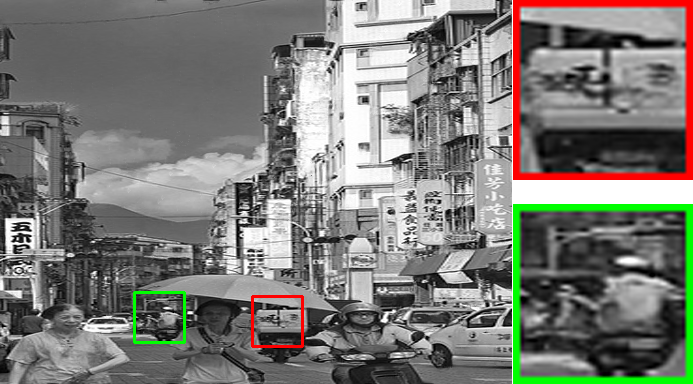}}
        &\frame{\includegraphics[width=\linewidth]{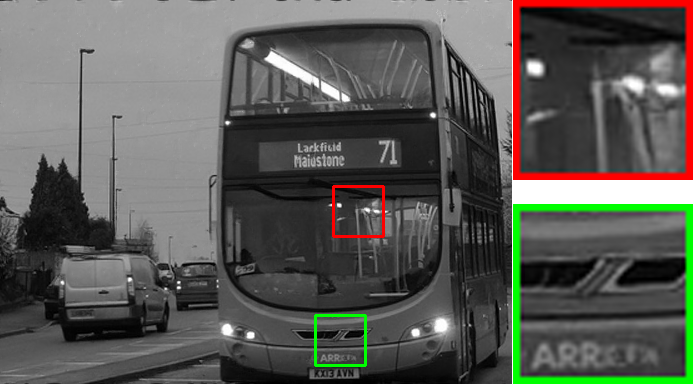}}
        &\frame{\includegraphics[width=\linewidth]{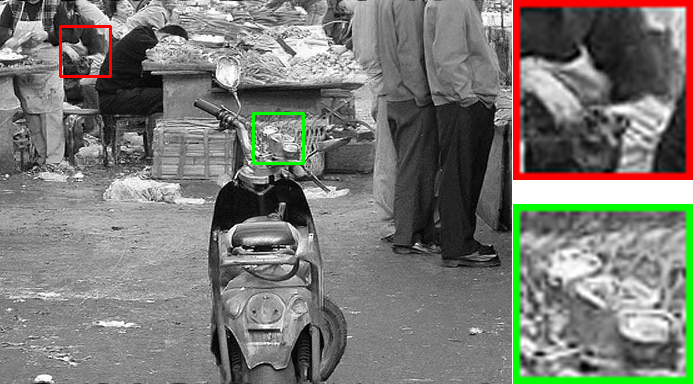}}
        &\frame{\includegraphics[width=\linewidth]{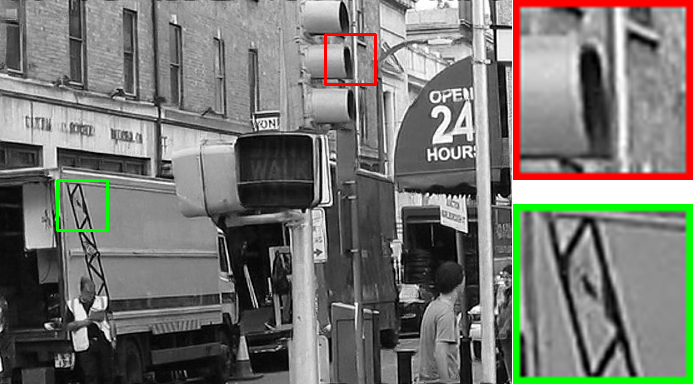}}
        &\frame{\includegraphics[width=\linewidth]{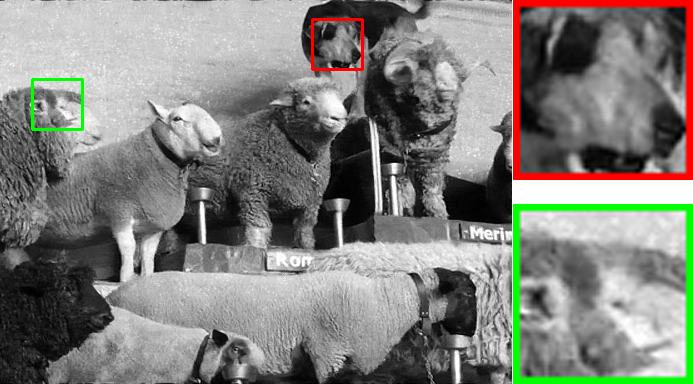}}
        \\
		\rotatebox{90}{\makecell{ZITS \cite{dong22cvpr}}}     
        &\frame{\includegraphics[width=\linewidth]{figures/images/qual_results_syn/zits/0013.png}} 
        & \frame{\includegraphics[width=\linewidth]{figures/images/qual_results_syn/zits/0134.png}}
        &\frame{\includegraphics[width=\linewidth]{figures/images/qual_results_syn/zits/0251.png}}
        &\frame{\includegraphics[width=\linewidth]{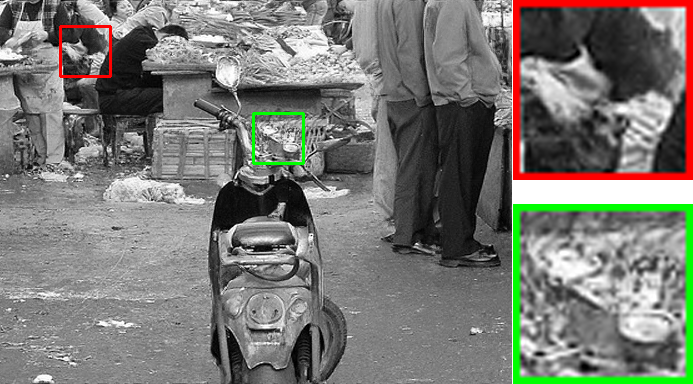}}
        &\frame{\includegraphics[width=\linewidth]{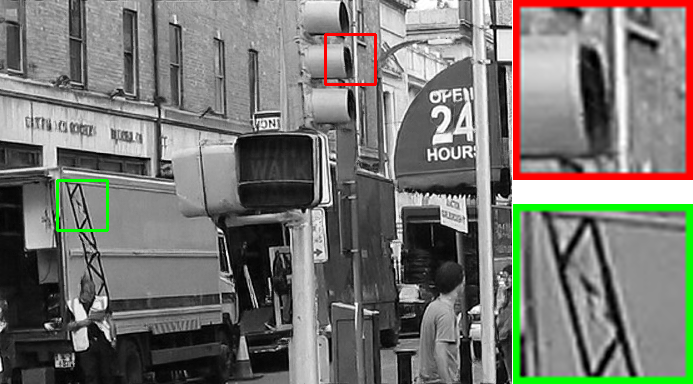}}
        &\frame{\includegraphics[width=\linewidth]{figures/images/qual_results_syn/zits/0454.png}}
        \\
		\rotatebox{90}{\makecell{Ours}}     
        &\frame{\includegraphics[width=\linewidth]{figures/images/qual_results_syn/ours/0013.png}} 
        & \frame{\includegraphics[width=\linewidth]{figures/images/qual_results_syn/ours/0134.png}}
        &\frame{\includegraphics[width=\linewidth]{figures/images/qual_results_syn/ours/0251.png}}
        &\frame{\includegraphics[width=\linewidth]{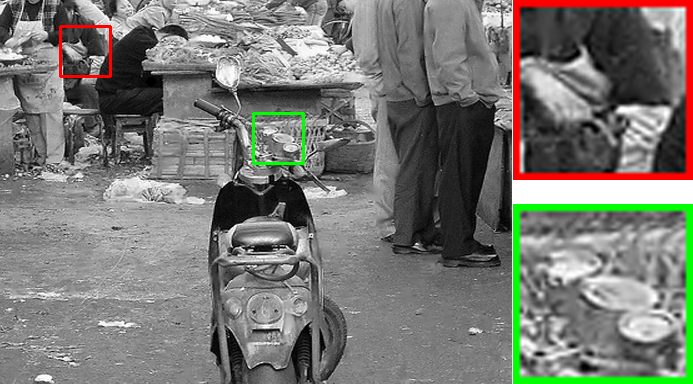}}
        &\frame{\includegraphics[width=\linewidth]{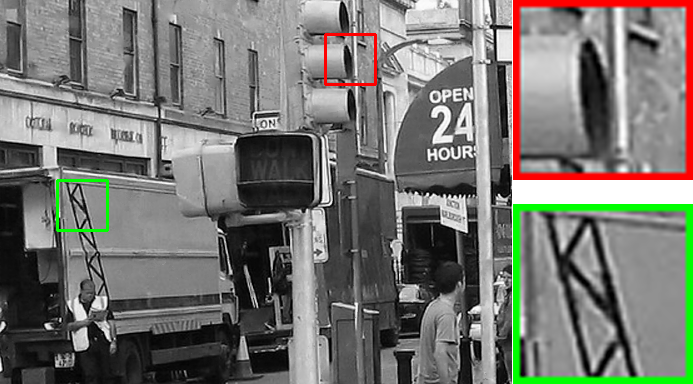}}
        &\frame{\includegraphics[width=\linewidth]{figures/images/qual_results_syn/ours/0454.png}}
        \\
		\rotatebox{90}{\makecell{Groundtruth}}       
        &\frame{\includegraphics[width=\linewidth]{figures/images/qual_results_syn/gt/0013.png}}
        &\frame{\includegraphics[width=\linewidth]{figures/images/qual_results_syn/gt/0134.png}}
        &\frame{\includegraphics[width=\linewidth]{figures/images/qual_results_syn/gt/0251.png}}
        &\frame{\includegraphics[width=\linewidth]{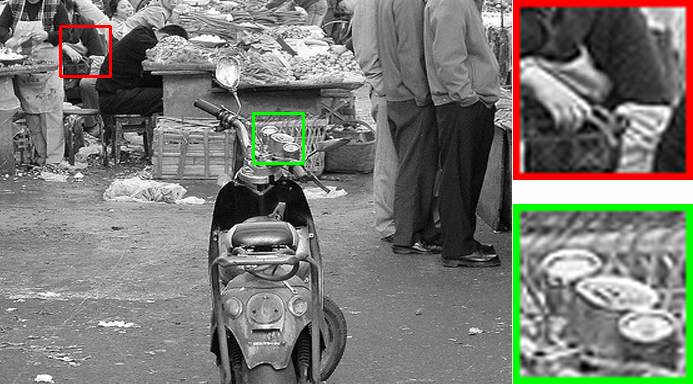}}
        &\frame{\includegraphics[width=\linewidth]{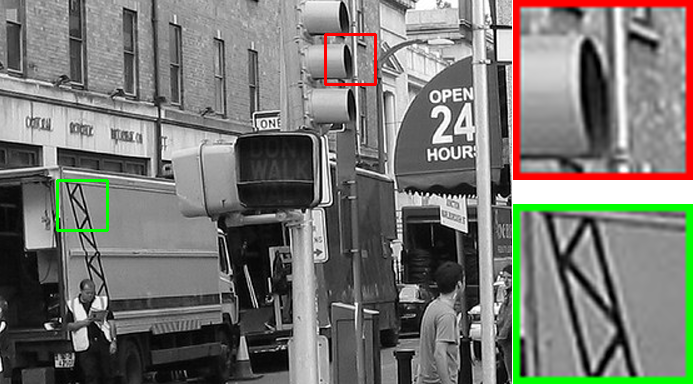}}
        &\frame{\includegraphics[width=\linewidth]{figures/images/qual_results_syn/gt/0454.png}}
      	\\

	\end{tabular}
    \vspace{-1ex}
	\caption{Images showing the occluded input frame, the reconstructed frame of all methods, and the ground truth frame for our synthetic dataset.
 }
	\label{fig:suple_syn_qual_all}
\end{figure*}

\global\long\def\figWidth{0.165\linewidth}
\begin{figure*}
	\centering
    \setlength{\tabcolsep}{1pt}
	\begin{tabular}{
	M{0.25cm}
	M{\figWidth}
	M{\figWidth}
	M{\figWidth}
	M{\figWidth}
	M{\figWidth}
	M{\figWidth}
 }
		\\
		\rotatebox{90}{\makecell{Occluded}} 
        & \includegraphics[width=\linewidth]{figures/images/qual_results_real/occ/1selected.png} 
        &  \includegraphics[width=\linewidth]{figures/images/qual_results_real/occ/4selected.png}
        &  \includegraphics[width=\linewidth]{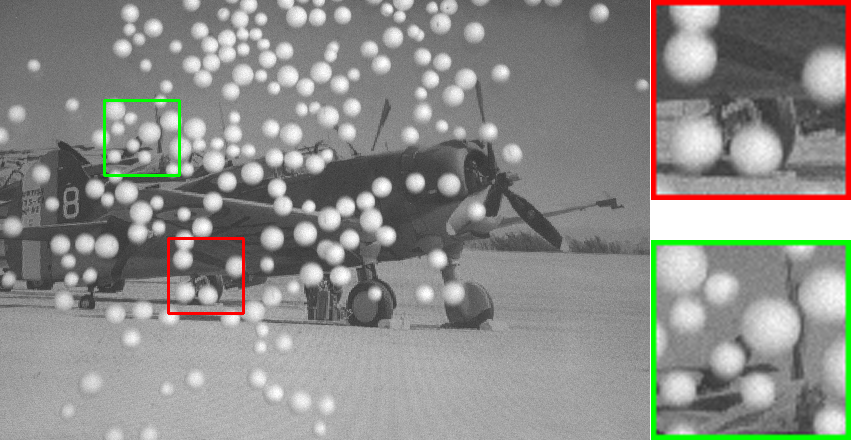}
        &  \includegraphics[width=\linewidth]{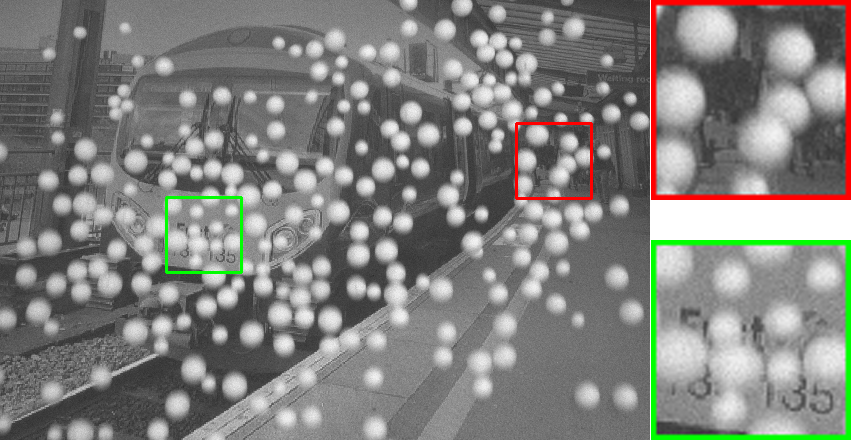}
        &  \includegraphics[width=\linewidth]{figures/images/qual_results_real/occ/31selected.png}
        &  \includegraphics[width=\linewidth]{figures/images/qual_results_real/occ/59selected.png}
        \\
        \rotatebox{90}{\makecell{E2VID \cite{Rebecq19cvpr}}}    
        &\frame{\includegraphics[width=\linewidth]{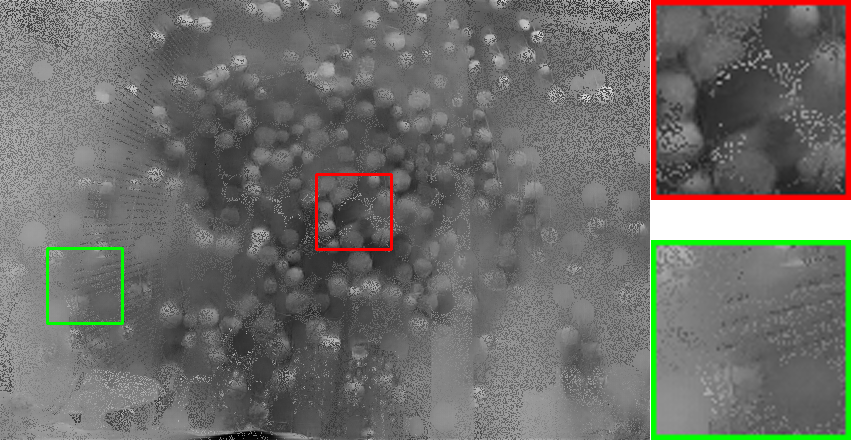}}
        &\frame{\includegraphics[width=\linewidth]{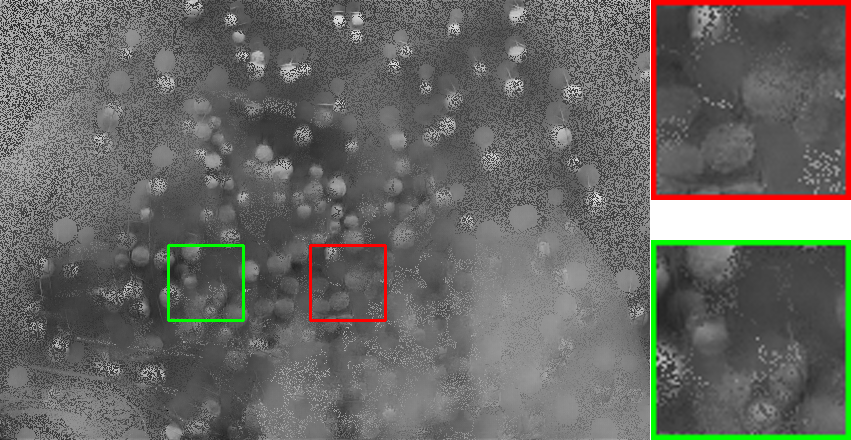}}
        &\frame{\includegraphics[width=\linewidth]{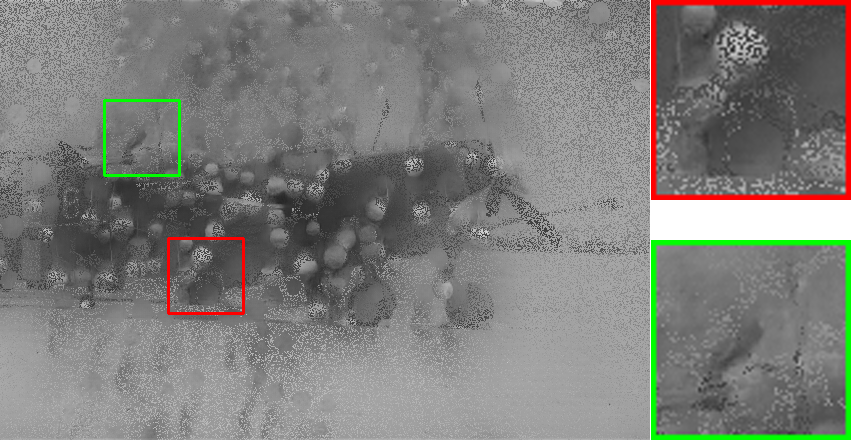}}
        &\frame{\includegraphics[width=\linewidth]{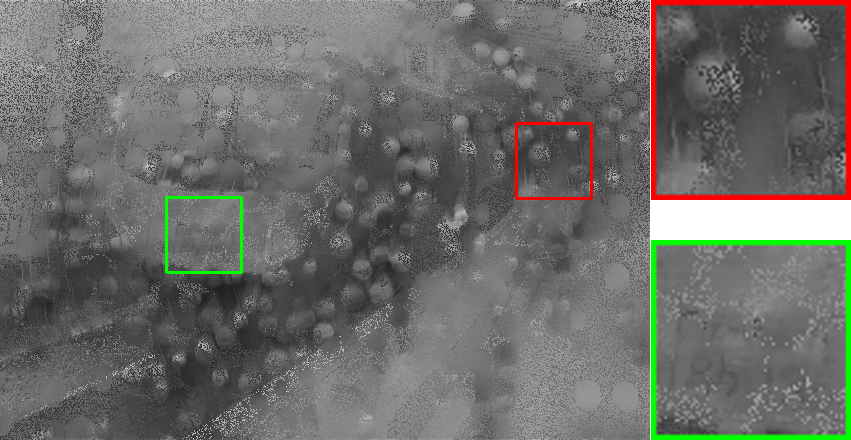}}
        &\frame{\includegraphics[width=\linewidth]{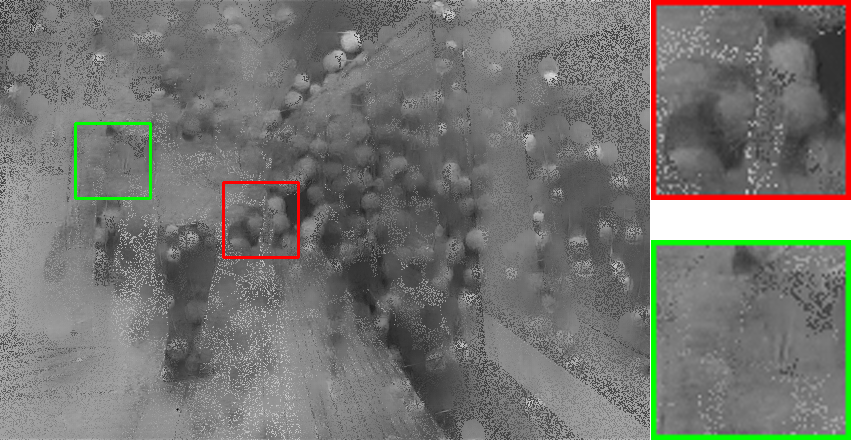}}
        &\frame{\includegraphics[width=\linewidth]{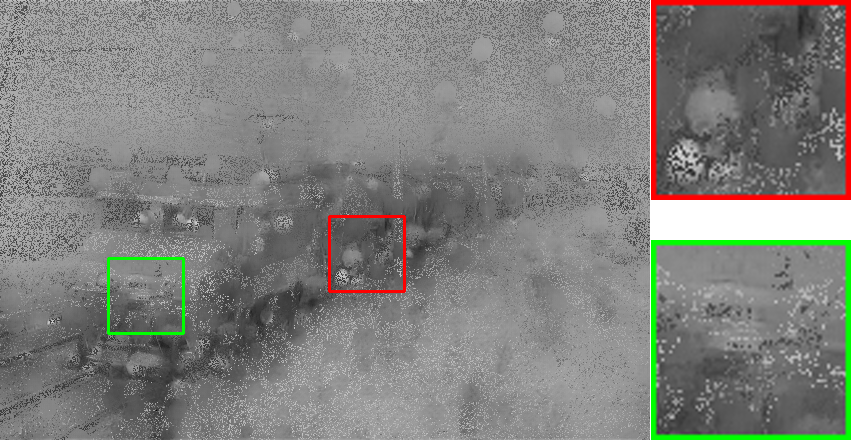}}
        \\
        \rotatebox{90}{\makecell{PUT \cite{liu22cvpr}}}  
        &\frame{\includegraphics[width=\linewidth]{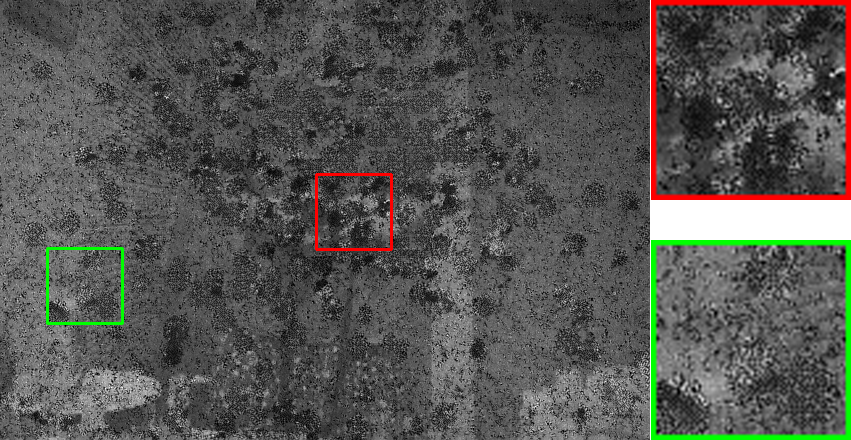}}
        &\frame{\includegraphics[width=\linewidth]{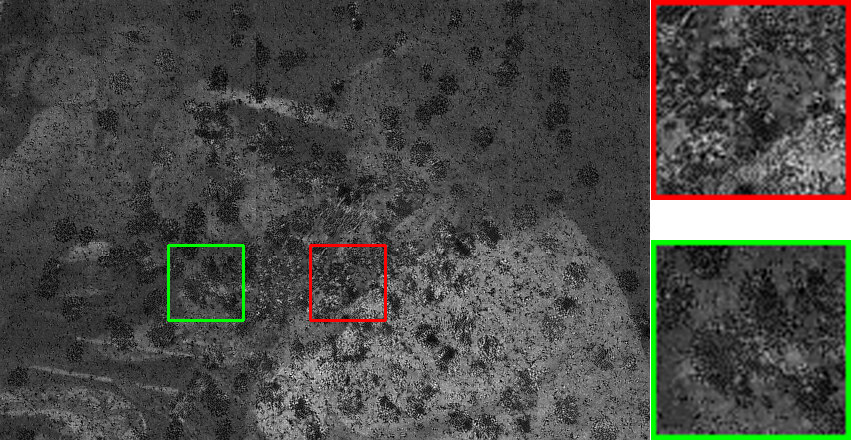}}
        &\frame{\includegraphics[width=\linewidth]{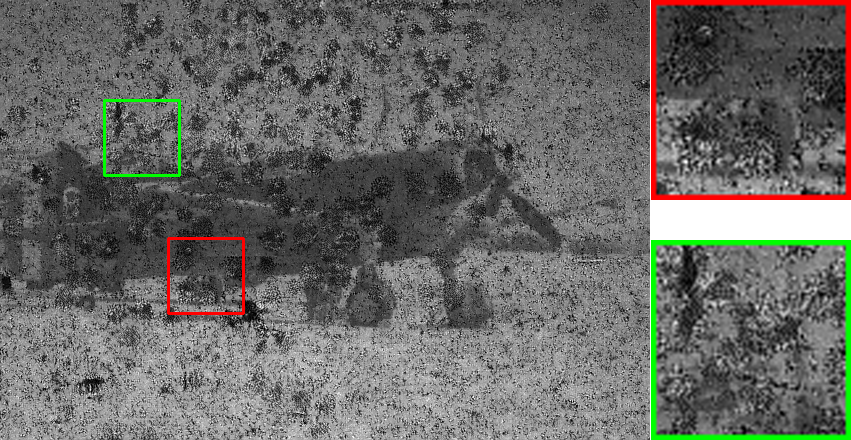}}
        &\frame{\includegraphics[width=\linewidth]{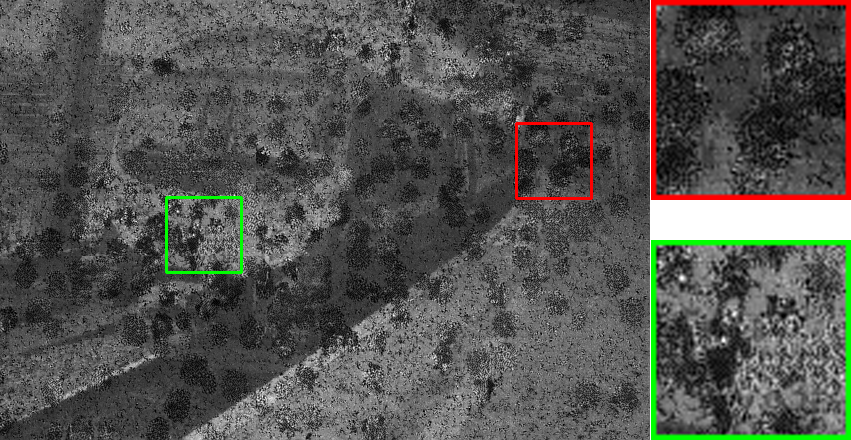}}
        &\frame{\includegraphics[width=\linewidth]{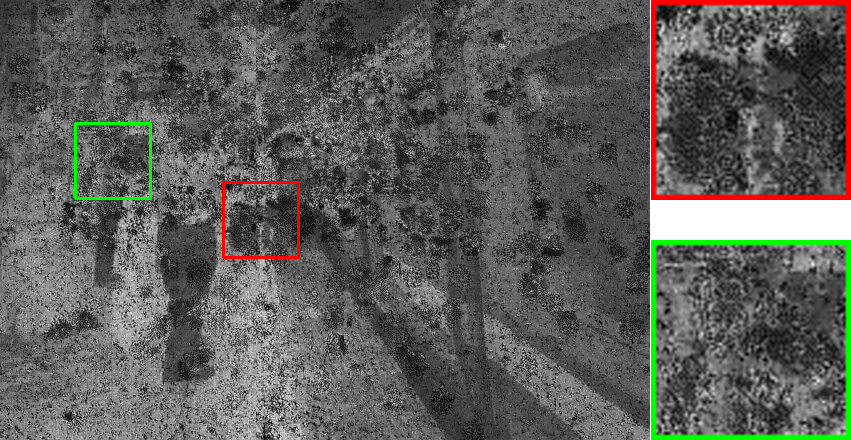}}
        &\frame{\includegraphics[width=\linewidth]{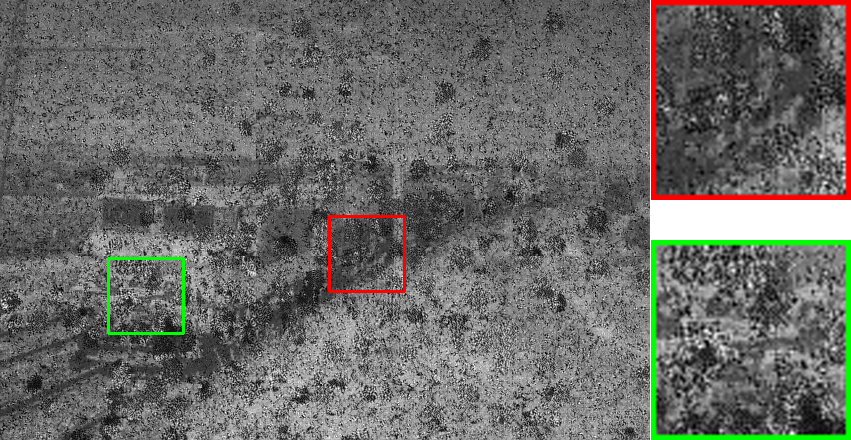}}
        \\
		\rotatebox{90}{\makecell{Ours (Acc.)}}  
        &\frame{\includegraphics[width=\linewidth]{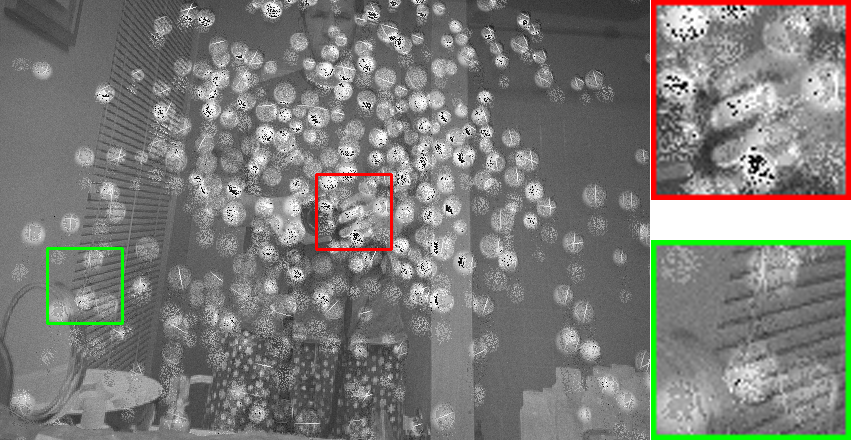}}
        &\frame{\includegraphics[width=\linewidth]{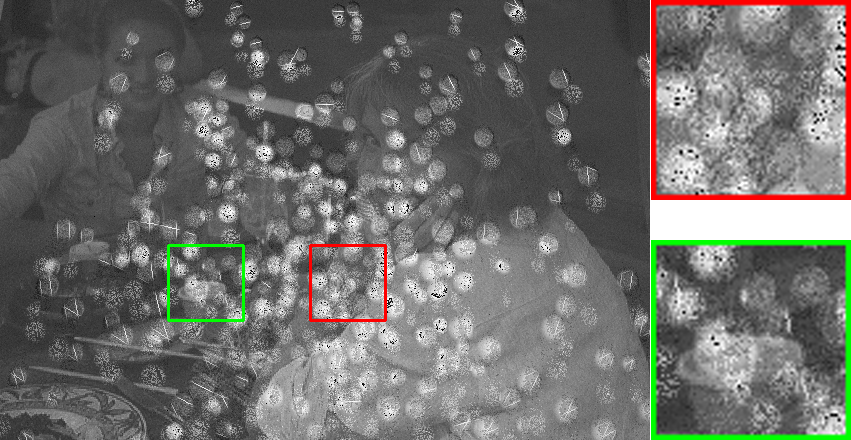}}
        &\frame{\includegraphics[width=\linewidth]{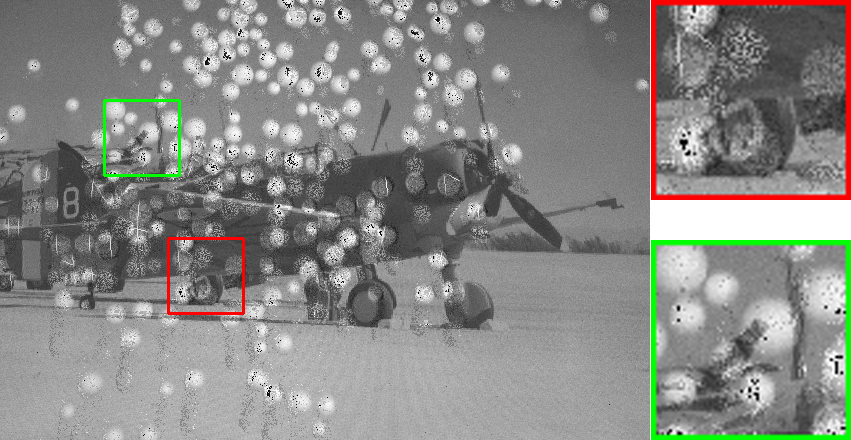}}
        &\frame{\includegraphics[width=\linewidth]{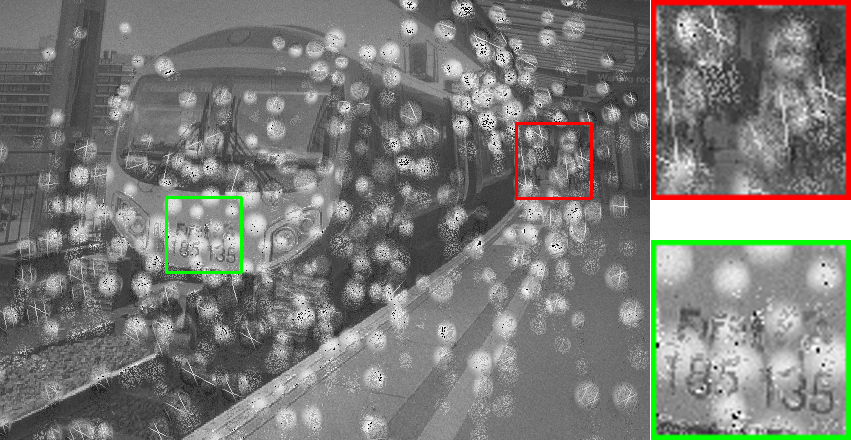}}
        &\frame{\includegraphics[width=\linewidth]{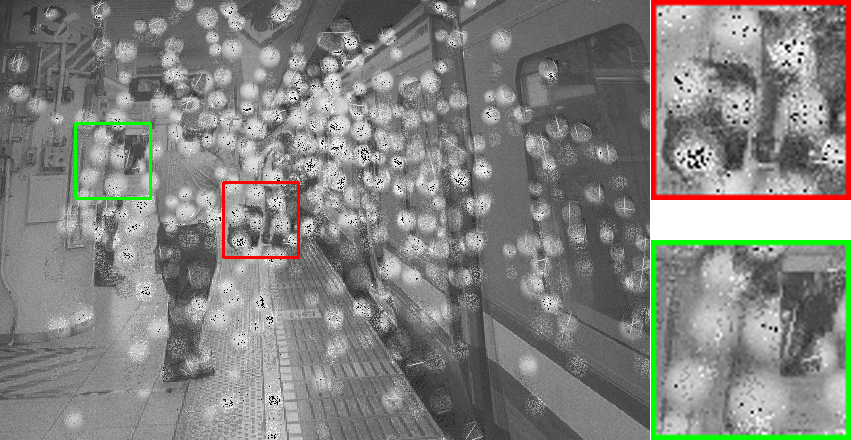}}
        &\frame{\includegraphics[width=\linewidth]{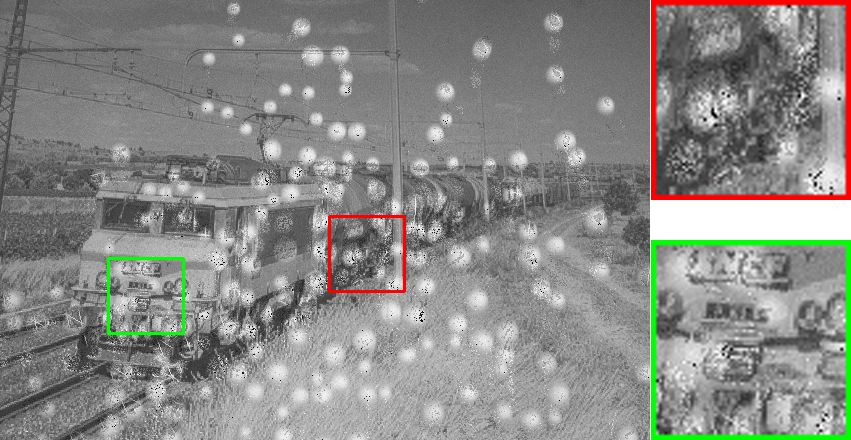}}
        \\ 
        \rotatebox{90}{\makecell{EF-SAI \cite{liao22cvpr}}}    
        &\frame{\includegraphics[width=\linewidth]{figures/images/qual_results_real/efsai/1selected.png}}
        &\frame{\includegraphics[width=\linewidth]{figures/images/qual_results_real/efsai/4selected.png}}
        &\frame{\includegraphics[width=\linewidth]{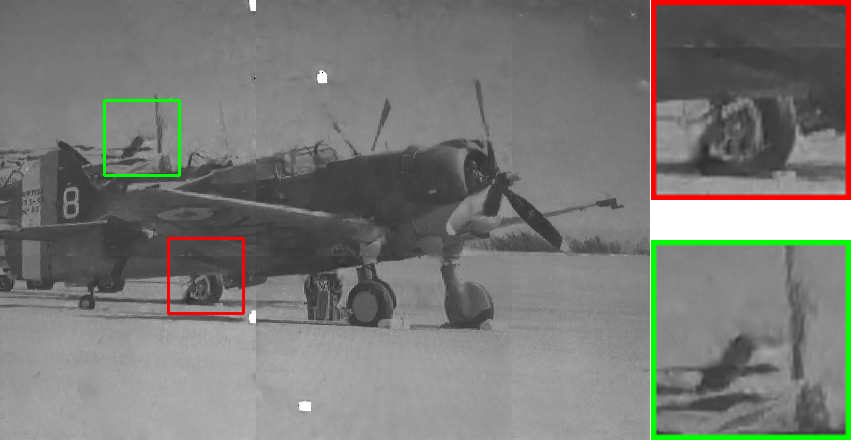}}
        &\frame{\includegraphics[width=\linewidth]{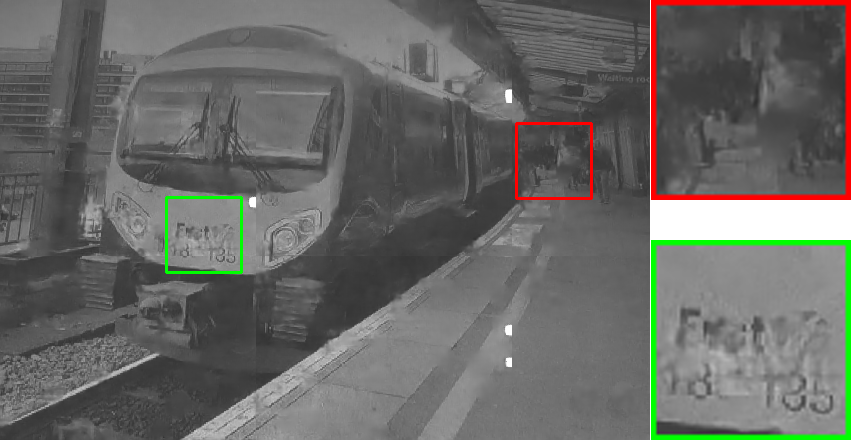}}
        &\frame{\includegraphics[width=\linewidth]{figures/images/qual_results_real/efsai/31selected.png}}
        &\frame{\includegraphics[width=\linewidth]{figures/images/qual_results_real/efsai/59selected.png}}
        \\       
        \rotatebox{90}{\makecell{MAT \cite{li_mat22cvpr}}}  
        &\frame{\includegraphics[width=\linewidth]{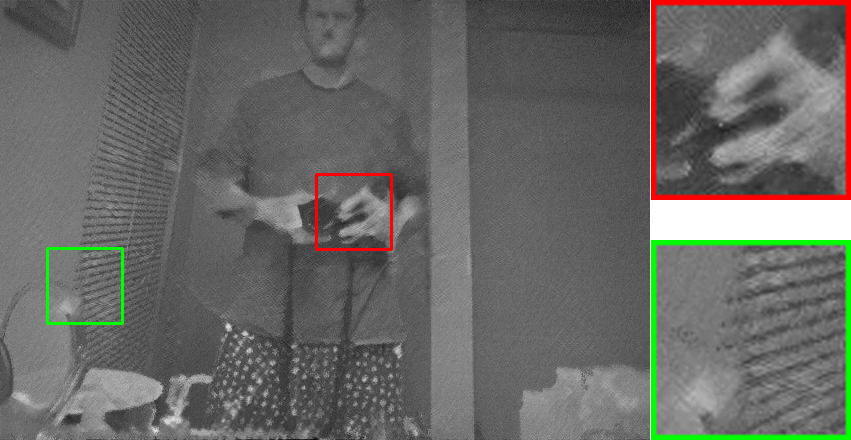}} 
        &\frame{\includegraphics[width=\linewidth]{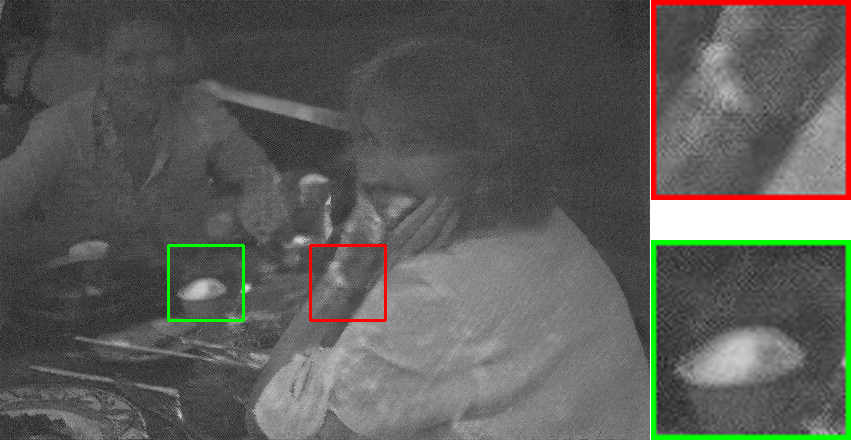}}
        &\frame{\includegraphics[width=\linewidth]{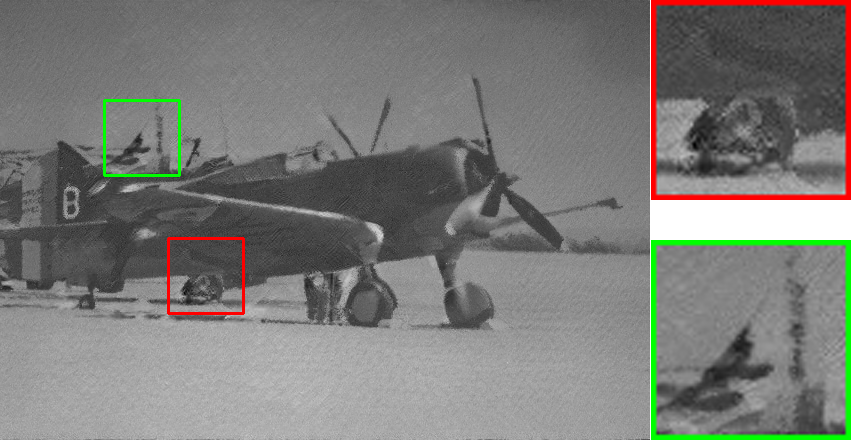}}
        &\frame{\includegraphics[width=\linewidth]{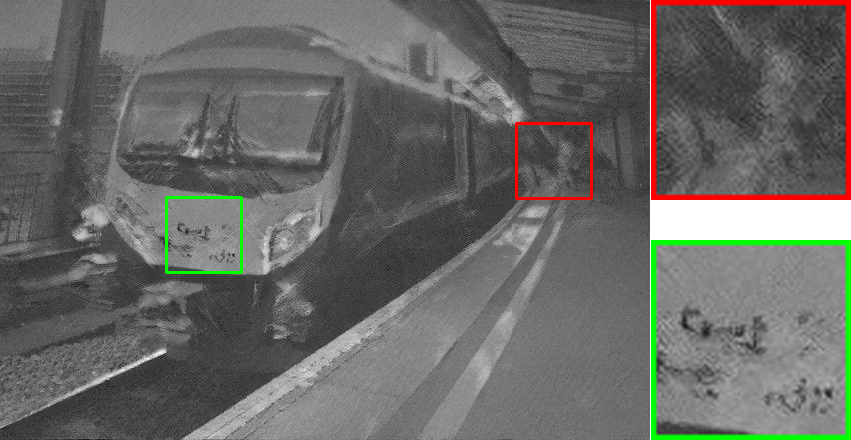}}
        &\frame{\includegraphics[width=\linewidth]{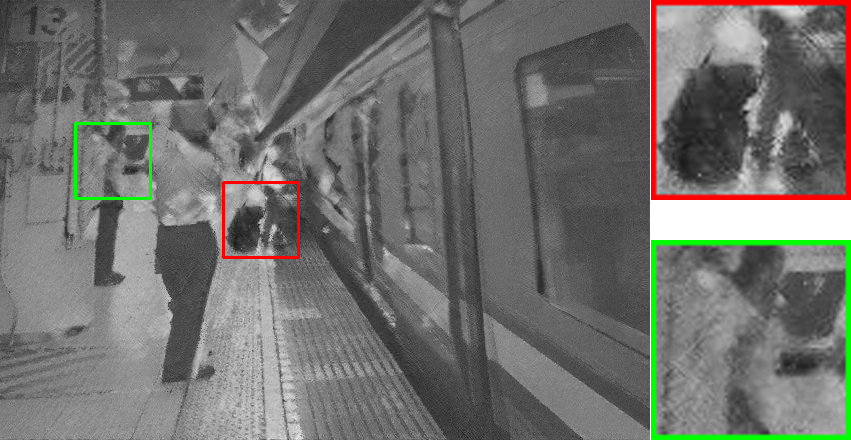}}
        &\frame{\includegraphics[width=\linewidth]{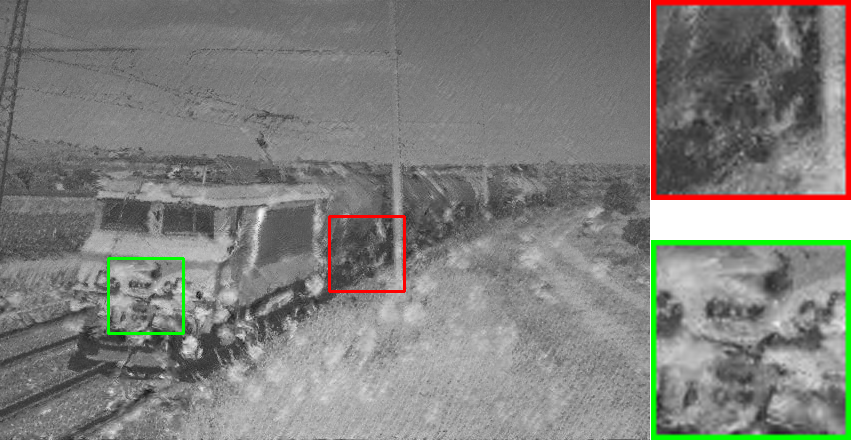}}
        \\        
		\rotatebox{90}{\makecell{MISF \cite{li_misf22cvpr}}}  
        &\frame{\includegraphics[width=\linewidth]{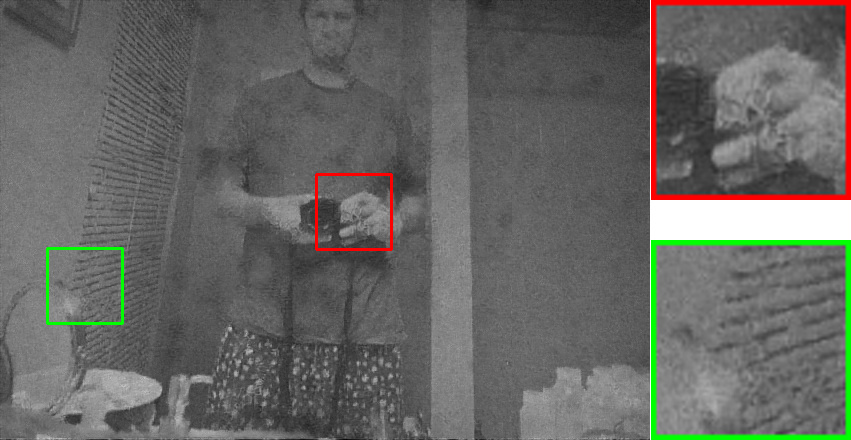}} 
        &\frame{\includegraphics[width=\linewidth]{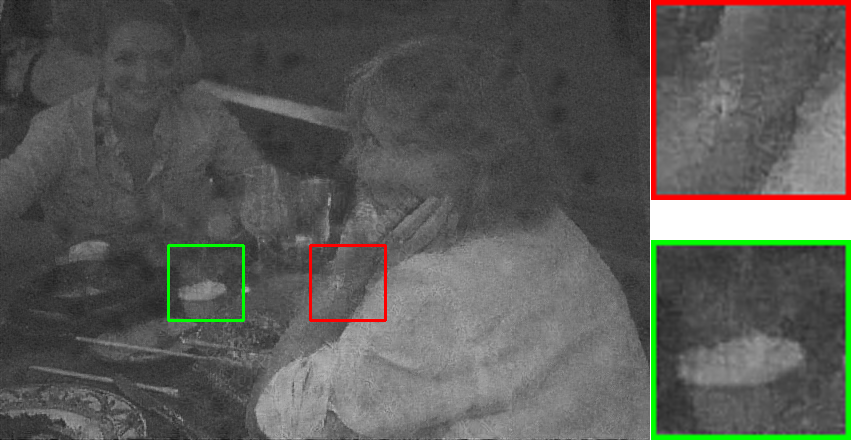}}
        &\frame{\includegraphics[width=\linewidth]{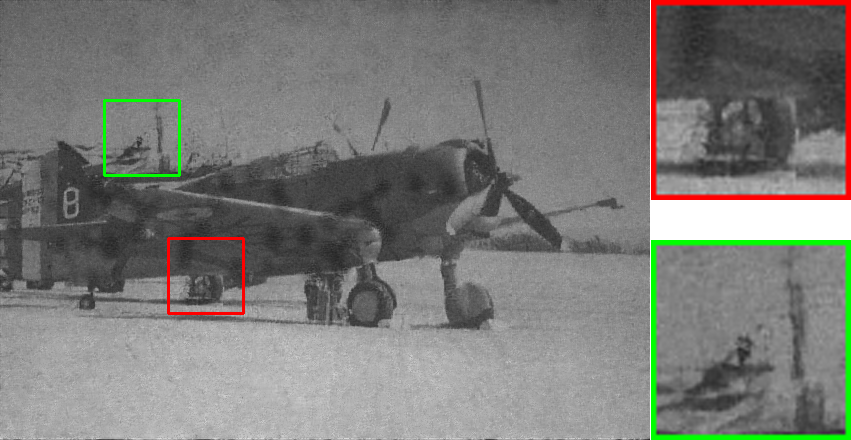}}
        &\frame{\includegraphics[width=\linewidth]{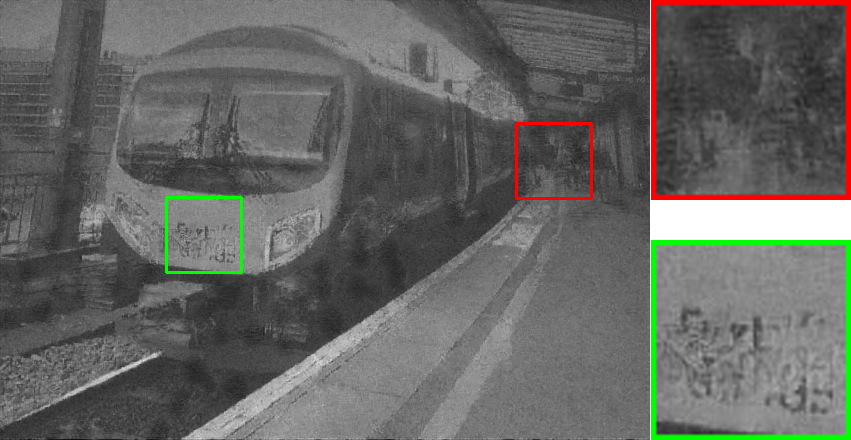}}
        &\frame{\includegraphics[width=\linewidth]{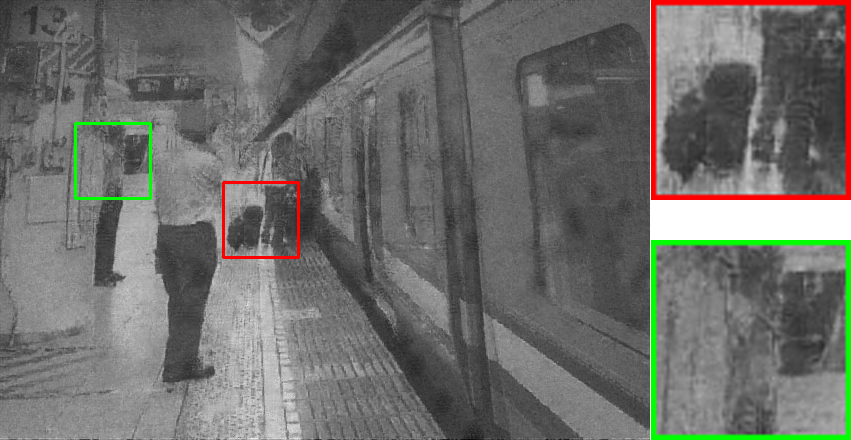}}
        &\frame{\includegraphics[width=\linewidth]{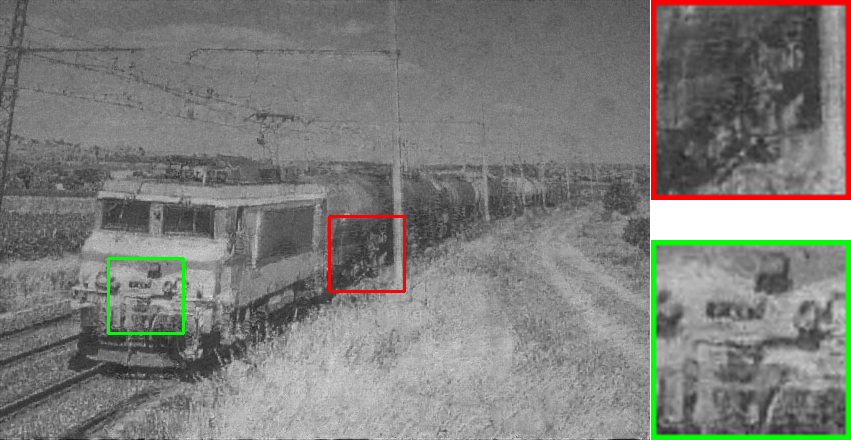}}
        \\
		\rotatebox{90}{\makecell{ZITS \cite{dong22cvpr}}}     
        &\frame{\includegraphics[width=\linewidth]{figures/images/qual_results_real/zits/1selected.png}} 
        & \frame{\includegraphics[width=\linewidth]{figures/images/qual_results_real/zits/4selected.png}}
        &\frame{\includegraphics[width=\linewidth]{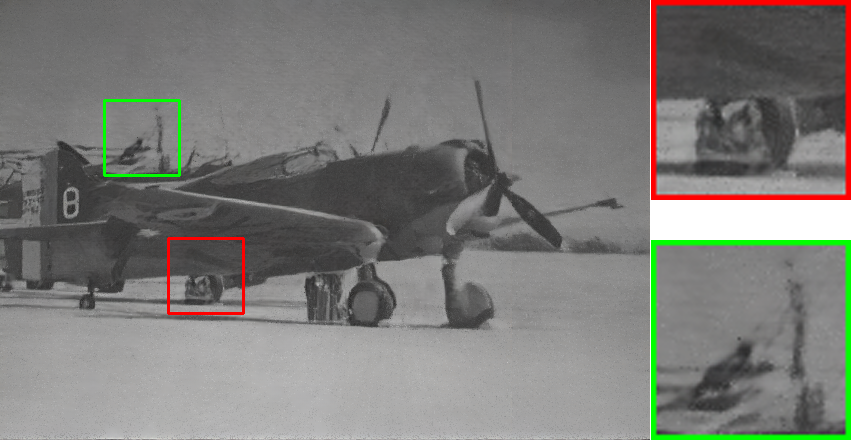}}
        &\frame{\includegraphics[width=\linewidth]{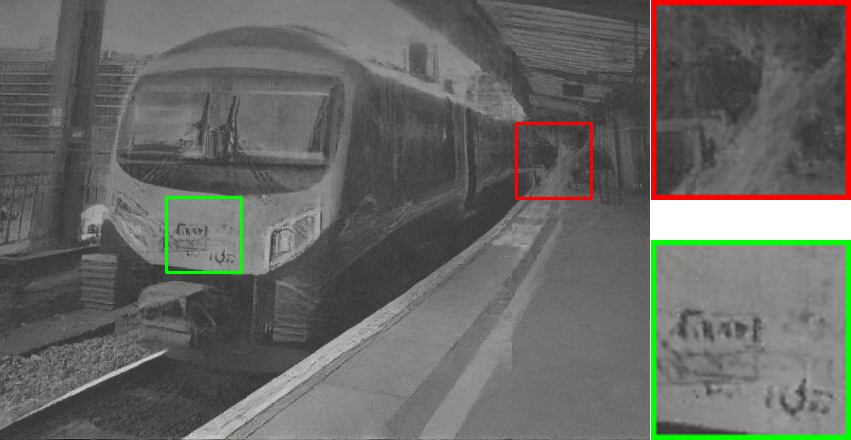}}
        &\frame{\includegraphics[width=\linewidth]{figures/images/qual_results_real/zits/31selected.png}}
        &\frame{\includegraphics[width=\linewidth]{figures/images/qual_results_real/zits/59selected.png}}
        \\
		\rotatebox{90}{\makecell{Ours}}     
        &\frame{\includegraphics[width=\linewidth]{figures/images/qual_results_real/ours/1selected.png}} 
        & \frame{\includegraphics[width=\linewidth]{figures/images/qual_results_real/ours/4selected.png}}
        &\frame{\includegraphics[width=\linewidth]{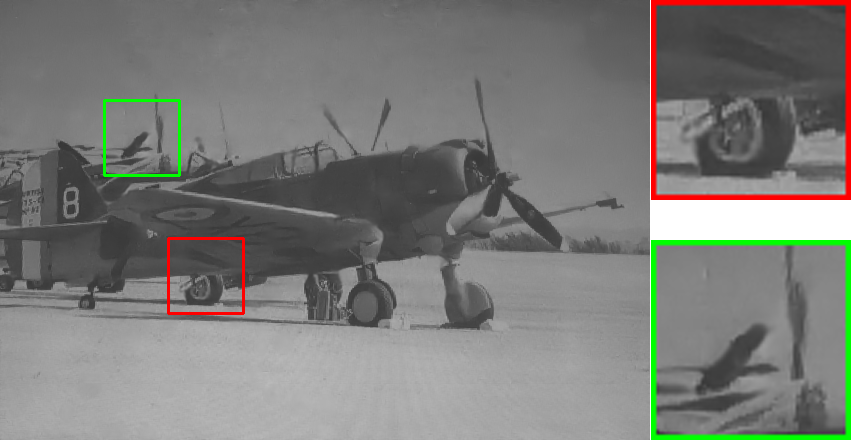}}
        &\frame{\includegraphics[width=\linewidth]{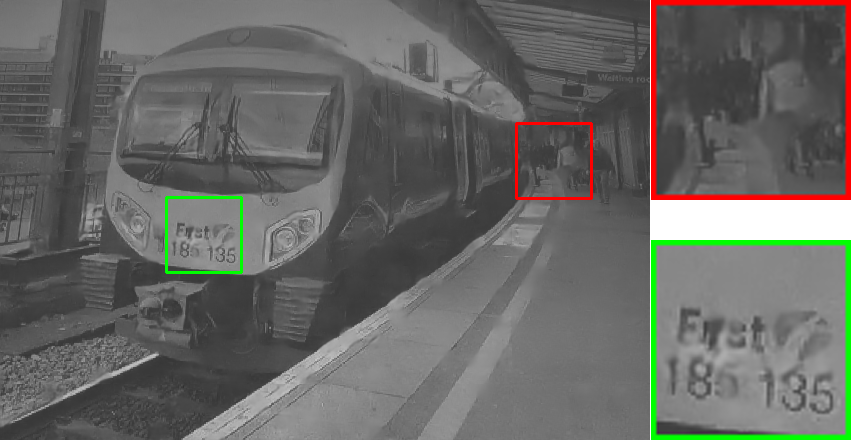}}
        &\frame{\includegraphics[width=\linewidth]{figures/images/qual_results_real/ours/31selected.png}}
        &\frame{\includegraphics[width=\linewidth]{figures/images/qual_results_real/ours/59selected.png}}
        \\
		\rotatebox{90}{\makecell{Groundtruth}}       
        &\frame{\includegraphics[width=\linewidth]{figures/images/qual_results_real/gt/1selected.png}}
        &\frame{\includegraphics[width=\linewidth]{figures/images/qual_results_real/gt/4selected.png}}
        &\frame{\includegraphics[width=\linewidth]{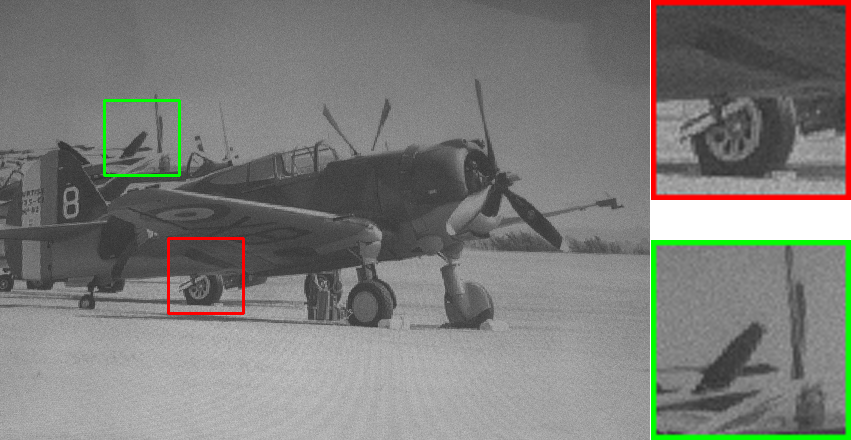}}
        &\frame{\includegraphics[width=\linewidth]{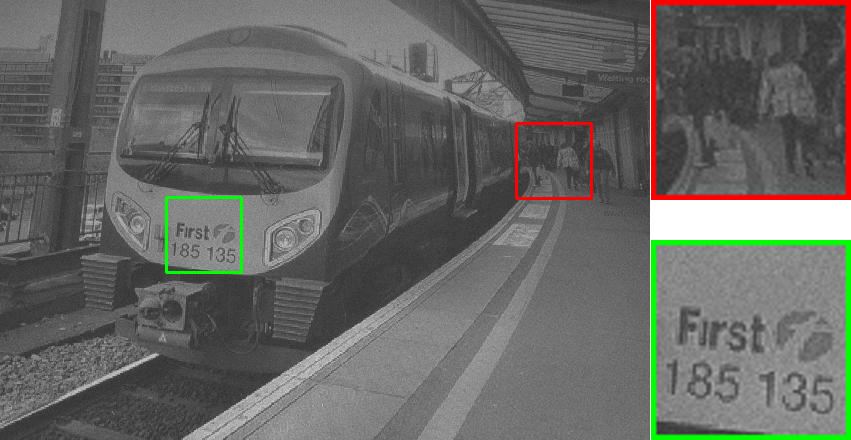}}
        &\frame{\includegraphics[width=\linewidth]{figures/images/qual_results_real/gt/31selected.png}}
        &\frame{\includegraphics[width=\linewidth]{figures/images/qual_results_real/gt/59selected.png}}
      	\\

	\end{tabular}
    \vspace{-1ex}
	\caption{Images showing the occluded input frame, the reconstructed frame of all methods, and the ground truth frame for our real-world dataset.
 }
	\label{fig:suple_real_qual_all}
\end{figure*}

We show the qualitative comparisons between our method and all the baselines on our synthetic and real-world datasets in \Fig \ref{fig:suple_syn_qual_all} and \Fig \ref{fig:suple_real_qual_all} respectively.
Qualitatively, our method outperforms other baselines for both datasets.
We re-state our conclusions that image inpainting baselines tend to hallucinate the missing areas, resulting in clean but inaccurate image reconstructions.
In comparison, our method is better at preserving the details of the original image, as can be seen in the highlighted patches.

\section{Inference Time}
In this section, we compare the computational complexity of the methods in terms of inference time.
Note that while it is not feasible to compute the precise inference time, as GPU loads vary over time due to other processes running simultaneously, we approximate this number by averaging the inference time over the entire test set.
The runtime is computed for a batch size of 1 using a Quadro RTX 8000 GPU for the learning-based baselines and Intel(R) Core(TM) i7-3720QM CPU @ 2.60GHz for our model-based baseline.
The comparison for all the baselines is presented in \Tab \ref{tab:inference}. 
Note that only our model-based baseline is run on CPU and therefore marked with a star.
Our method has an inference time of \SI{0.18}{\sec}, slightly slower than the fastest image inpainting baseline MAT \cite{li_mat22cvpr}. The best performing image-inpainting baseline ZITS \cite{dong22cvpr}, on the other hand, has an inference time of \SI{1.94}{\sec}.
\begin{table}[!h]
    \centering
    \begin{adjustbox}{max width=\linewidth}
    \setlength{\tabcolsep}{4pt}
    {\small
    \begin{tabular}{lcccccccccc}
        \toprule
        Method &  MAT \cite{li_mat22cvpr} & MISF \cite{li_misf22cvpr}  & PUT \cite{liu22cvpr} &  ZITS \cite{dong22cvpr} & EF-SAI \cite{liao22cvpr} & E2VID \cite{Rebecq19cvpr}& Ours (Acc. Method)* & Ours (Learning)\\ 
        \midrule
        Input & I & I & I & I & I+E & E & E & I+E \\
        Infer. time & 0.15  & 0.16 & $>$60 &  1.94 &  3.13  & 0.01 & 0.001 &  0.18\\

        \bottomrule
    \end{tabular}}
    \end{adjustbox}
    \caption{Comparison of inference time (\SI{}{\sec}).}
    \label{tab:inference}
\end{table}

{\small
\bibliographystyle{ieee_fullname}
\bibliography{all}
}

\end{document}